\documentclass[hidelinks,twoside,11pt]{article}
\usepackage{hyperref}

\usepackage{jair}
\usepackage{theapa}
\usepackage{rawfonts}
\jairheading{73}{2022}{1435-1471}{12/2021}{04/2022}
\ShortHeadings{Adversarial Framework with Certified Robustness for Time-Series Domain}
{Belkhouja \& Doppa}
\firstpageno{1}

\usepackage{graphicx}
\usepackage{amsmath}
\usepackage{amssymb}
\usepackage{amsthm}
\usepackage{amsfonts} 
\usepackage{float}
\usepackage{tcolorbox}
\usepackage{algorithm}
\usepackage{algorithmic}

\usepackage{booktabs}       
\usepackage{tcolorbox}
\numberwithin{equation}{section}
\newtheorem{lemma}{Lemma}
\newtheorem{theorem}{Theorem}

\begin{document}
\setcounter{page}{1435}
\title{Adversarial Framework with Certified Robustness for Time-Series Domain via Statistical Features}

\author{\name Taha Belkhouja \email taha.belkhouja@wsu.edu \\
      \name Janardhan Rao Doppa \email jana.doppa@wsu.edu \\
      \addr School of Electrical Engineering and Computer Science\\ \addr Washington State University \\ \addr Pullman, Washington 99163, USA}

\maketitle

\begin{abstract}
Time-series data arises in many real-world applications (e.g., mobile health) and deep neural networks (DNNs) have shown great success in solving them. Despite their success, little is known about their robustness to adversarial attacks. In this paper, we propose a novel adversarial framework referred to as {\em {\bf T}ime-{\bf S}eries {\bf A}ttacks via {\bf STAT}istical Features (TSA-STAT)}. To address the unique challenges of time-series domain, TSA-STAT employs constraints on statistical features of the time-series data to construct adversarial examples. Optimized polynomial transformations are used to create attacks that are more effective (in terms of successfully fooling DNNs) than those based on additive perturbations. We also provide certified bounds on the norm of the statistical features for constructing adversarial examples.  Our experiments on diverse real-world benchmark datasets show the effectiveness of TSA-STAT in fooling DNNs for time-series domain and in improving their robustness. The source code of TSA-STAT algorithms is available at \href{https://github.com/tahabelkhouja/Time-Series-Attacks-via-STATistical-Features.git}{https://github.com/tahabelkhouja/Time-Series-Attacks-via-STATistical-Features}
\end{abstract}

\section{Introduction}

We are seeing a significant growth in the Internet of Things (IoT) and mobile applications which are based on predictive analytics over time-series data collected from various types of sensors and wearable devices. Some important applications include smart home automation \cite{aminikhanghahi2018real}, mobile health \cite{ignatov2018real}, smart grid management \cite{zheng2017wide}, and finance \cite{ozbayoglu2020deep}. Deep neural networks (DNNs) have shown great success in learning accurate predictive models from time-series data \cite{wang2017time}. In spite of their success, very little is known about the adversarial robustness of DNNs for time-series domain. Most of the prior work on adversarial robustness for DNNs is focused on image domain \cite{kolter2018materials} and natural language domain \cite{wang2019deep} to a lesser extent. Adversarial methods rely on small perturbations to create worst possible scenarios from a learning agent's perspective. These perturbations are constructed by bounding $l_p$-norm (with $p$=2 or $\infty$, and sometimes $p$=1) and depend heavily on the input data space: they can be a small noise to individual pixels of an image or word substitutions in a sentence. Adversarial examples expose the brittleness of DNNs and motivate methods to improve their robustness. 

Time-series domain poses unique challenges (e.g., sparse peaks, fast oscillations) that are not encountered in both image and natural language processing domains. The standard approach of imposing an $l_p$-norm bound is not applicable as it doesn't capture the true similarity between time-series instances. Consequently, $l_p$-norm constrained perturbations can potentially create adversarial examples which correspond to a completely different class label. There is no prior work on filtering methods in the signal processing literature to {\em automatically} identify such invalid adversarial candidates. Hence, adversarial examples from prior methods based on $l_p$-norm will confuse the learner when they are used to improve the robustness of DNNs via adversarial training, i.e., augmenting adversarial examples to the original training data. In other words, the accuracy of DNNs can potentially degrade on real-world time-series data after adversarial training. Indeed, our experiments corroborate this hypothesis on diverse real-world datasets.

\begin{figure}[!h]
    \centering
    \includegraphics[width=.9\linewidth]{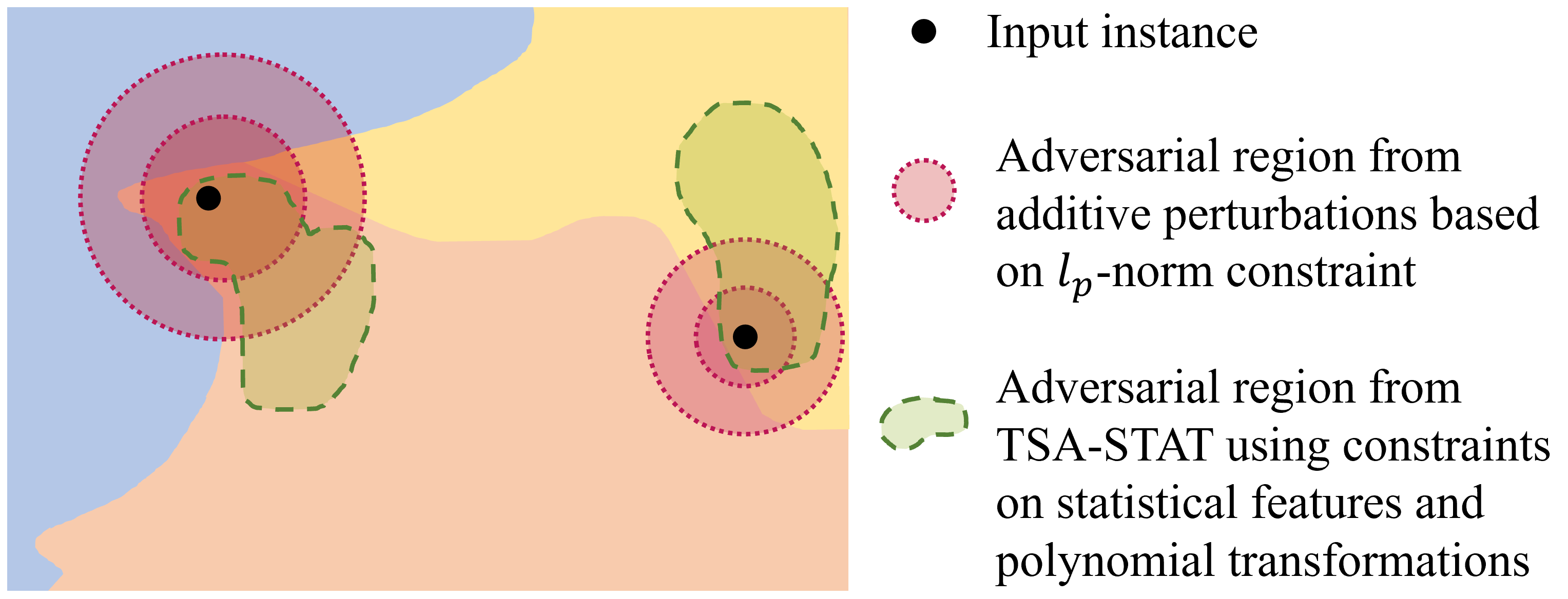}
    \caption{Conceptual illustration of adversarial regions for different attack strategies for three classes shown in blue, orange, and yellow colors. The dotted circles represent adversarial regions based on $l_p$-norm using standard additive perturbations. The green areas correspond to adversarial regions of TSA-STAT based on statistical constraints and polynomial transformations. The dotted circles cover multiple classes (invalid adversarial examples) and green areas cover only the true class label (valid adversarial examples). The intersection of green area and corresponding dotted circle represents valid additive perturbations with statistical constraints. Statistical constraints allow us to create valid adversarial examples and polynomial transformations expand the valid adversarial region.}
    \label{fig:thadvspace}
    \vspace{-3ex}
\end{figure}

In this paper, we propose a novel framework  referred to as {\em {\bf T}ime-{\bf S}eries {\bf A}ttacks via {\bf STAT}istical Features (TSA-STAT)} and provide certified bounds on robustness. TSA-STAT relies on three key ideas. First, we create adversarial examples by imposing {\em constraints on statistical features} of the clean time-series signal. This is inspired by the observation that time-series data are comprehensible using multiple statistical tools rather than the raw data \cite{ignatov2018real,christ2016distributed,ge2016feature}. 
The statistical constraints allow us to create valid adversarial examples that are much more similar to the original time-series signal when compared to $l_p$-norm constrained perturbations as demonstrated in Section \ref{sec:minimFGS}. Second, we employ {\em polynomial transformations} to create adversarial examples. For a given polynomial transformation with fixed parameters and an input time-series signal, we get an adversarial time-series as the output. We theoretically prove that polynomial transformations expand the space of valid adversarial examples over traditional additive perturbations, i.e., identify blind spots of additive perturbations. Our experiments demonstrate that polynomial transformation based attacks are more effective (in terms of successfully fooling time-series DNNs) than those based on additive perturbations. Third, to create attacks of different types, we solve an appropriate optimization problem to {\em identify the parameters of the polynomial transformation} via gradient descent. Certifiable robustness studies DNN classifiers whose prediction for any input $X$ is verifiably constant within some neighborhood around $X$, e.g., $l_p$ ball. We derive a certified bound for robustness of adversarial attacks using TSA-STAT. Our TSA-STAT framework and certification guarantees are applicable to DNNs for time-series domain with different network structures.

Figure \ref{fig:thadvspace} provides a conceptual illustration that captures the intuition behind TSA-STAT to create more effective and valid adversarial examples over $l_p$-norm constrained attacks: statistical constraints allow us to create valid adversarial examples and polynomial transformations extend the space of valid adversarial examples. Our experiments demonstrate the practical benefits of extending the space of valid adversarial examples over those from prior $l_p$-norm based methods. One potential advantage of the overall approach is the transferability of the attack to different input instances and deep models, which we evaluate in our experiments. We employ TSA-STAT to create a variety of adversarial attacks (single-instance and universal) under both white-box and black-box settings. We demonstrate that the above three ideas collectively overcome the limitations of prior work in the image domain to create effective adversarial examples to meet the unique needs of the time-series domain.  Experimental results on diverse real-world time-series datasets show that the TSA-STAT framework creates more effective adversarial attacks to fool DNNs when compared to prior adversarial methods.

\vspace{1.0ex}

\noindent {\bf Contributions.} The key contribution of this paper is the development, theoretical analysis, and experimental evaluation of the TSA-STAT framework. Specific contributions include:

\begin{itemize}
\setlength\itemsep{0em}
    \item Development of a principled approach to create targeted adversarial examples for the time-series domain using statistical constraints and polynomial transformations. Theoretical analysis to prove that polynomial transformations expand the space of valid adversarial examples over additive perturbations.
    \item Derivation of a certified bound for adversarial robustness of TSA-STAT that is applicable to any deep model for time-series domain.
    \item Comprehensive experimental evaluation of TSA-STAT on diverse real-world benchmark datasets and comparison with state-of-the-art baselines. The source code of TSA-STAT algorithms is available at \href{https://github.com/tahabelkhouja/Time-Series-Attacks-via-STATistical-Features.git}{https://github.com/tahabelkhouja/Time-Series-Attacks-via-STATistical-Features}
\end{itemize}

\section{Problem Setup}

\begin{table}[!h]
\centering
\caption{Mathematical notations used in this paper.}
\vspace{1ex}
\begin{tabular}{|l|l|}  
\hline
\textbf{VARIABLE}  & \textbf{DEFINITION} \\ \hline

$F_{\theta}$ & DNN classifier with parameters $\theta$ \\  \hline
$\mathbb{R}^{n \times T}$ &  Time-series input space, where $n$ is the number of channels \\ &and $T$ is the window-size\\  \hline
$Y$ & Set of output class labels\\  \hline
$\mathcal{PT}$ & Polynomial transformation on the input space $\mathbb{R}^{n \times T}$ \\  \hline
$a_k$ & Coefficient in $\mathbb{R}^{n\times T}$ of the polynomial transformation defined \\ &in Section \ref{sec:framework} \\  \hline
$S_i(X)$ & A statistical feature of time-series input $X$ \\  \hline
$\mathcal{S}^m(X)$ & A set of $m$ statistical features of time-series $X$ \\  \hline
$y_{target}$ & The class-label in $Y$ which an attack intends for DNN classifier $F_{\theta}$ \\ &to predict \\  \hline
$\mathcal{N}(\cdot, \cdot)$ & Multivariate Gaussian distribution\\ \hline
$\delta$  & Certified bound for a given time-series input $X$ \\&and DNN classifier $F_{\theta}$ \\    \hline
\end{tabular}
\label{tab:notation}
\end{table}

Let $X\in \mathbb{R}^{n \times T}$ be a multi-variate time-series signal, where $n$ is the number of channels and $T$ is the window-size of the signal. For this input space, we consider a DNN classifier $F_{\theta}: \mathbb{R}^{n \times T} \rightarrow Y$, where $\theta$ stands for weights/parameters and $Y$ is the set of candidate (classification) labels. For example, in a health monitoring application using physiological sensors for patients diagnosed with cardiac arrhythmia, we use the measurements from wearable devices to predict the likelihood of a cardiac failure.

\vspace{1.0ex}

$X_{adv}$ is called an adversarial example of input $X$ if:
\begin{equation*}
\bigg\{X_{adv} ~\bigg/~ \|X_{adv}-X\|_p \le \epsilon \text{~and~} F_{\theta}(X) \neq F_{\theta}(X_{adv})\bigg\}
\end{equation*}

\noindent where $\epsilon$ defines the neighborhood of highly-similar examples for input $X$ to create worst-possible outcomes from the learning agent's perspective and $\|.\|_p$ stands for $l_p$ norm. Given a DNN classifier $F_\theta$ and time-series signal $X$ with class label $y$, our goal is to create a valid adversarial example $X_{adv}$ which belongs to the semantic space of the true class label $y$. Table \ref{tab:notation} summarizes the different mathematical notations used in this paper.

\vspace {1.0ex}

\noindent {\bf Challenges for time-series domain.} The standard $l_p$-norm based distance doesn't capture the unique characteristics (e.g., fast-pace oscillations, sharp peaks) and the appropriate notion of invariance for time-series signals. As a consequence, perturbations based on $l_p$-norm can lead to a time-series signal that semantically belongs to a different class-label as illustrated in Figure \ref{fig:thadvspace}.  Indeed, our experiments demonstrate that small perturbations result in adversarial examples whose distance ($l_2$ and $l_{\infty}$-norm) from the original time-series signal is greater than the distance between time-series signals from two different class labels (see Section \ref{sec:minimFGS}). Therefore, there is a great need for studying adversarial methods focused on deep models for the time-series domain by exploiting the structure and unique characteristics of time-series signals. {\em The goal of this paper is to precisely fill this gap in our knowledge.}

\section{Related Work}

\vspace{1.0ex}

\noindent {\bf Adversarial methods.} Prior work for creating adversarial examples mostly focus on image and natural language processing (NLP) domains \cite{kolter2018materials,wang2019deep}. For the image domain, such methods include general attacks such as Carlini \& Wagner (CW) attack \cite{carlini2017towards} and universal attacks \cite{moosavi2017universal}. CW is an instance-specific attack that relies on solving an optimization problem to create adversarial examples by controlling the adversarial confidence score to fool the target deep model. Universal attacks are a class of adversarial methods that are not input-dependent. The goal of universal attacks is to create a universal perturbation that can be added to any input to create a corresponding adversarial example. The Frank-Wolfe attack \cite{jinghui20frank} improves the optimization strategy for adversarial examples to overcome the limitations of  projection methods.

Recent work regularizes adversarial example generation methods to obey intrinsic properties of images. The work of \cite{laidlaw2019functional} enforces a smoothness regularizer on the adversarial output such that similar-color pixels are perturbed following the same direction. Other works have employed spatial transformation within a perceptual threshold \cite{xiao2018spatially} or a semantic-preserving transformation \cite{hosseini2017limitation} to regularize the output. These methods exploit the intrinsic characteristics of images to control and regularize the algorithm to create adversarial examples. 
Expectation Over Transformation (EOT) \cite{Anish18synthesizing} approach creates robust adversarial examples that are effective over an entire distribution of transformations by maximizing an expectation of the log-likelihood given transformed inputs. These transformations include perceptual distortion of a given image such as rotation or texture modification. RayS method \cite{chen2020rays} was also proposed to improve the search over adversarial examples using a sanity check that is specific for the image domain.

\cite{Baluja18learning} proposed to use Adversarial Transformation Network (ATN) to automatically create adversarial examples for any given input. The role of polynomial transformation function in our TSA-STAT framework is similar to ATN. However, polynomial transformation functions are simpler and does not require hyper-parameter tuning. \cite{karim2020adversarial} investigated the use of ATNs for time-series data. The main findings include ATN fails to find adversarial examples for many inputs and not all targeted attacks are successful to fool DNNs.

While adversarial attacks perturb pixel values in the image domain, they perturb characters and words in the NLP domain. For example, adversarial attacks may change some characters to obtain an adversarial text which seems similar to the reader, or change the sentence structure to obtain an adversarial text which is semantically similar to the original input sentence (e.g., paraphrasing). One method to fool text classifiers is to employ the saliency map of input words to generate adversarial examples while preserving meaning under the white-box setting \cite{samanta2017towards}. A second method named DeepWordBug \cite{gao2018black} employs a black-box strategy to fool classifiers with simple character-level transformations. Since characteristics of time-series (e.g., fast-pace oscillations, sharp peaks) are different from images and text, most transformations in both domains are not applicable to time-series data. As a consequence, prior methods are not suitable for the time-series domain. Our proposed TSA-STAT method employs constraints on statistical features of time-series and polynomial transformations to create effective adversarial examples for time-series domain.

\vspace{1.0ex}

\noindent{\bf Certified robustness.} Early studies of adversarial robustness relied on \textit{empirical defenses}. The most successful empirical defense known so far is adversarial training \cite{tramer2020adaptive} that employs adversarial algorithms to augment training data. This method is intuitive as it relies on feeding DNNs with adversarial examples in order to be robust against adversarial attacks. Other defense methods  have been designed to overcome the injection of adversarial examples and the failure of deep models. \cite{athalye2018obfuscated} proposed different attack techniques to show that a defense method such as obfuscated gradients is unable to create a robust deep model. Distillation technique \cite{papernot2016distillation} has also been proposed as a defense against adversarial perturbations. It was shown empirically that such techniques can reduce the success rate of adversarial example generation. \cite{kurakin2018ensemble} analyzed adversarial training and its transferability property to explain how robust deep models should be attained. To improve adversarial training through a min-max optimization formulation, \cite{xiong2020improved} tries to learn a recurrent neural network to guide the optimizer to solve the inner maximization problem of the min-max training objective. However, such defense methods either offer specialized solutions or unquantifiable improvement in robustness for a given adversarial attack strategy. Importantly, for time-series domain, as $l_p$-norm based perturbations may not guarantee preserving the semantics of the true class label, adversarial examples may mislead the deep model during the adversarial training phase.

\par To improve over empirical defenses, the concept of \textit{certifiable robustness} was introduced. A deep model is certifiably robust for a given input $X$, if the prediction of $X$ is guaranteed to be constant within a small neighborhood of $X$, e.g., $l_p$ ball. \cite{raghunathan2018certified} provide certificates for one-hidden-layer neural networks using semi-definite relaxation. In \cite{hein2017formal}, certification is an instance-specific lower bound on the tampering required to change the classifier's decision with a small loss in accuracy. In a recent work \cite{cohen2019certified,li2019certified}, the robustness of deep models against adversarial perturbation is connected to random noise. These methods certify adversarial perturbations for deep models under the $l_2$ norm. \cite{cohen2019certified} defined two families for certification methods: 1) \textit{Exact} methods report the existence or the absence of a possible adversarial perturbation within a given bound. This goal has been achieved using feed-forward multi-layer neural networks based on Satisfiability Modulo Theory  \cite{huang2017safety}  or modeling the neural network as a 0-1 Mixed Integer Linear Program \cite{fischetti2018deep}. However, these methods suffer from scalability challenges. 2) \textit{Conservative} methods either confirm that a given network is robust for a given bound or report that robustness is inconclusive \cite{li2019certified}. Our proposed robustness certificate for TSA-STAT falls in the conservative category and extends the recent method based on random noise \cite{li2019certified}.

\vspace{1.0ex}

\noindent {\bf Adversarial attacks for time-series domain.} There is little to no principled prior work on adversarial methods for time-series domain. \cite{fawaz2019adversarial} employed the standard Fast Gradient Sign method with $l_2$-norm bound \cite{kurakin2016adversarial} to create  adversarial noise with the goal of reducing the confidence of deep convolutional models for classifying {\em uni-variate} signals. Network distillation is also employed to train a student model for creating adversarial attacks \cite{karim2020adversarial}. In an orthogonal work, the study from \cite{siddiqui2019tsviz} concluded that time-series signals are highly-complex, and their interpretability is ambiguous. Additionally, there is no previous work on certification algorithm for time-series domain. Prior methods can only certify the deep models using $\l_p$-norm and are not specific to time-series domain.
Adversarial examples can also be studied for regression tasks over time-series data. However, there is very limited work in this direction as explained by \cite{siddiqui2019tsviz}. These methods consider Euclidean distance and employ standard methods from the image domain such as FGSM \cite{mode2020adversarial}. In an orthogonal/complementary direction, generative adversarial networks are used to impute missing values in time-series data \cite{Yonghong19e2gan,Yonghong18imputation}. 

In summary, existing methods for the time-series domain are lacking in the following ways: 1) Do not create targeted adversarial attacks; 2) Employ $l_p$-norm based perturbations\footnote{A concurrent work \cite{AAAI2022} developed min-max optimization methods to explicitly train robust deep models for time-series domain based on the global alignment kernel measure.}, which do not take into account the unique characteristics of time-series data; and 3) Do not provide theoretical guarantees for adversarial robustness. This paper overcomes these drawbacks and improves the state-of-the-art in adversarial robustness for time-series domain through the proposed TSA-STAT framework.

\section{The TSA-STAT Framework}
\label{sec:framework}
In this section, we first provide a high-level overview of the TSA-STAT framework. Subsequently, we describe the key elements, and instantiate the framework to create white-box and black-box attacks.

\begin{figure*}[!h]
    \centering
    \includegraphics[width=\linewidth]{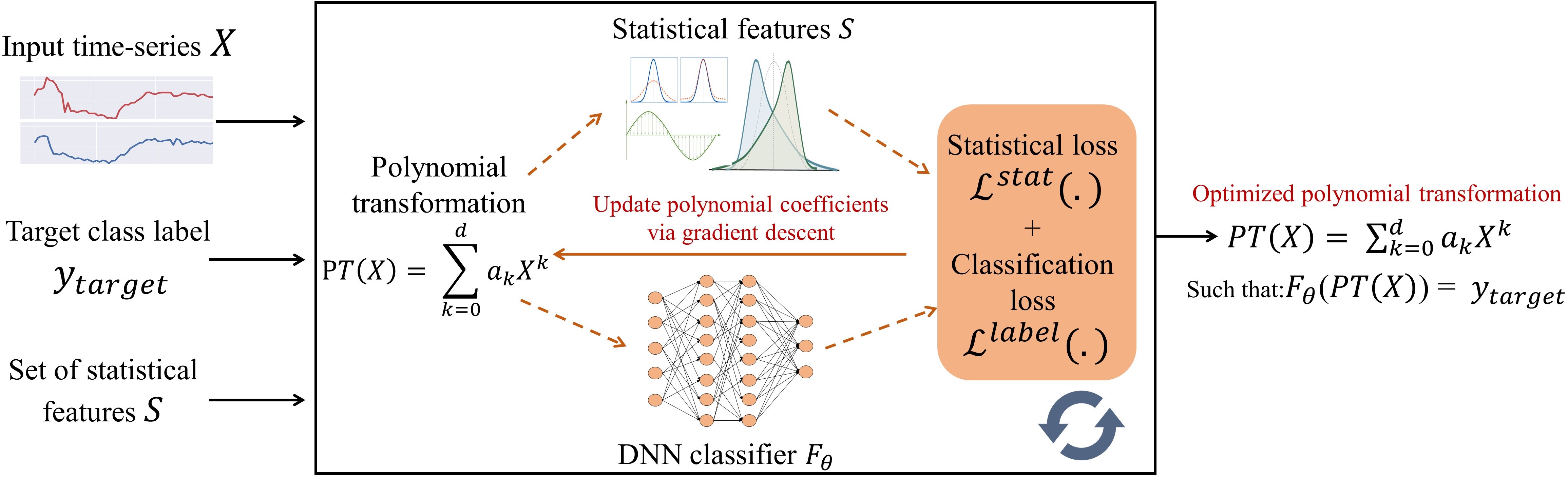}
    \caption{High-level overview of the TSA-STAT framework to create adversarial examples using optimized polynomial transformations. Given an input time-series signal $X$, a target label $y_{target}$, a DNN classifier $F_\theta$, and a set of statistical features $\mathcal{S}$, TSA-STAT solves an optimization problem over two different losses to find the parameters of the polynomial transformation: 1) A statistical loss to ensure that original time-series signal $X$ and the generated adversarial example $X_{adv}$ are highly similar by imposing constraints on their statistical features; and 2) A classification loss to make sure that the DNN classifier $F_\theta$ classifies the generated adversarial example $X_{adv}$ with the target class label $y_{target}$. The optimized polynomial transformation will take the time-series signal $X$ as input and produce adversarial example $X_{adv}$ as output.} 
    \label{fig:highlevel}
    \vspace{-4ex}
\end{figure*}

\noindent {\bf Overview of TSA-STAT.} Our framework creates targeted adversarial examples using polynomial transformations. For a given input time-series signal $X$, a target label $y_{target}$, a set of statistical features $\mathcal{S}$ and a DNN classifier $F_\theta$, TSA-STAT generates adversarial examples using two key ideas: 1) {\em Constraints on the statistical features} to regularize the similarity of adversarial example $X_{adv}$ to the original time-series $X$; and 2) A {\em polynomial transformation} that allows us to explore a larger space of adversarial examples over the traditional additive perturbations. Figure \ref{fig:highlevel} provides a high-level overview of the TSA-STAT framework. The effectiveness of adversarial examples critically depends on the coefficients of the polynomial transformation. TSA-STAT solves an optimization problem over two different losses via gradient descent to find the parameters of the polynomial transformation. First, a statistical loss is employed to ensure that original time-series signal $X$ and the generated adversarial example $X_{adv}$ are highly similar by imposing constraints on their statistical features. Second, a classification loss to make sure that the DNN classifier $F_\theta$ classifies the generated adversarial example $X_{adv}$ with the target class label $y_{target}$. The polynomial transformation with the optimized parameters will take the time-series signal $X$ as input and produce adversarial example $X_{adv}$ as output.


\subsection{Key Elements}

\vspace{1.0ex}

\noindent \textbf{1) Statistical constraints.} Time-series data is often analyzed using diverse statistical tools \cite{silvey2017statistical}. Machine learning models have achieved good classification performance using statistical features of time-series data \cite{fulcher2014highly}. These prior studies motivate us to use statistical features of time-series data to develop adversarial algorithms. We propose a new definition to create adversarial examples for time-series signals. Let $\mathcal{S}^m(X)=\{S_1(X), S_2(X), \cdots, S_m(X)\}$ be the set of statistical features of a given input $X$ (e.g., mean, standard deviation, kurtosis). We define an adversarial example $X_{adv}$ derived from $X$ as follows:
\begin{equation}
\begin{cases}
\forall ~ 1 \le i\le m,~  \|S_i(X_{adv})-S_i(X)\|_{\infty} \le \epsilon_i \\
\text{and} ~~ F_{\theta}(X) \neq F_{\theta}(X_{adv})
\end{cases}
\label{eq:advdef}
\end{equation}

where $\epsilon_i$ is the bound for the $i^{th}$ statistical feature. Using this definition, we call to change the conventional $l_p$ distance-based neighborhood-similarity to one based on statistical features for creating valid adversarial examples. We conjecture that this definition is better suited for adversarial examples in time-series domain. Indeed, our experiments strongly support this claim.

\vspace{1.0ex}

\noindent \textbf{2) Polynomial transformation-based attacks.} To explore larger and powerful space of valid adversarial examples when compared to traditional additive perturbations, we propose polynomial transformation based attacks. The aim of this approach is to find a transformation over the input space that creates effective adversarial attacks. This transformation considers the entire time-series input to decide the output for each channel and time-step of the adversarial example. Hence, we propose an adversarial transformation on the input time-series space. We define polynomial transformation $\mathcal{PT}: \mathbb{R}^{n \times T} \rightarrow \mathbb{R}^{n \times T}$ as follows:
\begin{equation}
    X_{adv} = \mathcal{PT}(X) = \mathcal{PT}(X_{i,j}) ~~\forall (i,j) \in [n] \times [T]
\end{equation}
where $X\in \mathbb{R}^{n \times T}$ is the input time-series signal and $X_{adv}$ is the corresponding adversarial example. The key idea is to create a threat model that does not require calling back the deep model for every new adversarial attack. Our goal is to preserve dependencies between features of the input space by having a transformation $\mathcal{PT}(\cdot)$ that depends on the input time-series $X$, unlike the standard additive perturbations. Inspired by power series \cite{drensky2006constants}, we approximate this transformation $\mathcal{PT}(\cdot)$ using a polynomial representation with a chosen degree $d$: $\mathcal{PT}(X)=\sum_{k=0}^{d} a_k~X^k + \mathcal{O}(X^{d+1})$, where $a_k \in \mathbb{R}^{n\times T}$ denote the polynomial coefficients and $\mathcal{O}$ stands for Big O notation.

\begin{theorem}
\label{th:poly}
For a given input space $\mathbb{R}^{n\times T}$ and $d \ge 1$, polynomial transformations allow more candidate adversarial examples than additive perturbations in a constrained space. If $X\in \mathbb{R}^{n\times T}$ and $\mathcal{PT}:X\rightarrow \sum_{k=0}^{d} a_k~X^k$, then $\forall X_{adv}$ s.t. $\|S_i(X_{adv})-S_i(X)\|_{\infty} \le \epsilon_i$:
\begin{equation*}
\bigg\{X_{adv}=\mathcal{PT}(X), ~\forall a_k  \bigg\} 
\supsetneq
\bigg\{X_{adv}=X+\delta, ~\forall \delta  \bigg\}
\end{equation*}
, $S_i \in  \mathcal{S}^m(X)\bigcup I$dentity.
\end{theorem}

The above theorem states that polynomial transformations expand the space of valid adversarial examples and identify blind spots of additive perturbations. In other words, the theorem explains that {\em some} of the adversarial examples created using polynomial transformations are not possible using standard additive perturbations. We show through the proof provided in {\bf Appendix \ref{append:proof}} that an example created using a standard additive perturbation can be created by a polynomial transformation, however, the inverse is not always true. This theorem motivates the use polynomial transformations within the TSA-STAT framework instead of additive perturbations in order to uncover more adversarial examples.

\vspace{1.0ex}

\noindent \textbf{3) Optimization based adversarial attacks.} To create powerful adversarial examples to fool the deep model $F_{\theta}(X)$, we need to find optimized coefficients $a_k$, $\forall ~k$=0 to $d$, of the polynomial transformation $\mathcal{PT}(X)$. Our approach is based on minimizing a {\em loss function} $\mathcal{L}$ using gradient descent that {\bf a)} Enforces an input signal $X$ to be mis-classified to a target class $y_{target}$ (different from true class label $y^* \in Y$); and {\bf b)} Preserves close proximity to statistical features in the given set $\mathcal{S}^m$. 

\vspace{1.0ex}

{\bf Classification loss.} To achieve the mis-classification goal, we employ the formulation of \cite{carlini2017towards} to define a loss function:
\begin{equation}
    \mathcal{L}^{label}(\{a_k\}, X) =  \max \Bigg[  \max_{y \neq y_{target}} \left( \mathcal{Z}_y\left(\sum_{k=0}^{d} a_k~X^k\right) \right) - \mathcal{Z}_{y_{target}}\left( \sum_{k=0}^{d} a_k~X^k\right)\textbf{,} ~~\rho \Bigg]
    \label{eq:classloss}
\end{equation}

\noindent where $\rho < 0$. This loss function will ensure that the adversarial example will be moving towards the space where it will be classified by the DNN as class $y_{target}$ with a confidence $|\rho|$ using the output of the pre-softmax layer $\{\mathcal{Z}_y\}_{y \in Y}$. 
\vspace{1.0ex}

{\bf Statistical loss.} To satisfy the constraints on statistical features of the set $\mathcal{S}^m$, we propose another loss function. This loss function overcomes the impractical use of projection functions on the statistical feature space. 

\begin{equation}
    \mathcal{L}^{stat}(\{a_k\}, X, \mathcal{S}^m) \triangleq  \sum_{S_i \in \mathcal{S}^m} \|S_i(\sum_{k=0}^{d} a_k~X^k) - S_i(X)\|_{\infty}
    \label{eq:statloss}
\end{equation}

\vspace{1.0ex}

{\bf Combined loss.} The final loss function $\mathcal{L}$ that we want to minimize  to obtain coefficients $a_k$ of the polynomial transformation $\mathcal{PT}(\cdot)$ is as follows:
\begin{equation}
    \mathcal{L}(\{a_k\}, X, \mathcal{S}^m) = \beta_{label} \times \mathcal{L}^{label}(\{a_k\}, X) +  \beta_{stat} \times \mathcal{L}^{stat}(\{a_k\}, X, \mathcal{S}^m)
    \tag{$\bigstar$}
    \label{eq:floss}
\end{equation}
where $\beta_{label}$ and $\beta_{stat}$ are hyper-parameters that can be used to change the trade-off between the adversarial classification loss $\mathcal{L}^{label}$ and the statistical loss $\mathcal{L}^{stat}$. We note that our experiments showed good results with the simple configuration of $\beta_{label}$=1 and $\beta_{stat}$=1.

\subsection{Instantiations of TSA-STAT}

\vspace{1.0ex}

\noindent \textbf{White-box setting.} Our goal is to create targeted adversarial attacks on a classifier $F_{\theta}$. Adversarial transformation $X_{adv}$ for a single-instance $X$: 
$X_{adv} =\mathcal{PT}_{y_{target}}(X)=\sum_{k=0}^{d} a_k~X^k$ s.t.:
\begin{align*}
    \begin{cases}
    \|S_i(X_{adv})-S_i(X)\|_{\infty} \le \epsilon_i~\forall S_i \in \mathcal{S}^m \\ F_{\theta}(X_{adv})=y_{target}
    \end{cases}
\end{align*}
where  $y_{target}$ is the target class-label of the attack. 

\vspace{1.0ex}

We employ gradient descent based optimizer to minimize the loss function in Equation \ref{eq:floss} over $\{a_k\}_{0\le k \le d}$, where $d$ is the polynomial degree for $\mathcal{PT}(\cdot)$. The parameter $\rho$ introduced in Equation \ref{eq:classloss} plays an important role here. $\rho$ will push gradient descent to minimize mainly the second term when the first one plateaus at $\rho$ first. Otherwise, the gradient can minimize the general loss function by pushing $\mathcal{L}^{label}(\{a_k\}, X)$ to $-\infty$, which is counter-productive for our goal.

\vspace{1.0ex}

\noindent We can also extend this procedure to create adversarial examples under universal perturbations. A universal perturbation generates a single transformation that is applicable for any input $X \in \mathbb{R}^{n\times T}$. We introduce a targeted universal attack in this setting as:
\vspace{-2ex}
\begin{equation}
    X_{adv}  =\mathcal{PT}_{y_{target}}(X) =\sum_{k=0}^{d} a_k~X^k  ~\text{s.t.}~ \mathcal{PT}(F(X_{adv})=y_{target})>(1-e_t)
    \vspace{-2ex}
\end{equation}

where $e_t$ represents the error probability of creating an adversarial example that $F_{\theta}$ would classify it with label $y \neq y_{target}$. Our proposed algorithm analyzes a given set of inputs to find coefficients $\{a_k\}_{0\le k \le d}$ that would push image of multiple inputs  $\mathcal{TF}(X)=\sum_{k=0}^{d} a_k~X^k$ to the decision boundary of a target class-label $y_{target}$ defined by the classifier $F_{\theta}$. 
As the algorithms for both universal attack and instance-specific attack are similar and follow the same general steps, we present the universal attack algorithm of TSA-STAT in Algorithm \ref{alg:univ}. The instance-specific attack is a {\em special-case} of the universal attack: Since the universal algorithm generates a single polynomial transformation that is applicable for {\em any} time-series $X$, the instance-specific transformation is just applicable for a {\em single} time-series $X$. Algorithm \ref{alg:univ} can degenerate to the case of instance-specific attack by changing the value of $l$ (the number of time-series inputs) in Line 2 to the value of 1 and optimize over only one time-series $X$.

\vspace{1.0ex}

\noindent \textbf{Black-box setting.}  Black-box attacks are adversarial examples that are created with no knowledge about the target deep model parameters $\theta$. In the best scenario, the attacker has the ability to query the target model to get the predicted label for any input time-series $X$. This allows the creation of a proxy deep model to mimic the behavior of the target model. This technique can be more effective when a target scenario is well-defined \cite{tramer2020adaptive,papernot2017practical}. 
For the instantiation of TSA-STAT, we consider the general case where we do not query the black-box target DL. We create adversarial examples using optimized transformations as in white-box setting and prove through experimental results that the same transformations generalize to fool other black-box deep learning models. 

\begin{algorithm}[!h]
\caption{Optimized universal adversarial transformation}
\label{alg:univ}
\textbf{Input}: A set of $l$ inputs $\{X_i\}_{i=1}^{l}$; $d$, maximum degree; $y_{target}$, target class; $F_{\theta}$, target model; $\mathcal{S}^m$, statistical feature set; $\eta$, learning rate\\
\textbf{Output}: $\{a^y_k\}_{0\le k\le d,~y \in Y}$
\begin{algorithmic}[1] 
\STATE Random initialization of $\{a_k^y\}$.
\FOR{i=1 to $l$}
\IF {$F_{\theta}(X_i) \neq y_{target}$}
\STATE $\hat{y} \leftarrow F_{\theta}(X_i)$
\STATE $\delta \leftarrow \nabla_{\{a^{\hat{y}}_k\}}  \mathfrak{L}(\{a^{\hat{y}}_k\}, X_i, \mathcal{S}^m) ~~ \forall k$
\STATE $\forall k: \{a^{\hat{y}}_k\}\leftarrow \{a^{\hat{y}}_k\} - \eta \times \delta$ 
\ENDIF
\ENDFOR
\STATE \textbf{return} $\{a^y_k\}_{0\le k\le d,~y \in Y}$
\end{algorithmic}
\end{algorithm}

\section{Certified Bounds for Adversarial Robustness of TSA-STAT}
\begin{figure}[!h]
    \centering
    \begin{minipage}{\linewidth}
    \centering
    \hspace{5em}
        \begin{minipage}{.25\linewidth}
                \centering
                \includegraphics[width=\linewidth]{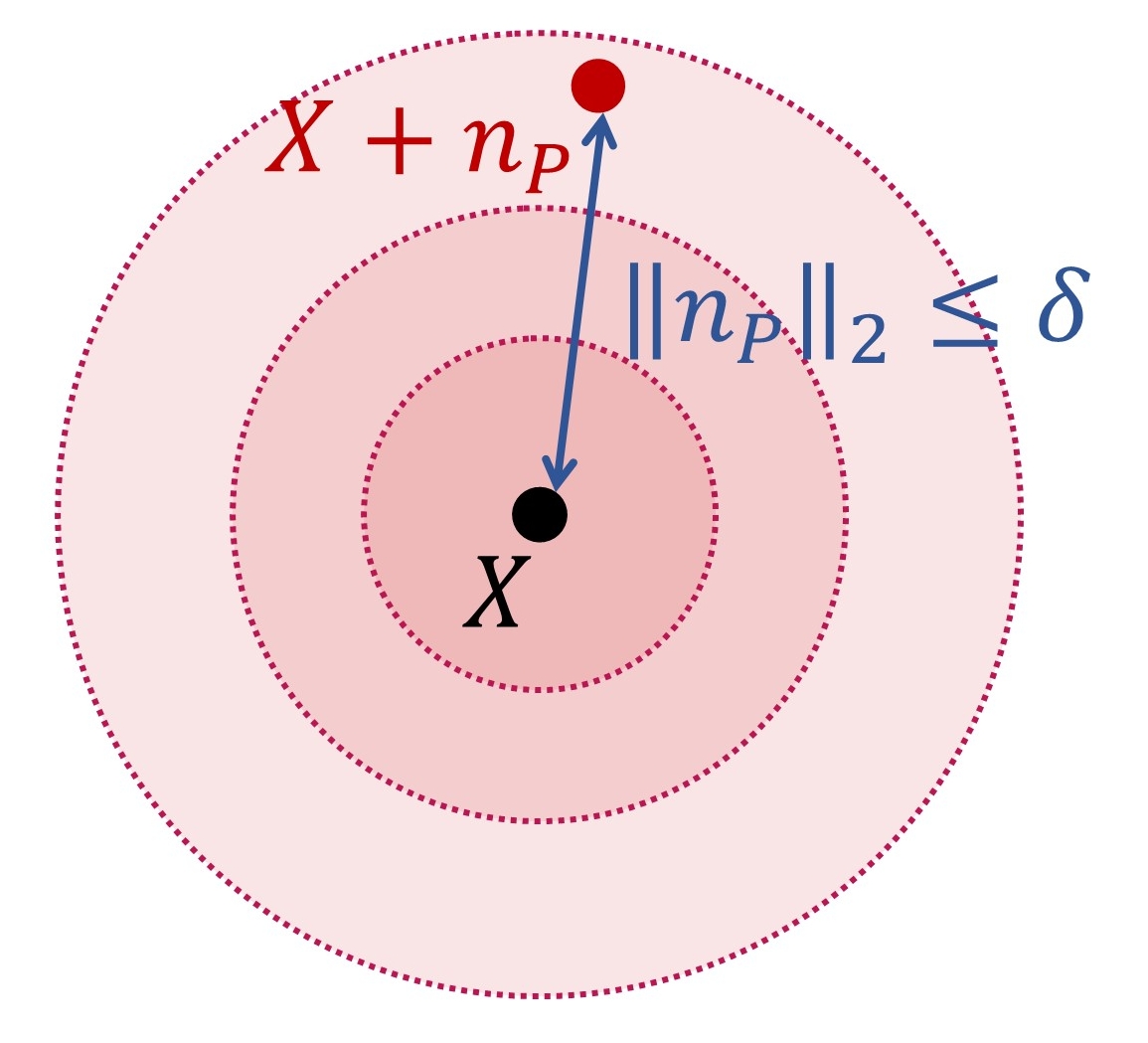}
            \end{minipage}%
            \hspace{4em}
        \begin{minipage}{.4\linewidth}
                \centering
                \includegraphics[width=\linewidth]{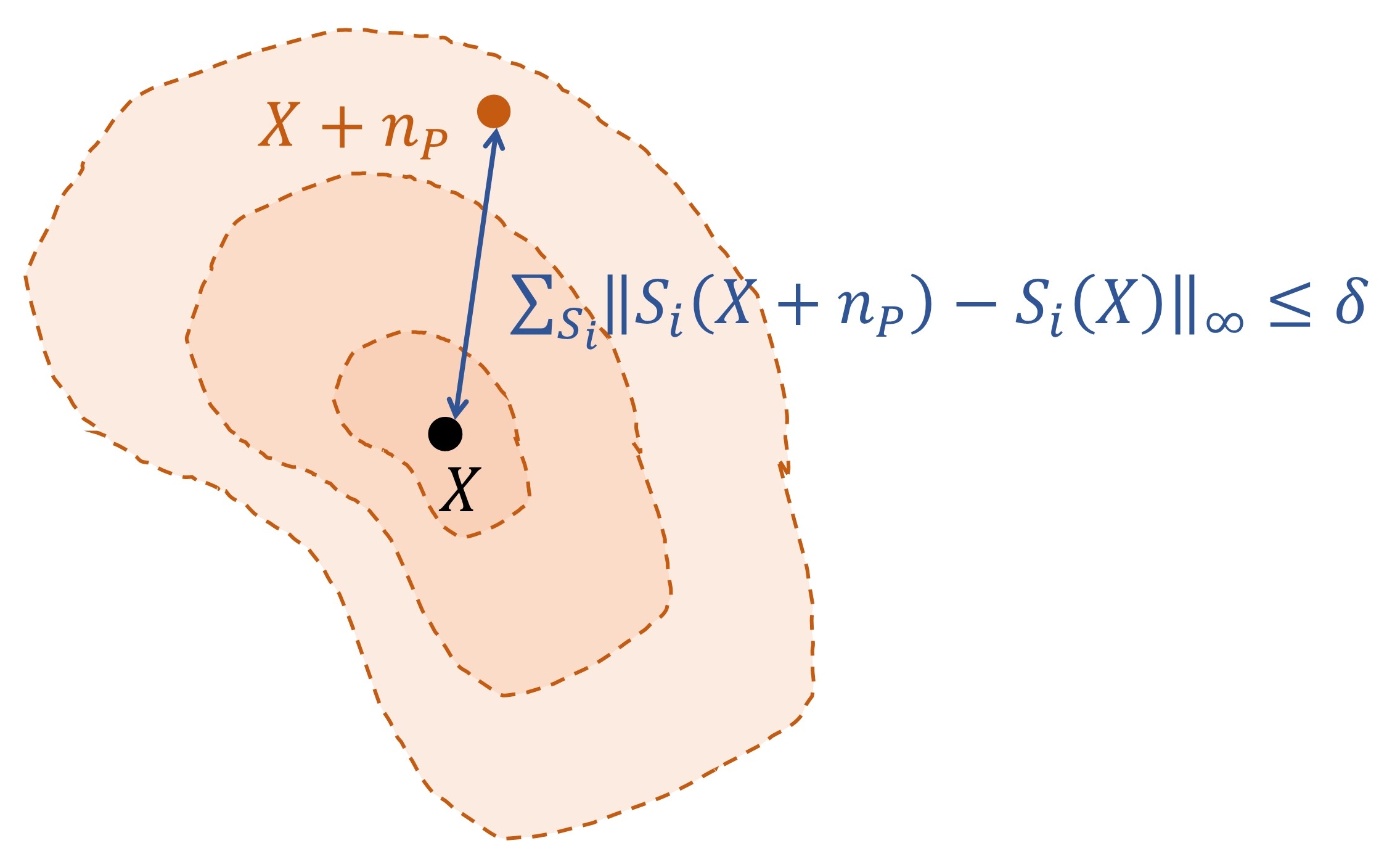}
            \end{minipage}\\
    \hspace{5em}
        \begin{minipage}{.25\linewidth}
                \centering
                (a)
            \end{minipage}%
            \hspace{4em}
        \begin{minipage}{.4\linewidth}
                \centering
                (b)
            \end{minipage}
    \end{minipage}
    \caption{Conceptual illustration of the perturbation region of an input $X$ with respect to noise $n_P$ as considered by (a) Standard $l_2$ norm where $\delta$ is a constant representing the Euclidean distance between $X$ and $X+n_P$; and (b) Statistical constraints as considered by TSA-STAT, where $\delta$ is a constant representing the cumulative sum of the maximum difference between statistical features computed over $X$ and $X+n_P$ for each time-series channel separately. $S_i$ represents one statistical feature (e.g., mean), and $S_i(X)$ and $S_i(X+n_P)$ represent the vector of values for a given statistical feature, one for each time-series channel (e.g., a vector of mean values for each channel).}
    \label{fig:deltaspaces}
\end{figure}

\vspace{1.0ex}
In this section, we propose a novel certification approach for adversarial robustness of the TSA-STAT framework. Given a time-series input $X\in \mathbb{R}^{n\times T}$ and a classifier $F_{\theta}$, our overall goal is to provide a certification bound $\delta$ on the $\|\cdot\|_{\infty}$ over the statistical features $\mathcal{S}^m(X)$ of the time-series signal $X$. Traditionally, the certification bound is a constant $\delta$ that constrains the distance between an input $X$ and a perturbed version $X_{adv}$=$X+n_P$ ($n_P$ is a multi-variate noise) as shown in Figure \ref{fig:deltaspaces}(a). Using TSA-STAT, our goal is to derive a certification bound $\delta$ that constraints the  difference between the statistical features of a given time-series input $X$ and a perturbed version $X+n_P$ as shown in Figure \ref{fig:deltaspaces}(b). This bound will guarantee the robustness of classifier $F_{\theta}$ in predicting $F_{\theta}(X_{adv})=F_{\theta}(X)$ for any adversarial time-series $X_{adv}$ such that $\sum_{S_i \in \mathcal{S}^m} \|S_i(X_{adv}) - S_i(X)\|_{\infty}\le \delta$, where $S_i$ is a statistical feature (e.g., a vector of mean values, one for each time-series channel) and ${\infty}$ norm takes the maximum of the difference between statistical feature values for each channel separately (e.g., maximum of the difference between mean for each channel separately).

As explained in Section 3, there are two families for certification methods, namely, exact and conservative. It has been shown that exact certification approaches do not scale well with the network in question \cite{cohen2019certified}. Hence, we propose a certification algorithm that belongs to the conservative family. Our aim is to provide a bound that asserts that the prediction of $F_{\theta}(X)$ remains unchanged for any adversarial instance $X_{adv}$ such that $\|S_i(X_{adv})-S_i(X)\|_{\infty}$ is bounded by $\delta$. State-of-the-art methods such as Gaussian smoothing \cite{cohen2019certified} rely on the Euclidean distance to measure the similarity between the original input and its adversarial example. Since TSA-STAT investigates statistical features of time-series for similarity purposes, the $l_2$ bounds derived by prior work are not sufficient to cover time-series adversarial examples. The certification provided in prior work cannot be extended to assess the robustness of DNNs for time-series domain as TSA-STAT relies on complex statistical features. To overcome this challenge, we propose a new robustness certification approach for TSA-STAT that is well-suited for time-series domain by bounding the statistical features. 
\begin{figure}[!h]
    \centering
    \includegraphics[width=\linewidth]{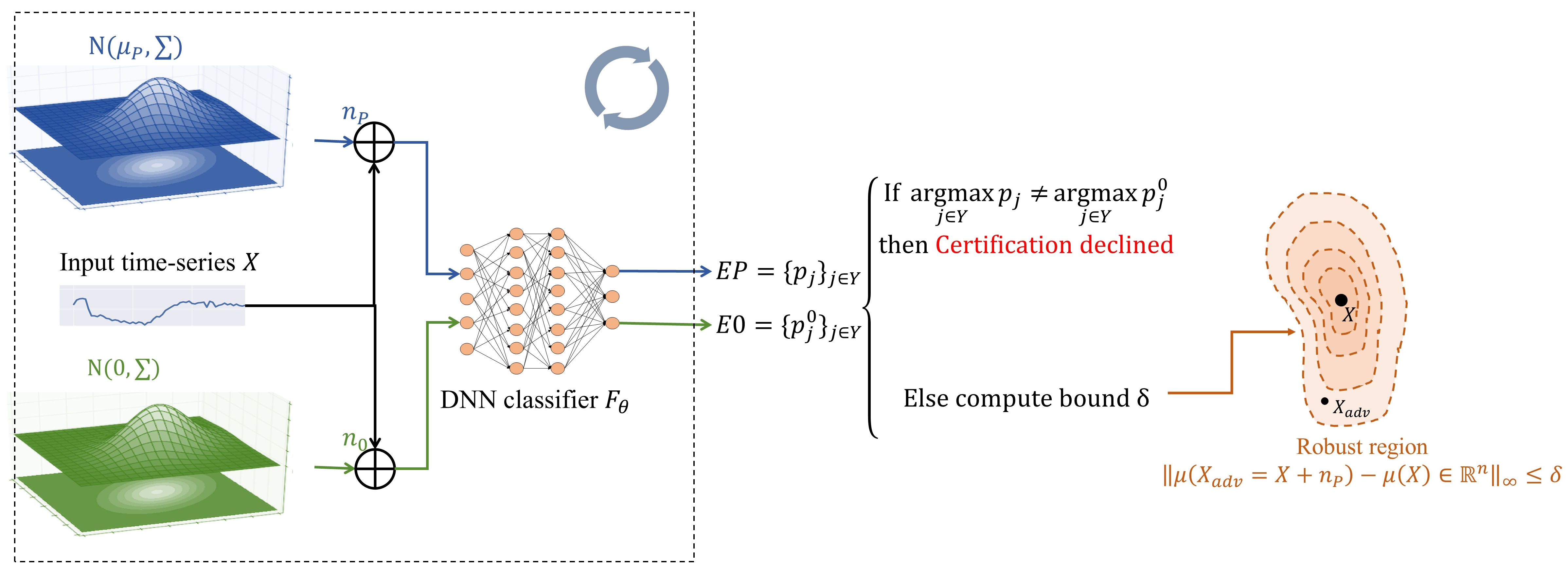}
    \caption{High-level illustration of the TSA-STAT certification approach to estimate the statistical perturbation space of a given time-series input $X$ where the classifier $F_{\theta}$ is robust. This illustration is for mean only. For a given number of iterations, we repeatedly generate perturbations $n_P$ and $n_0$ and add them to the time-series $X$ to assess the robustness of classifier $F_{\theta}$ using the mean statistical feature. $n_P\sim \mathcal{N}(\mu_P, \cdot)$ is generated to mimic the perturbation that can affect the input time-series signal by producing $EP$ (probability distribution over candidate class labels) that is used to characterize the robustness of classifier $F_{\theta}$ for predicting the same label for time-series $X$. $n_0 \sim \mathcal{N}(0, \cdot)$ is generated as an arbitrary noise that does not affect the mean vector (statistical feature for each channel) of $X$ and produces $E0$ (probability distribution over candidate class labels) needed for the computation of the certification bound $\delta$. Once $\delta$ is estimated, TSA-STAT guarantees the  robustness of classifier $F_{\theta}$ for predicting  $F_{\theta}(X_{adv})$=$F_{\theta}(X)$ for any $X_{adv}$ such that $\|\mu(X_{adv}) - \mu(X)\|_{\infty}\le \delta$, where $\mu(.)$ is the vector of mean values, one for each time-series channel separately and ${\infty}$ norm takes the maximum value of a given vector.}
    \label{fig:certifAlg}
\end{figure}
Intuitively, we aim to provide a certification for an input $X$ that considers the statistical feature space of the time-series signal $X$. TSA-STAT's certification relies on adding random multi-variate perturbation $n_P$ to quantify the robustness of the classifier on the surrounding region using statistical constraints as shown in Figure \ref{fig:certifAlg}.

\begin{enumerate}
    \item If the classifier $F_{\theta}$ predicts a class-label on the perturbation $X_{adv}$ which differs from the prediction on the original time-series signal $X$, then the classifier is prone to adversarial attacks on the input.
    \item If the classifier $F_{\theta}$ yields the correct classification in spite of all the perturbations, then it is easy to say that the classifier is robust against any perturbation (represented by $n_P$) on the time-series input $X$.
    \item If the classifier $F_{\theta}$ yields the correct classification on most perturbation cases, then we develop an algorithmic approach to compute the conditions that $n_P$ must satisfy in order to not affect the classifier's prediction. Therefore, the certification bound can be deduced.
\end{enumerate}

\begin{algorithm}[!t]
\caption{TSA-STAT Certification Algorithm}
\label{alg:certif}
\textbf{Input}: A multivariate time-series signal $X$, $F_{\theta}$, DNN classifier; $Y$, the set of class labels\\
\textbf{Parameters}: $\mu_P$, multivariate mean; $\sum$, covariance matrix; $n$, the number of iterations\\
\textbf{Output}: $\hat{y}$, predicted class label; $\delta$, certification bound
\begin{algorithmic}[1] 

\FOR{$i$=1 to $MAX$}
\STATE Generate $n_P \sim \mathcal{N}(\mu_P, \sum)$ 
 and $n_0 \sim \mathcal{N}(0, \sum)$ 
\STATE Compute $\hat{y}^{p}(i)$=$F_{\theta}(X+n_P)$ and $\hat{y}^{0}(i)$=$F_{\theta}(X+n_0)$
\ENDFOR
\STATE Estimate EP=$\{p_j=\frac{\sum_{i=1}^{MAX}\mathbb{I}[[\hat{y}^{p}(i)==j]]}{MAX}\}_{j\in Y}$
\STATE Estimate E0=$\{p^0_j=\frac{\sum_{i=1}^{MAX} \mathbb{I}[[\hat{y}^0(i) == j]]}{MAX}\}_{j\in Y}$

\IF{$\displaystyle arg\max_{j \in Y} \; p_j \neq \displaystyle arg\max_{j \in Y} \; p^0_j$}
\STATE \textbf{return} \; \texttt{Certification declined} 
\ELSIF{$\displaystyle \max_{j\in Y}\; p_j$ equals 1}
\STATE \textbf{return} predicted label $\hat{y}=\displaystyle arg\max_{j \in Y} \; p_j$ and certification bound $\delta = \|\mu_P\|_{\infty}$ 
\ELSE
\STATE Compute the upper bound:

\begin{equation*}
    \delta^2=\displaystyle \max_{\alpha \neq 1} \frac{2}{\alpha \cdot \sum\nolimits^{(S)}}  \cdot \Bigg(  -ln\bigg(1-p_{(1)}-p_{(2)}  + 2\left( \frac{1}{2}\left( p_{(1)}^{1-\alpha} + p_{(2)}^{1-\alpha} \right)  \right)^{\frac{1}{1-\alpha}} \bigg) \Bigg)
\end{equation*}
\STATE \textbf{return} predicted label $\hat{y}$=$\displaystyle arg\max_{j\in Y} \; p_j$ and certification bound $\delta$ 
\ENDIF
\end{algorithmic}
\end{algorithm}

Our certification study relies on Rényi Divergence \cite{van2014renyi}. Rényi divergence is a generalization of the well-known Kullback-Leibler (KL) divergence \cite{van2014renyi}. For a positive order $\alpha\neq 1 $ and two probability distributions $EP$=$(p_1,\cdots, p_k)$ and $E0$=$(p^0_1,\cdots, p^0_k)$, which are estimated in our case, the Rényi divergence is defined as:

\begin{equation}
    D_\alpha(EP\|E0) = \frac{1}{\alpha-1}~ln\left(\sum_{i=1}^{k}p_i^\alpha \cdot (p^0_i)^{1-\alpha}\right)
\end{equation}

For the purpose of this paper, we define the estimated probability distribution $EP$ as the empirical probabilities $p_i$ that the class $i$ is predicted by $F_{\theta}$ on $X+n_P$ ($n_P$ being a random perturbation). Our TSA-STAT framework is general to handle multivariate time-series data. Hence, we define a multi-variate Gaussian distribution $\mathcal{N}(\mu, \sum)$ characterized by a mean vector $\mu$ and a covariance matrix $\sum$ to generate $n_P$.
To compute the divergence between multi-variate Gaussian distributions, the expression is provided by the following Lemma \cite{gil2013renyi}.
\begin{lemma}
\label{lemma:mvgdist}
For two multivariate Gaussian distributions $\mathcal{N}(\mu_1, \sum_1)$ ~and~ $\mathcal{N}(\mu_2, \sum_2)$:

\begin{equation*}
    D_\alpha(\mathcal{N}(\mu_1, \sum_1)\|\mathcal{N}(\mu_2, \sum_2)) = \frac{\alpha}{2}(\mu_1-\mu_2)^T\sum_{\alpha}(\mu_1-\mu_2) - \frac{1}{2(\alpha-1)}~ ln\frac{|\sum_{\alpha}|}{|\sum_1|^{1-\alpha}|\sum_2|^{\alpha}}
\end{equation*}

, where $\sum_{\alpha}=\alpha\sum_1+(1-\alpha)\sum_2$.
\end{lemma}

\noindent Lemma \ref{lemma:mvgdist} provides the expression of the divergence using the parameters of the multivariate Gaussian distributions. Consequently, we use it to produce the following theorem to provide certification bound over the mean of the time-series input space for adversarial robustness of TSA-STAT. For this purpose, we require a second multivariate Gaussian distribution $\mathcal{N}(\cdot, \cdot)$ to estimate $E0$ and to compute the divergence provided in Lemma 1. Therefore, we use an arbitrary distribution $\mathcal{N}(0, \sum)$ with a zero-vector mean. This way, the mean feature of the input time-series signal will not be disturbed. For a computationally-efficient derivation of the certification bound, we use the same covariance matirx $\sum$ as the multi-variate Gaussian distribution that generated $n_P$.

\begin{theorem}
\label{th:bound}
Let $X\in \mathbb{R}^{n\times T}$ be an input time-series signal. Let $n_P\sim\mathcal{N}(\mu_P\in\mathbb{R}^n, \sum)$ and $n_0 \sim\mathcal{N}(0, \sum)$. Given a classifier $F_{\theta}: \mathbb{R}^{n\times T}\rightarrow Y$ that produces a probability distribution $(p_1,\cdots,p_k)$ over $k$ labels for $F_{\theta}(X+n_P)$ and another probability distribution $(p^0_1,\cdots,p^0_k)$ for $F_{\theta}(X+n_0)$. To guarantee that $\displaystyle arg\max_{p_i} ~ p_i = \displaystyle arg\max_{p^0_i } ~ p^0_i$, the following condition must be satisfied:

\begin{equation*}
\|\mu_P\|^2_{\infty} \le \max_{\alpha \neq 1} \frac{2}{\alpha \cdot \sum\nolimits^{(S)}}  \cdot \left( -ln \left(1-p_{(1)}-p_{(2)} +2\left( \frac{1}{2}\left( p_{(1)}^{1-\alpha} + p_{(2)}^{1-\alpha} \right)  \right)^{\frac{1}{1-\alpha}} \right)\right)
\end{equation*}

where $\|\mu_P\|_{\infty}$ is the maximum perturbation over the mean of each time-series channel and $\sum\nolimits^{(S)}$ is the sum of all elements of ~$\sum$. 
\end{theorem}

\noindent This new certification formulation is suitable for the time-series domain, as it takes into account the different channels of the time-series signal input and  adversarial attacks using TSA-STAT explore a larger space of valid adversarial examples using statistical constraints and polynomial transformations. In the {\bf Appendix \ref{append:proof}}, we provide a discussion of the unique contributions of this Theorem compared to the certification method of \cite{li2019certified}.

\vspace{0.75ex}

To derive the certification bound for a given time-series signal $X\in \mathbb{R}^{n\times T}$ and a classifier $F_{\theta}$, we employ two different noise distributions to generate two different noise samples that we denote $n_P \in \mathbb{R}^{n \times T}$ and $n_0 \in \mathbb{R}^{n \times T}$, where $n$ is the number of channels and $T$ is the window size of the time-series signal. $n_P\sim \mathcal{N}(\mu_P, \cdot)$ is generated to mimic the perturbation that can affect the input time-series signal by producing $EP$ (probability distribution over candidate class labels) that is used to characterize the robustness of classifier $F_{\theta}$ for predicting the same label for time-series $X$. $n_0 \sim \mathcal{N}(0, \cdot)$ is generated as an arbitrary noise that does not affect the mean vector (statistical feature for each channel) of $X$ and produces $E0$ (probability distribution over candidate class labels) needed for the computation of the certification bound. If both perturbations result to the same classifier prediction, we compute the tolerable perturbation's upper bound $\delta=\max \|\mu_P\|_{\infty}$. As $\|\mu_P\|_{\infty}$ is upper-bounded by the RHS term of Theorem 2, the maximum value for $\|\mu_P\|_{\infty}$ is the RHS term.

The upper bound $\delta$ guarantees that for any noise $n_P$ with a mean feature for each channel constrained by $\delta$, the classifier's prediction is robust on the perturbed input $X+n_P$. In other words, following the formulation used in Equation 4.4, if for an adversarial time-series signal $X_{adv}$ such that $\|S_i(X_{adv}) - S_i(X)\|_{\infty} \le \delta$ where $S_i$ is the statistical feature mean (a vector of mean values, one for each channel) and ${\infty}$ norm takes the maximum of the difference between mean values for each channel separately, then $F_{\theta}(X_{adv})=F_{\theta}(X)$. We provide Algorithm \ref{alg:certif} to automatically assess the robustness of a classifier $F_{\theta}$ on a given multivariate time-series signal $X$ as input. To generalize this result for other statistical features, we provide Lemma \ref{lem:otherbounds}. Both proofs are present in {\bf Appendix \ref{append:proof}}.

\begin{lemma}
 If a certified bound $\delta$ has been generated for the mean of input time-series signal $X\in \mathbb{R}^{n\times T}$ and classifier $F_{\theta}$, then certified bounds for other statistical/temporal features can be derived consequently.
 \label{lem:otherbounds}
\end{lemma}

\section{Experiments and Results}

In this section, we discuss the experimental evaluation of TSA-STAT along different dimensions and compare it with prior methods.

\subsection{Experimental Setup}

\vspace{1.0ex}

\noindent \textbf{Datasets.} To evaluate the proposed TSA-STAT framework, we employed diverse uni-variate and multi-variate time series benchmark datasets \cite{ucrdata,Dua_2019,kwapisz2011activity}. Complete details are provided in Table \ref{tab:dsnames}. We employ the standard training/validation/testing splits from these benchmarks. Table \ref{tab:dsnames} describes each dataset employed in our evaluation: acronym to represent the dataset, the number of classes, and the dimensions of each input time-series signal. 
\begin{table}[!h]
\centering
\caption{Description of different benchmark time-series datasets.}
\vspace{1ex}
\begin{tabular}{|c|c|c|c|}  
\hline
\textbf{NAME}  & \textbf{ACRONYM} & \textbf{CLASSES} & \textbf{INPUT SIZE ($n\times T$) } \\ \hline
Chlorine Concentration & CC & 3 & 1$\times$166 \\ \hline
Synthetic Control & SC & 6 & 1$\times$~~30 \\ \hline
Cylinder-Bell-Funnel& CBF & 3 & 1$\times$128 \\ \hline
CricketX & CX & 12 & 1$\times$300 \\ \hline
CricketY & CY & 12 & 1$\times$300 \\ \hline
CricketZ & CZ & 12 & 1$\times$300 \\ \hline
Human Activities &&&\\ 
and Postural Transitions & HAPT & 12 & 6$\times$200 \\ \hline
WISDM & WD & 6 & 3$\times$200 \\ \hline
Character Trajectories& ChT & 20 & 3$\times$182\\  \hline
\end{tabular}
\label{tab:dsnames}
\vspace{-1ex}
\end{table}

\vspace{1.0ex}

\noindent \textbf{Algorithmic setup.} We employ three different 1D-CNN architectures --$A_0$, $A_1$, and $A_2$-- to create three deep models as target DNN classifiers: $WB$ for white-box setting, and $BB1$ and $BB2~~$ for the black-box setting respectively. $WB$ is a model using $A_0$ to evaluate the adversarial attack under a white-box setting, and trained using clean training examples. $BB_1$ and $BB_2$ use the architectures $A_1$ and $A_2$ respectively to evaluate the black-box setting. The architecture information of the deep learning models are presented in Table~\ref{tab:archs}.
\begin{table}[!t]
\centering
\caption{Details of DNN architectures. C: Convolutional layers, K: kernel size, P: max-pooling kernel size, and R: rectified linear layer.}
\begin{tabular}{|l|lllllll|l|}  
 \hline
 & C & K & C & K & P & R & R\\ \hline
$A_0$   & x & x & 66 & 12 & 12 & 1024 & x \\ \hline
$A_1$   & x & x & 20 & 12 & 2 & 512 & x \\ \hline
$A_2$   & 100 & 5 & 50 & 5 & 4 & 200 & 100 \\ \hline
\end{tabular}
\label{tab:archs}
\end{table}
To further evaluate the effectiveness of attacks, we create models that are trained using augmented data from baselines attacks that are not specific to the image domain: Fast Gradient Sign method (FGS) \cite{kurakin2016adversarial} that was used by \cite{fawaz2019adversarial}, Carlini \& Wagner (CW) \cite{carlini2017towards}, and Projected Gradient Descent (PGD) \cite{madry2017towards}. Finally, we evaluate the performance of adversarial examples from TSA-STAT on two RNN models. To effectively test the transferability of adversarial transformations, we only use the knowledge of $WB$ model. We assume that the framework is unaware of all other deep models.
\label{sec:minimFGS}
For FGS and PGD algorithms, we employed a minimal perturbation factor ($\epsilon <$ 0.4) for two main reasons. First, larger perturbations significantly degrade the overall performance of adversarial training. We also want to avoid the risk of leaking label information \cite{madry2017towards}. Second, while analyzing the datasets, we found that there are time-series signals from different classes that are separated by a distance less than what an $l_p$-norm bounded perturbation engenders. Therefore, $l_p$-norm bounded attacks will create adversarial examples that are inconsistent (i.e., examples for a semantically different class label) for adversarial training. For example, in the case of CC dataset, there are time-series signals from different classes with $l_2$-distances $\le 0.3$, while FGS's average perturbation is around $0.3$ for $\epsilon=0.3$. If we employ $l_{\infty}$-distance, CW causes several signal perturbations with $l_{\infty}$-norm $\ge 1.5$ on HAPT dataset, whereas many time-series signals with different class labels have $l_{\infty}$-distances $< 1.5$. We observed similar findings in most of the other datasets.

For TSA-STAT, we use $\beta_{label}$=1 and $\beta_{stat}$=1 for the loss function in Equation $\bigstar$ in all our experiments.  TSA-STAT's attack algorithm and adversarial training have both shown good performance with this simple configuration. Therefore, we chose not to fine-tune the hyper-parameters $\beta_{label}$ and $\beta_{stat}$ to avoid additional complexity. 
We use constraints over statistical features including mean, standard deviation, kurtosis, skewness, and root mean square \cite{brockwell2016introduction} of an input time-series signal. We explain the methodology that was used to select these statistical features below.

 \subsection{Selection of Statistical Features and Degree of Polynomial Transformation}
\label{append:feats}
\begin{figure}[!t]
    \centering
    \begin{minipage}{.45\linewidth}
        \centering
        \includegraphics[width=\linewidth]{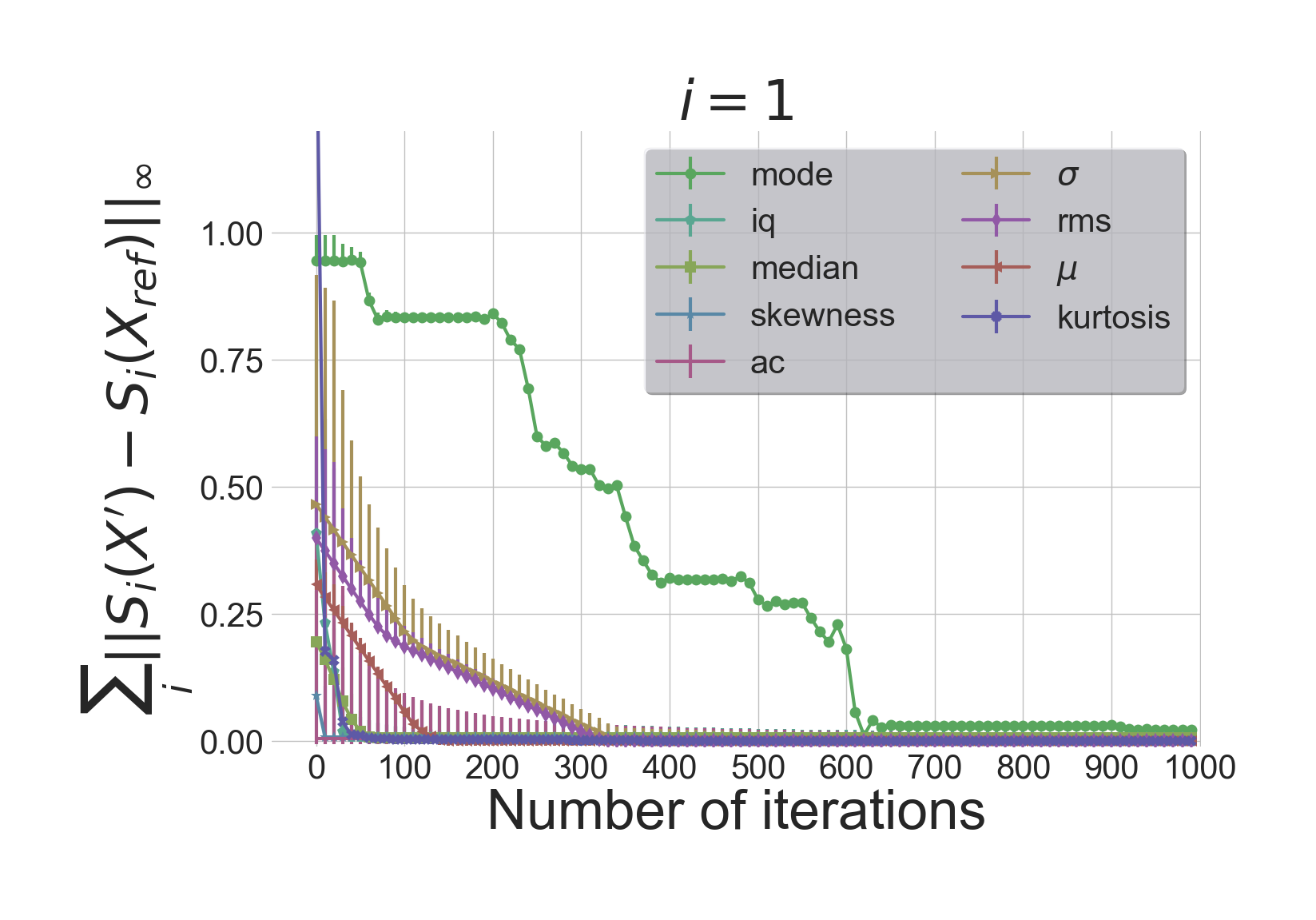}
        \vspace{-3em}
        \caption{Convergence of different\\ statistical constraints for $i\le1$}
        \label{fig:poolconv1}
    \end{minipage}%
    \begin{minipage}{.45\linewidth}
        \centering
        \includegraphics[width=\linewidth]{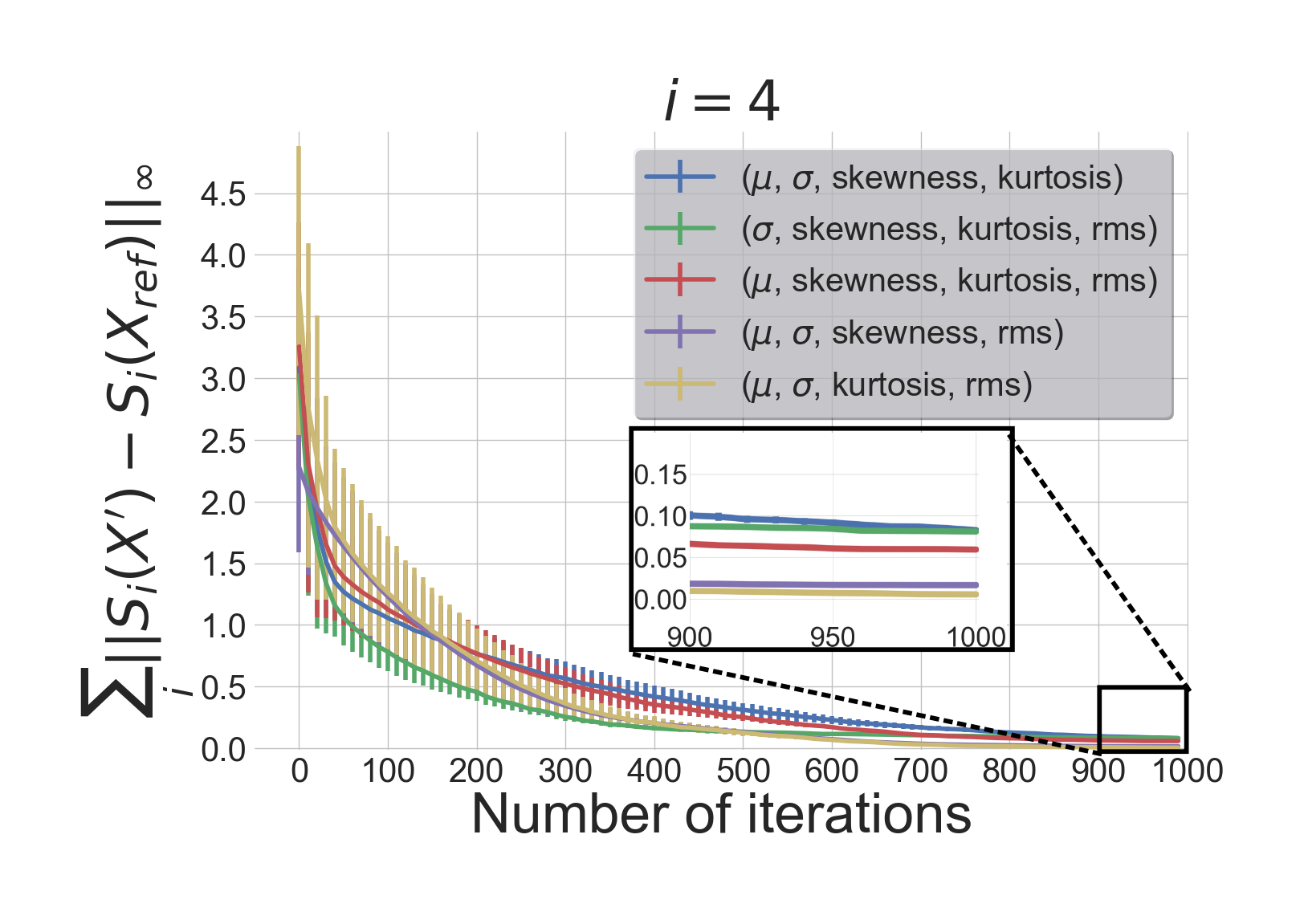}
        \vspace{-3em}
        \caption{Convergence of different\\ statistical constraints for $i\le4$}
        \label{fig:poolconv4}
    \end{minipage}
    \vspace{-2em}
\end{figure}

We initially started with the following statistical features of time-series signals: $\mathcal{S}^m$=\{Mean ($\mu$), Standard deviation ($\sigma$), Median, Mode, Interquartile range (iq), Skewness, Kurtosis, Root mean square (rms), Auto-correlation (ac)\}. To decide on the most appropriate subset to use for all our TSA-STAT experiments, we ran a convergence test on  $\sum_i \|S_i(X')-S_i(X_{ref})\|_{\infty}$ using a subset of the data from WD. We note that the TSA-STAT framework can be used with both $l_2$ norm and $l_\infty$ norm on the statistical features. For $X \in \mathbb{R}^{n \times T}$, we have one statistical feature for each channel, i.e., $S_i(X) \in \mathbb{R}^n$. Our goal from constraining $S_i(X)$ is to guarantee that for all the $n$ channels, the value of the statistical feature is less than the given bound. Hence, the use of $l_\infty$ norm is straightforward. However, any other norm can be used. To demonstrate the generality of TSA-STATE, we provide a comparison between  $l_2$ norm and $l_\infty$ norm on the statistical features in Figure \ref{fig:statnorm}. As Figure \ref{fig:poolconv1} illustrates, we ran the convergence test at first on each $S_i \in  \mathcal{S}^m$ individually ($i\le 1$). We eliminate the statistical feature $S_i$ which does not converge properly in contrast to other statistical constraints, and repeat the experiment each time by increasing $i$. Figure \ref{fig:poolconv4} illustrates the step at $i\le4$. We conclude from both Figures \ref{fig:poolconv1} and \ref{fig:poolconv4} that our approach empirically satisfies the $\epsilon_i$ bound presented in Equation \ref{eq:advdef}. Hence, for all our TSA-STAT experiments, we choose $\mathcal{S}^m$=\{Mean $\mu$, Standard deviation $\sigma$, Skewness, Root mean square\} or  $\mathcal{S}^m$=\{Mean $\mu$, Standard deviation $\sigma$, Kurtosis, Root mean square\}. We did not increase $i$ further to avoid increasing the time-complexity of the proposed algorithm for negligible benefits. We have observed similar patterns for all other datasets. We note that it is not possible to use the basic PGD method to satisfy the constraint over $\sum_i \|S_i(X')-S_i(X_{ref})\|_{\infty}$: a projection function on the statistical feature space is not a straightforward projection as in the Euclidean space. 

\begin{figure}[!h]
    \centering
    \begin{minipage}{.5\textwidth}
        \centering
        \includegraphics[width=\linewidth]{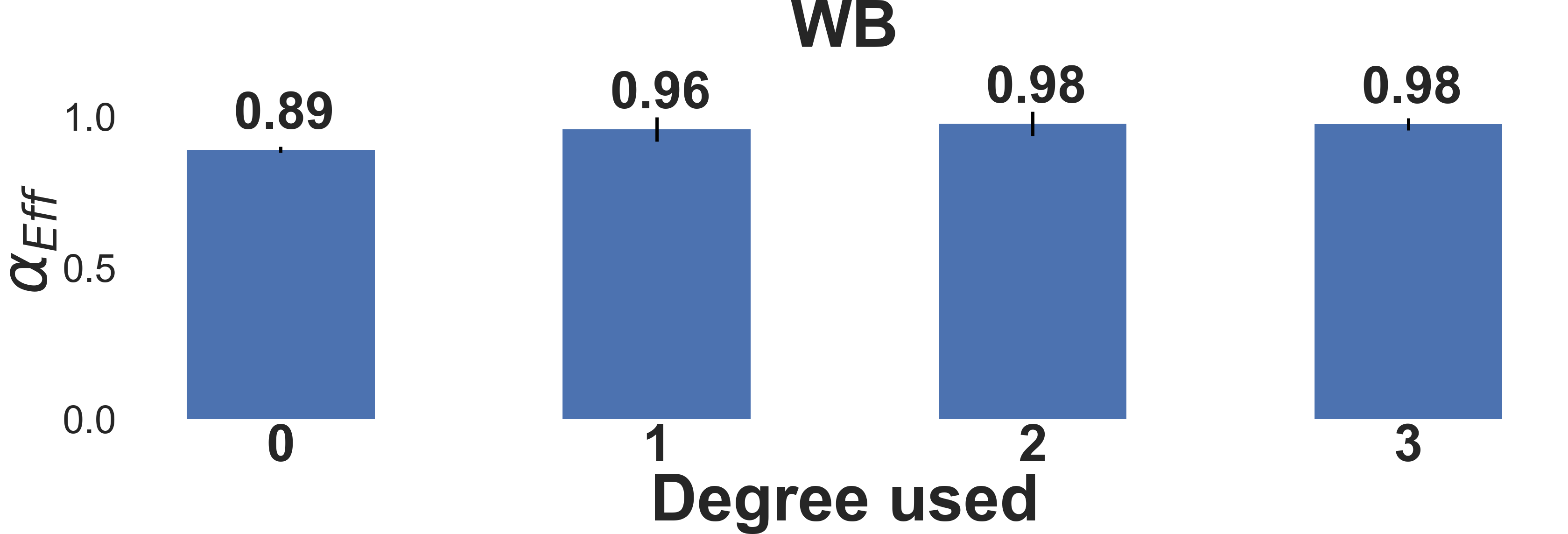}
\end{minipage}%
\begin{minipage}{.5\textwidth}
        \centering
        \includegraphics[width=\linewidth]{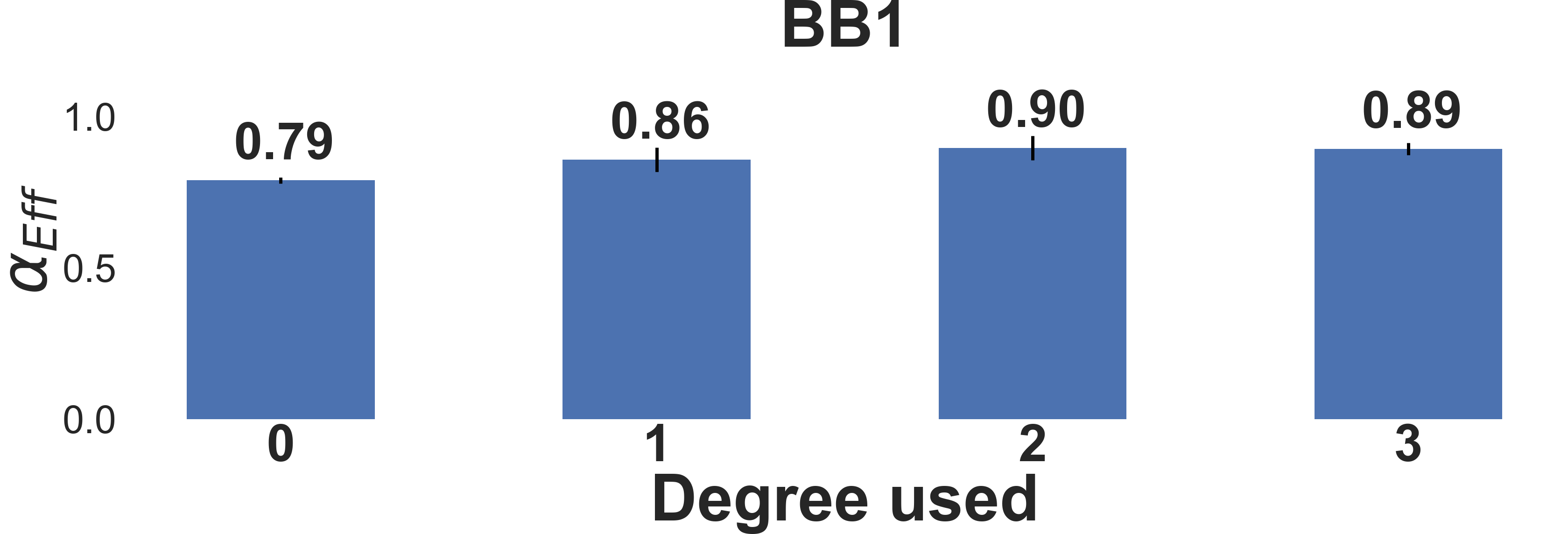}
\end{minipage}
\begin{minipage}{.5\textwidth}
        \centering
        \includegraphics[width=\linewidth]{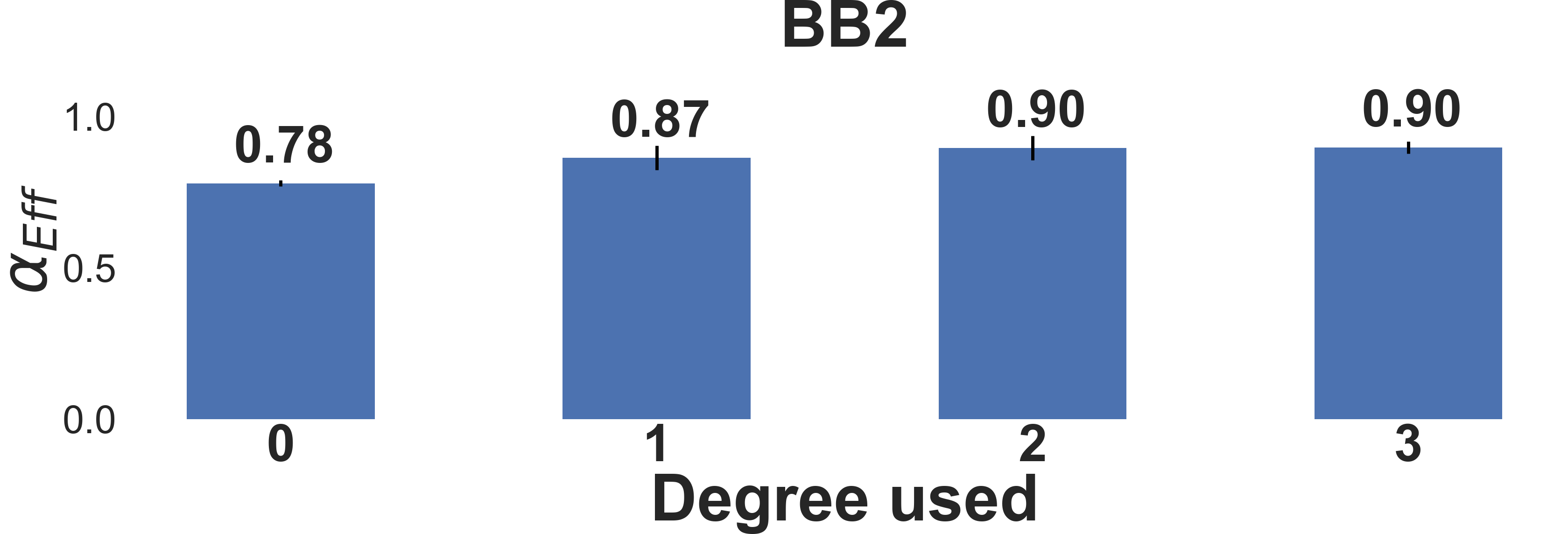}
\end{minipage}
\caption{Performance of TSA-STAT based universal adversarial attacks using polynomial transformations with different degrees on multiple DNN models.}
    \label{fig:degrees}
    \vspace{-1em}
\end{figure}

Regarding the degree of polynomial transformation for TSA-STAT, we employ $d$=1 and $d$=2 in all our experiments. Figure \ref{fig:degrees} shows the impact of different degrees used for the polynomial adversarial transformation when tested on the WD dataset noting that we observed similar patterns for all other datasets. Degree 0 corresponds to the standard constant $\delta$ additive perturbation. While the adversarial attack is still functional, degrees $\ge 1$ showed improved effectiveness of adversarial attacks. Starting from degree $3$, the attack's effectiveness did not increase significantly. To prevent increasing the time-complexity of the optimization method to find the coefficients of polynomial transformation, we chose degrees $d$=1 and $d$=2 to evaluate our TSA-STAT framework.

\subsection{Results and Discussion}

\textbf{Spatial distribution of TSA-STAT outputs.} One of the claims of this work is that adversarial examples relying on $l_p$-norm bounds are not applicable for time-series domain. To evaluate this claim, we employ a t-Distributed Stochastic Neighbor Embedding (t-SNE) \cite{maaten2008visualizing} technique to visualize the adversarial examples generated by TSA-STAT and PGD, an $l_p$-norm based attack. t-SNE provides a dimensionality reduction method that constructs a probability distribution for the high-dimensional samples to create a reduced feature space where similar instances are modeled by nearby points and dissimilar instances are modeled by distant points.
\begin{figure}[!h]
    \centering
    \begin{minipage}{\linewidth}
    \centering
        \begin{minipage}{.49\linewidth}
                \centering
                \includegraphics[width=\linewidth]{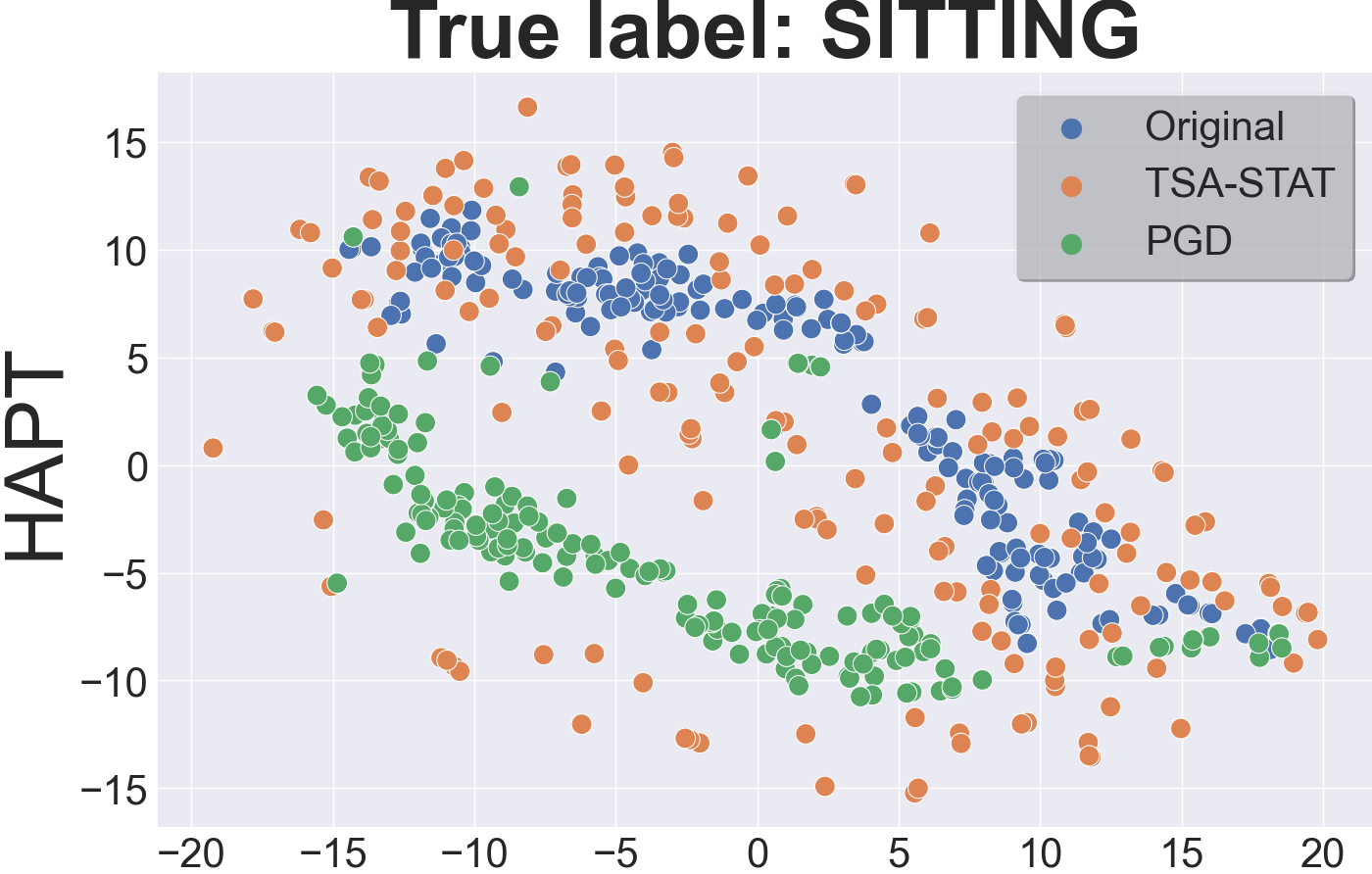}
            \end{minipage}%
        \begin{minipage}{.47\linewidth}
                \centering
                \includegraphics[width=\linewidth]{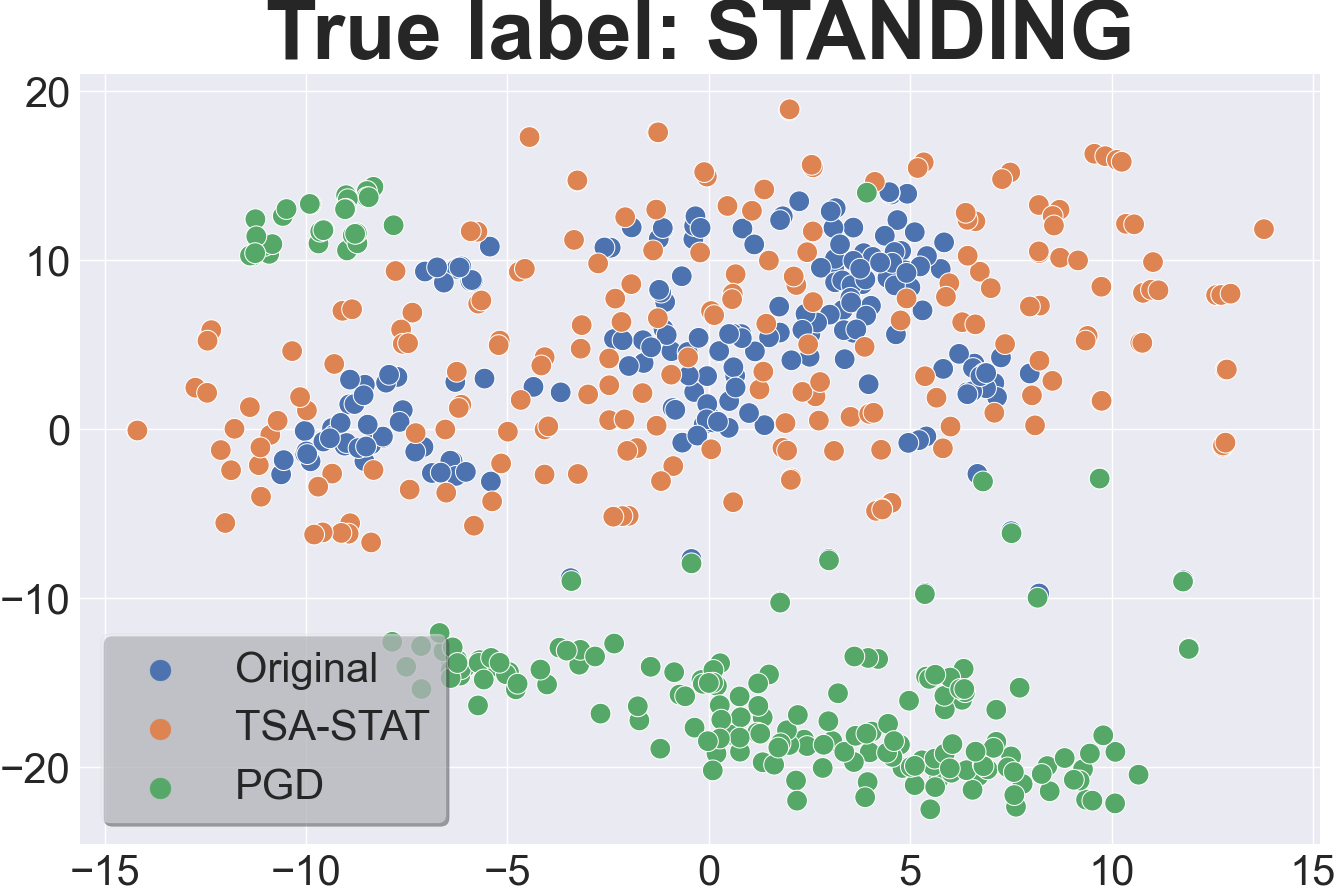}
            \end{minipage}
        \begin{minipage}{.49\linewidth}
                \centering
                \includegraphics[width=\linewidth]{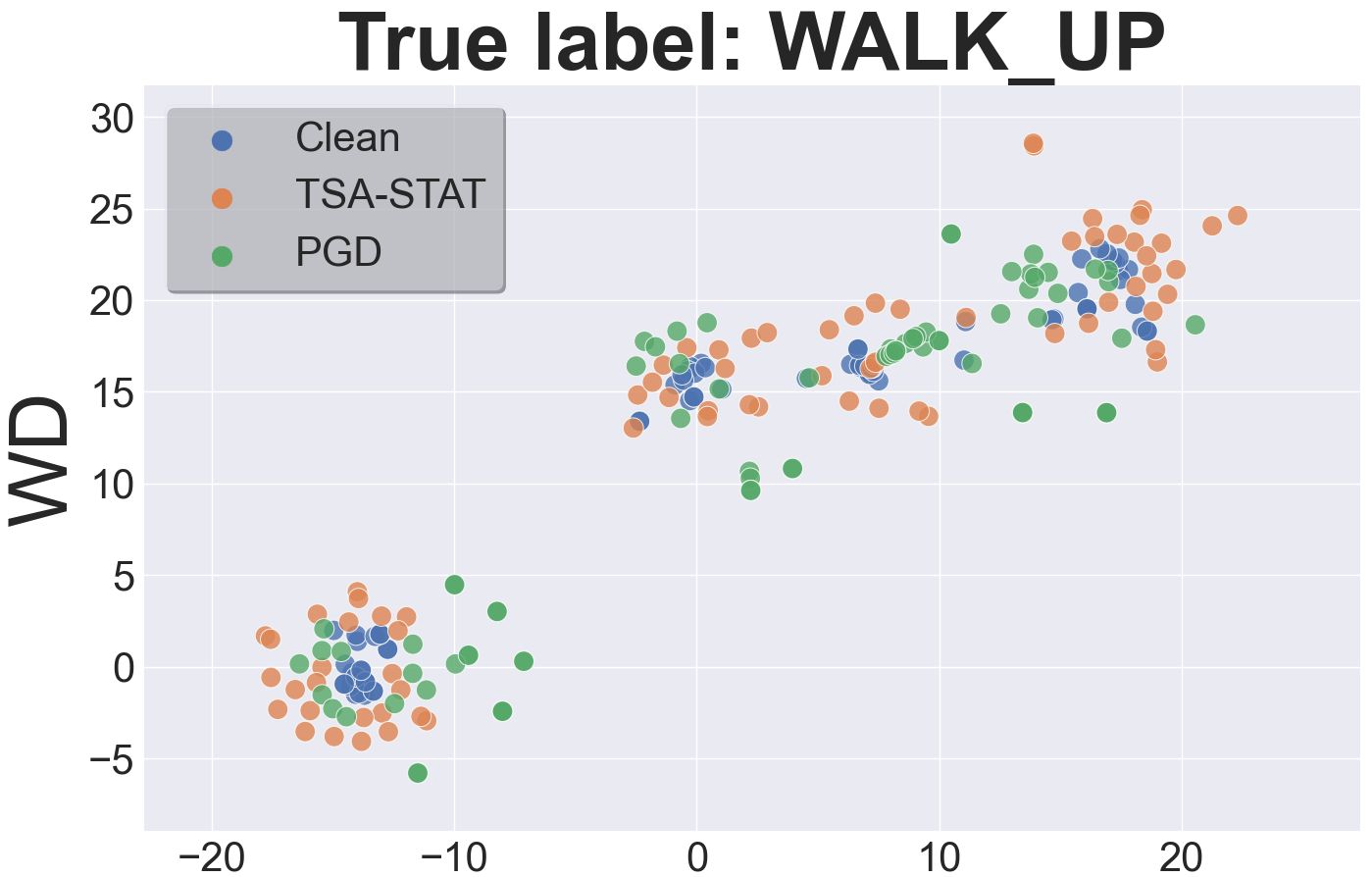}
            \end{minipage}%
            \hspace{.2em}
        \begin{minipage}{.47\linewidth}
                \centering
                \includegraphics[width=\linewidth]{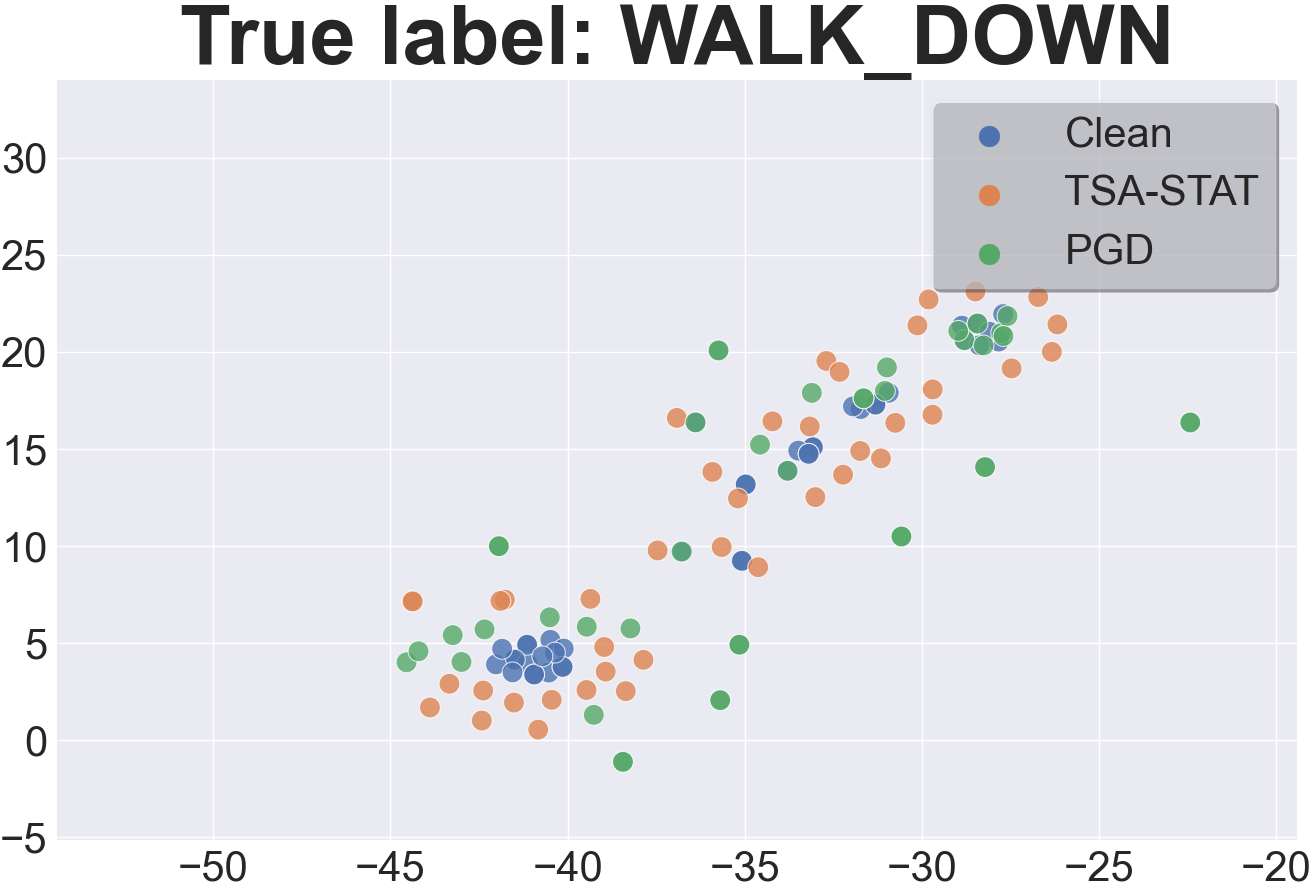}
            \end{minipage}
        \caption{t-Distributed Stochastic Neighbor Embedding showing the distribution of natural and adversarial examples from TSA-STAT and PGD. Adversarial examples from TSA-STAT are more or equally similar to the original time-series input than PGD-based adversarial examples.}
        \label{fig:tsne}
    \end{minipage}
    \vspace{-1em}
\end{figure}

Figure \ref{fig:tsne} illustrates a representative example of the spatial distribution between same-class data of HAPT and WD, and their respective adversarial examples using TSA-STAT and PGD. We can clearly see that TSA-STAT succeeds in preserving the similarity between the original and adversarial example pairs, and in most cases, better than PGD.

\vspace{1.0ex}

\begin{figure}[!h]
    \centering
    \begin{minipage}{.45\textwidth}
        \centering
        \includegraphics[width=.8\linewidth]{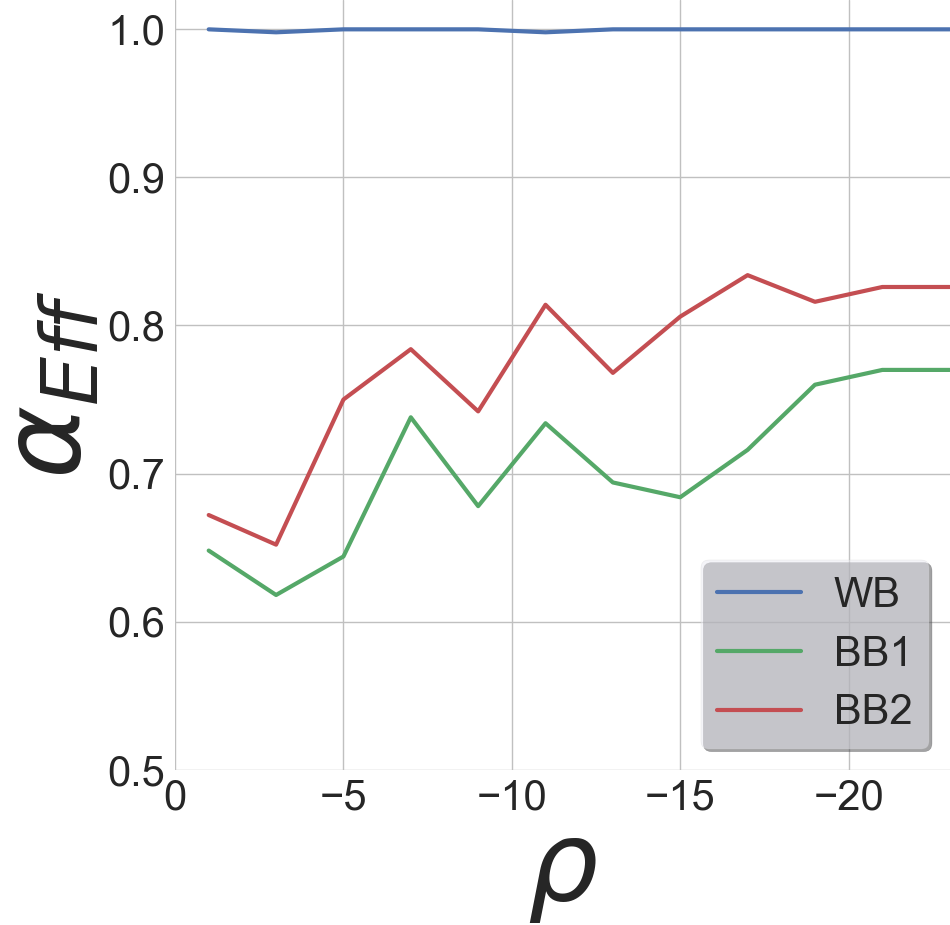}
        \caption{Performance of the fooling rate on a subset of WD dataset with a variable $\rho$ for the instance-specific attack setting.}
        \label{fig:rhosing}
\end{minipage}%
\hspace{2em}
\begin{minipage}{.45\textwidth}
        \centering
        \includegraphics[width=.8\linewidth]{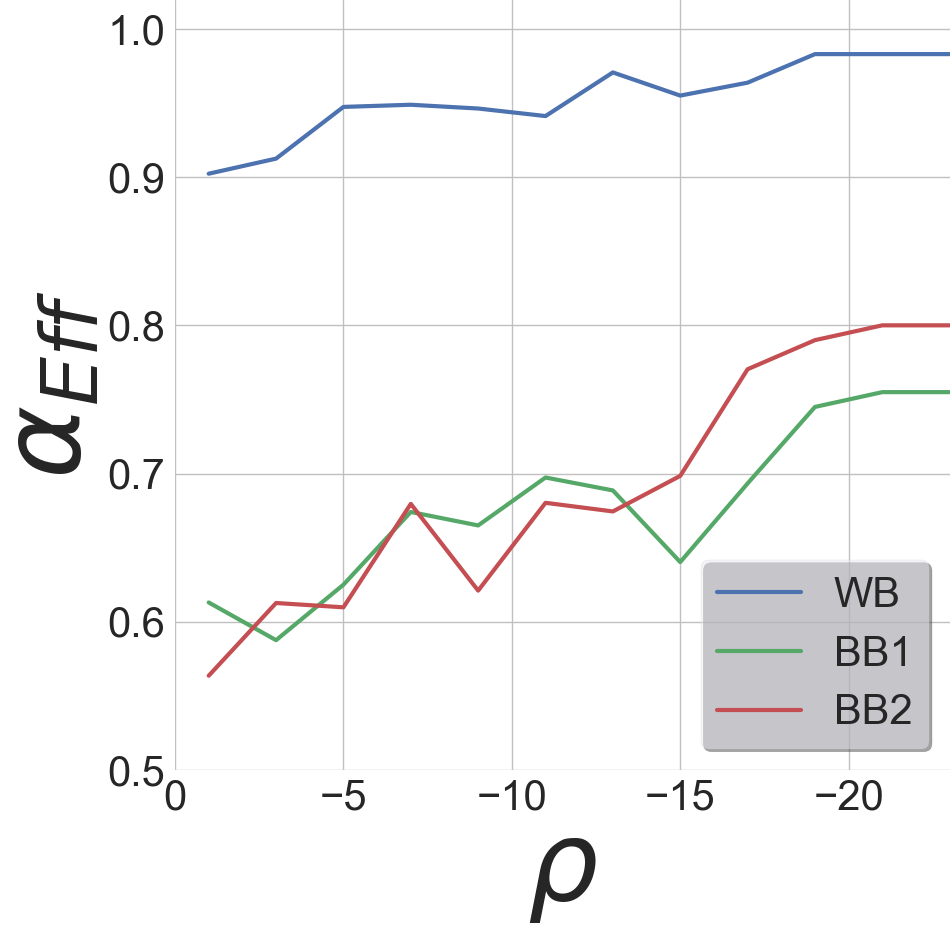}
        \caption{Performance of the fooling rate on a subset of WD dataset with a variable $\rho$ for the universal attack setting.}
        \label{fig:rhouniv}
\end{minipage}
\vspace{-1em}
\end{figure}

\noindent \textbf{Effectiveness of adversarial examples from TSA-STAT.} All following experiments were repeated 10 times and we report the averaged results (variance was negligible). We have used the standard benchmark training, validation, and test split on the datasets. We implemented the TSA-STAT framework using TensorFlow \cite{tensorflow2015} and the baselines using the CleverHans library \cite{papernot2018cleverhans}. We employ $\rho$=-20 for $\mathfrak{L}^{label}$ in Equation \ref{eq:classloss}. The choice was due to the observations made from Figures \ref{fig:rhosing} and \ref{fig:rhouniv}. A low value of $\rho$ has worse performance on generalization to unseen data or black-box models. However, higher values of $\rho$ slow down the convergence on each data point. Hence, we picked a confidence value of $\rho$ at which the fooling rate performance did not increase significantly.

Adversarial examples are generated for $\mathcal{L} < 0.1$ with a maximum of $5 \times 10^3$ iterations of gradient descent using the learning rate $\eta$=0.01.  We construct a group of transformations $\{\mathcal{PT}(X,y)\}_{y \in Y}$, one for each class $y$ in $Y$. The transformation to be used depends on the initial output class-label predicted by the deep model for the given input $X$. Therefore, the universal transformation $\mathcal{PT}(X, y)=\sum_{k=0}^{d} a^y_k~X^k$ will transform the inputs of the same class-label into adversarial outputs belonging to the target class-label. A targeted attack is a more sophisticated attack, which exposes the vulnerability of a DNN model better than an untargeted attack. From an attacker's perspective, having an attack model that allows choosing the target classification label of the adversarial example is better. Hence, we use targeted attacks for our experimental setup to show that TSA-STAT has the best opportunity for exploring time-series adversarial examples. We run the algorithm repeatedly on all the different class labels as targets.  If the maximum iteration number is reached, we select the coefficients $\{a_k^y\}$ with the lowest corresponding loss. 

\begin{figure}[!h]
\centering
    \begin{minipage}[t]{\linewidth}
    \centering
        \begin{minipage}{.49\linewidth}
            \includegraphics[width=\linewidth]{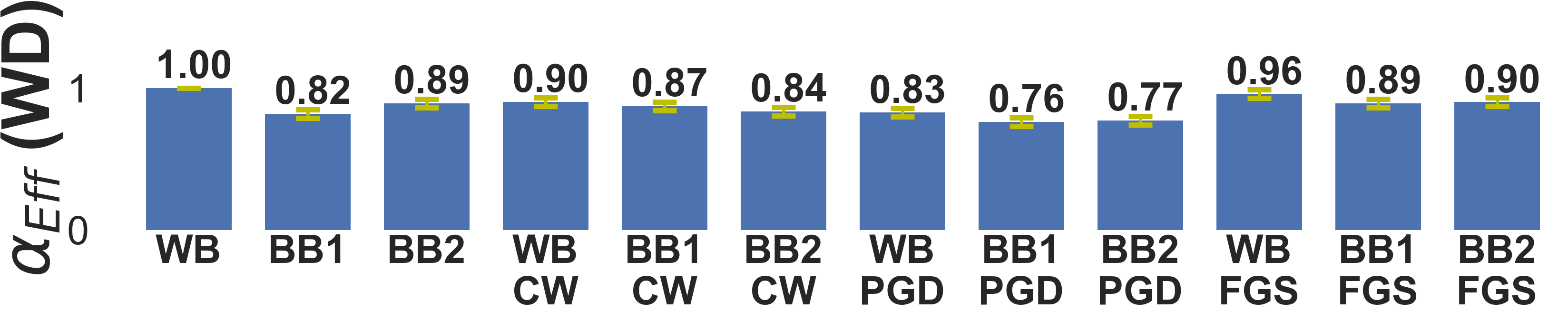}
        \end{minipage}%
        \begin{minipage}{.49\linewidth}
            \includegraphics[width=\linewidth]{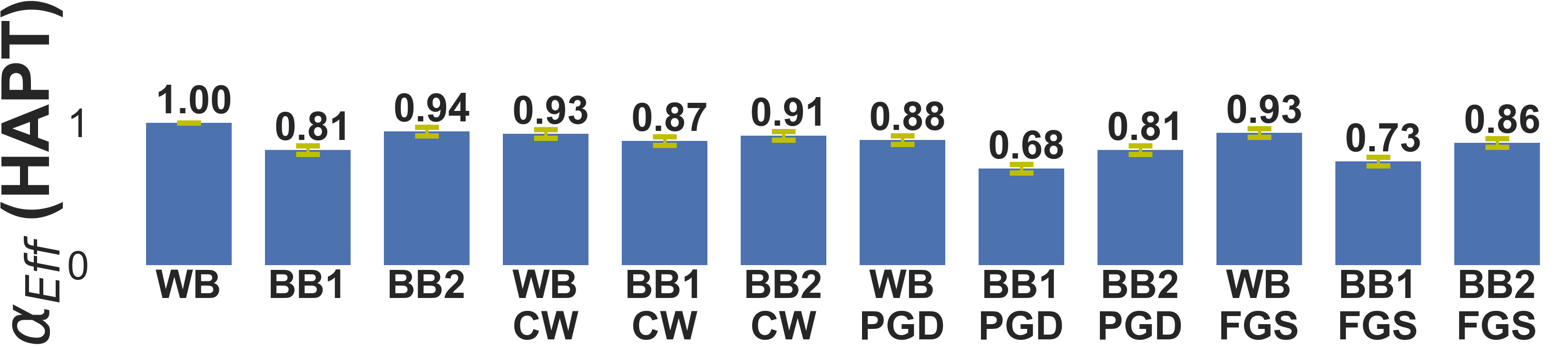}
        \end{minipage}
        \begin{minipage}{.49\linewidth}
            \includegraphics[width=\linewidth]{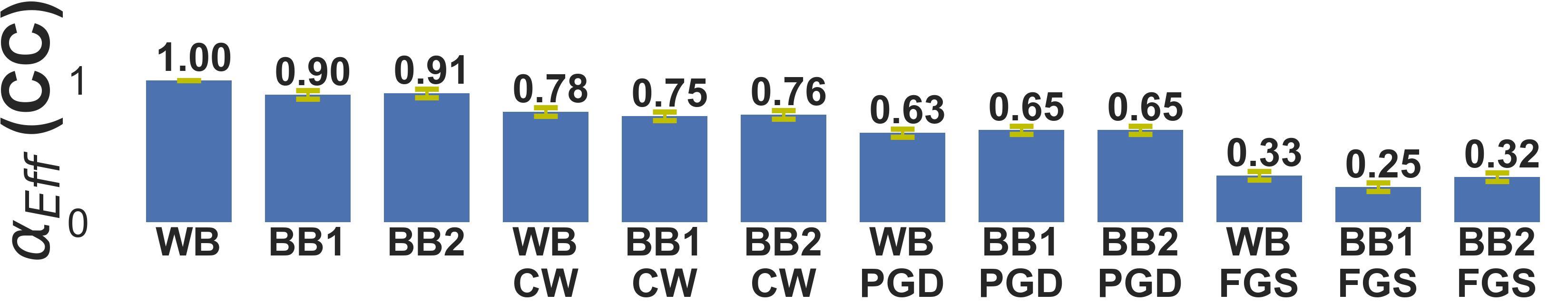}
        \end{minipage}%
        \begin{minipage}{.49\linewidth}
            \includegraphics[width=\linewidth]{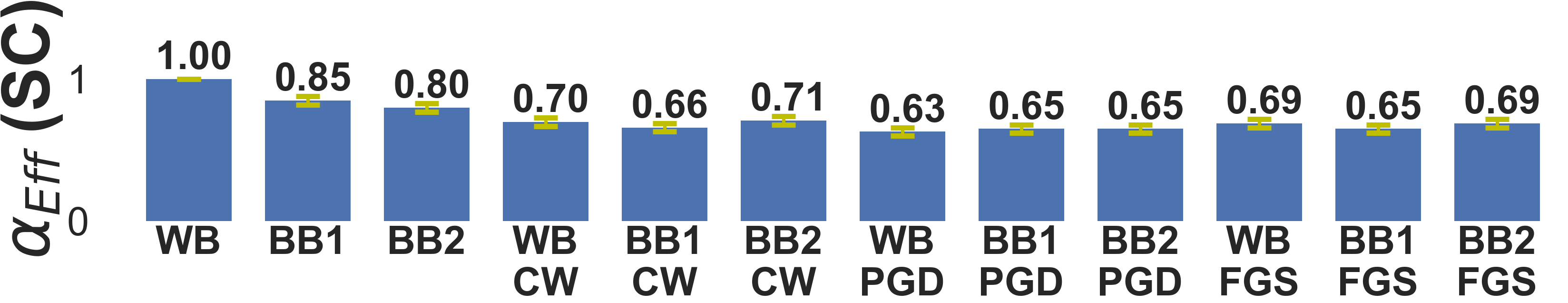}
        \end{minipage}
        \begin{minipage}{.49\linewidth}
            \includegraphics[width=\linewidth]{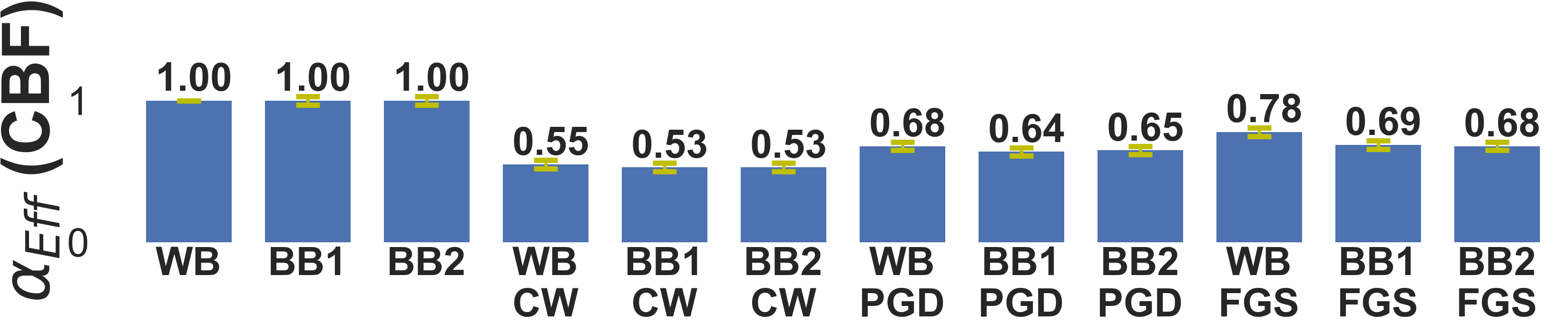}
        \end{minipage}%
        \begin{minipage}{.49\linewidth}
            \includegraphics[width=\linewidth]{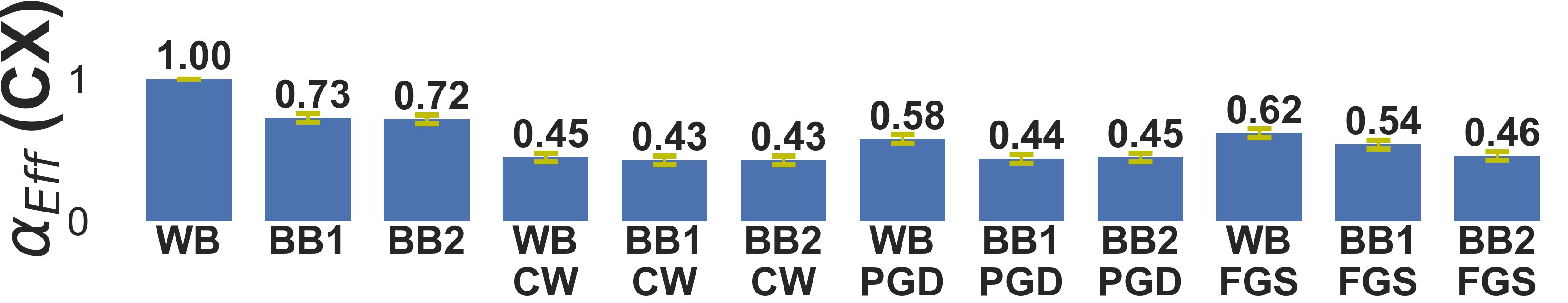}
        \end{minipage}
        \begin{minipage}{.49\linewidth}
            \includegraphics[width=\linewidth]{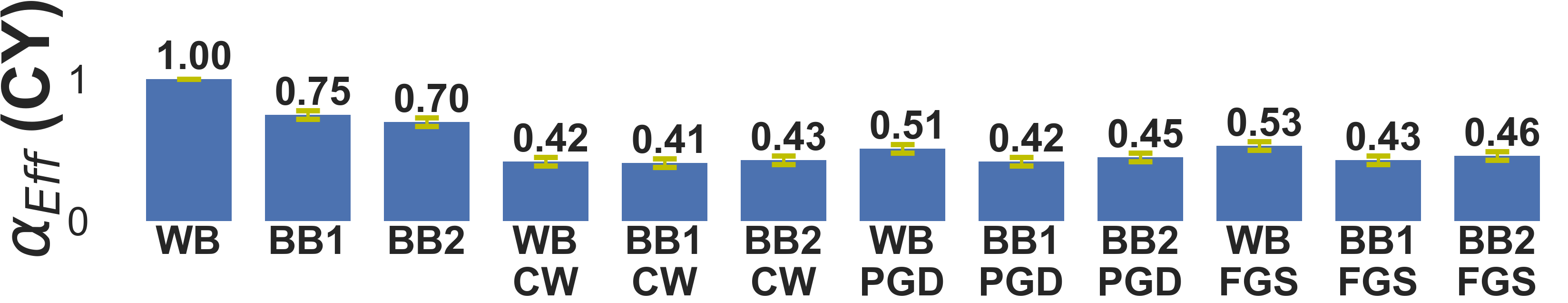}
        \end{minipage}%
        \begin{minipage}{.49\linewidth}
            \includegraphics[width=\linewidth]{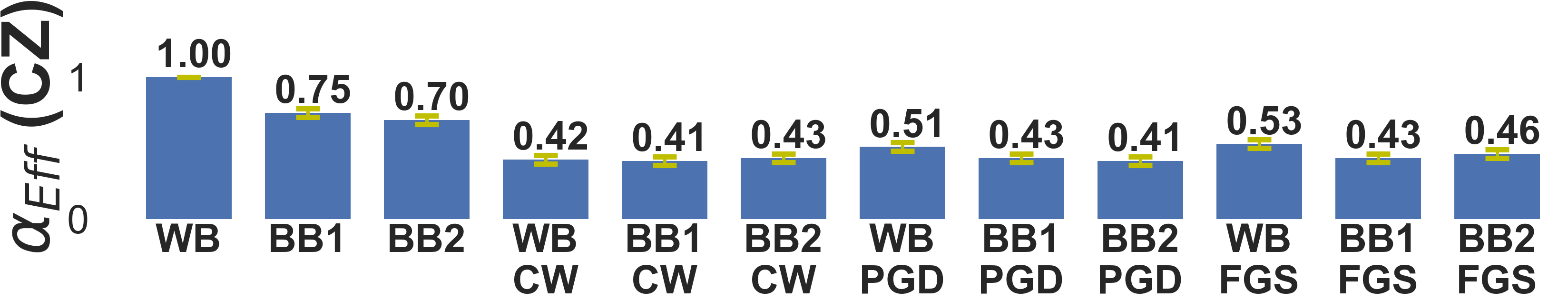}
        \end{minipage}
        \begin{minipage}{.49\linewidth}
            \includegraphics[width=\linewidth]{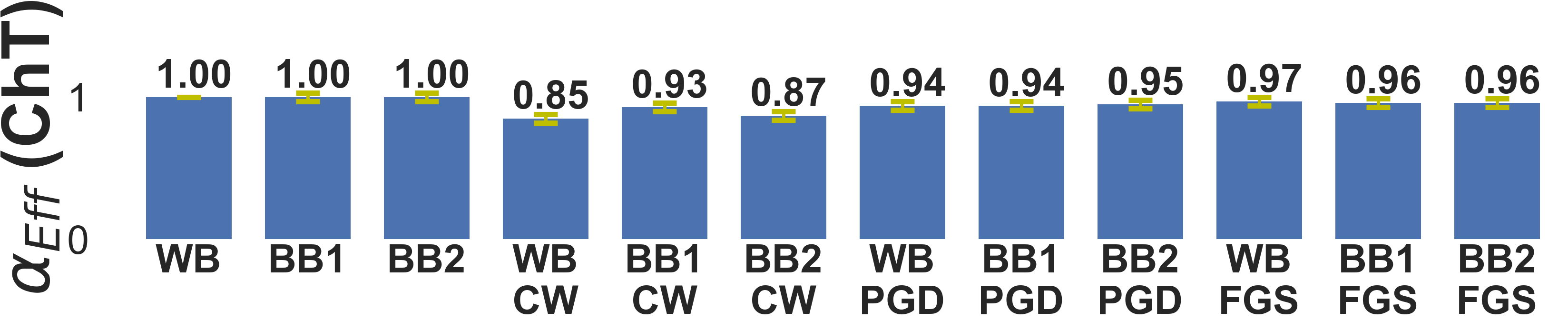}
        \end{minipage}
\end{minipage}
        \caption{Results for TSA-STAT instance-specific adversarial examples on different deep models trained with clean data and adversarial training baselines. 
        }
        \label{fig:singPerf}
        
        \vspace{-1em}
\end{figure}

\begin{figure}[!h]
\centering
    \begin{minipage}[t]{\linewidth}
    \centering
        \begin{minipage}{.49\linewidth}
            \includegraphics[width=\linewidth]{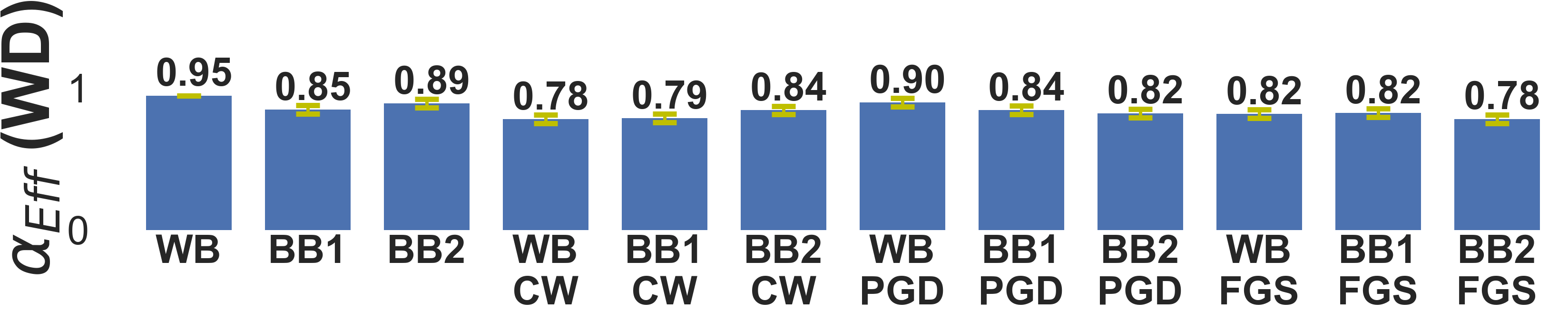}
        \end{minipage}%
        \begin{minipage}{.49\linewidth}
            \includegraphics[width=\linewidth]{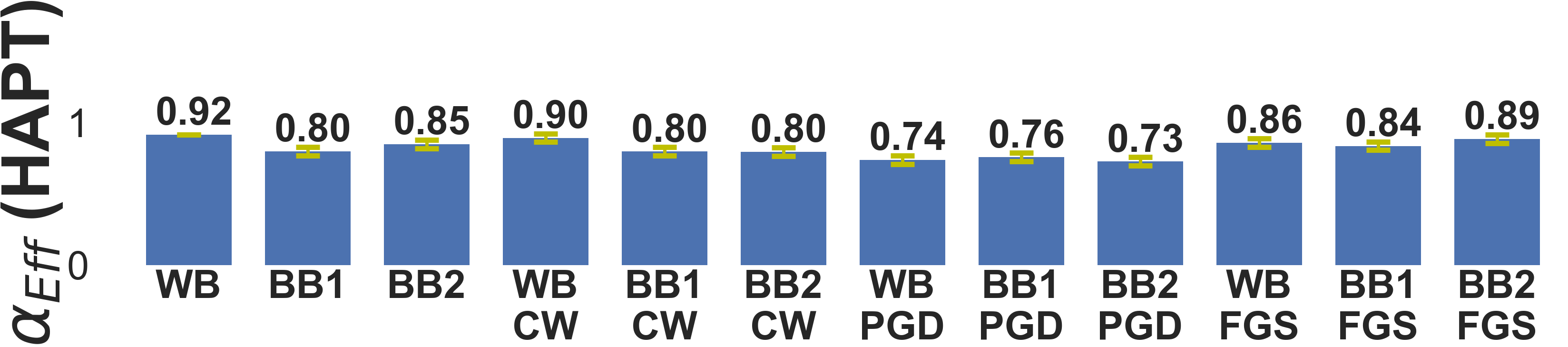}
        \end{minipage}
        \begin{minipage}{.49\linewidth}
            \includegraphics[width=\linewidth]{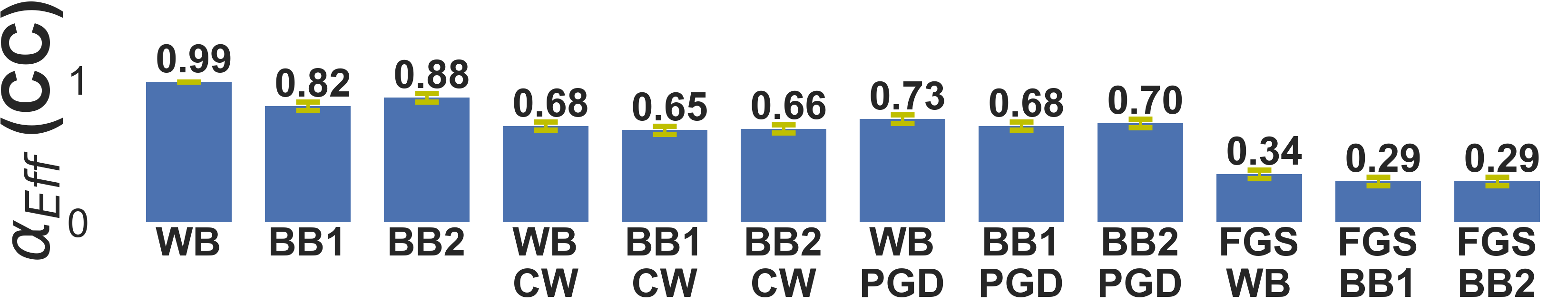}
        \end{minipage}%
        \begin{minipage}{.49\linewidth}
            \includegraphics[width=\linewidth]{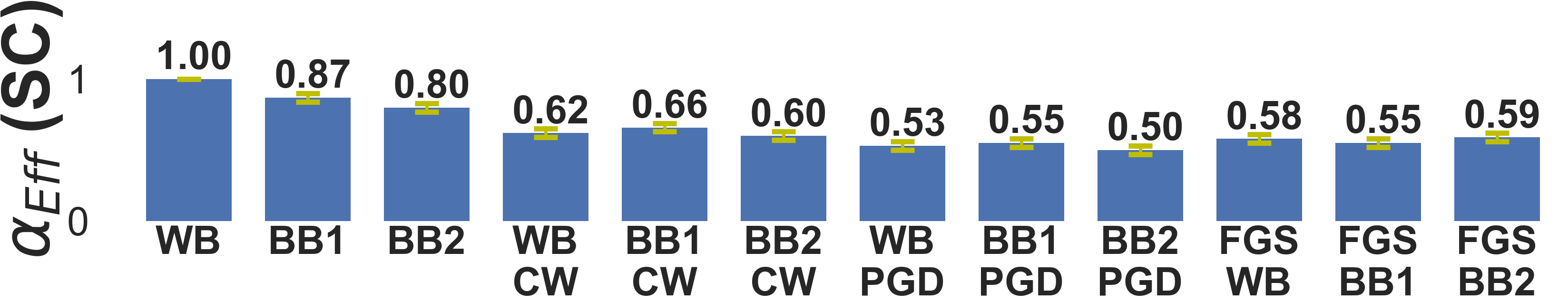}
        \end{minipage} 
        \begin{minipage}{.49\linewidth}
            \includegraphics[width=\linewidth]{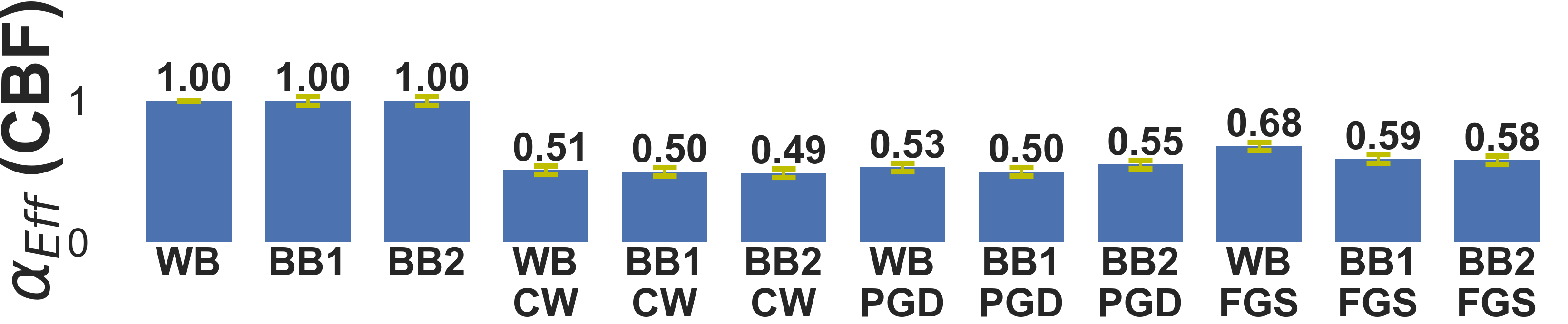}
        \end{minipage}%
        \begin{minipage}{.49\linewidth}
            \includegraphics[width=\linewidth]{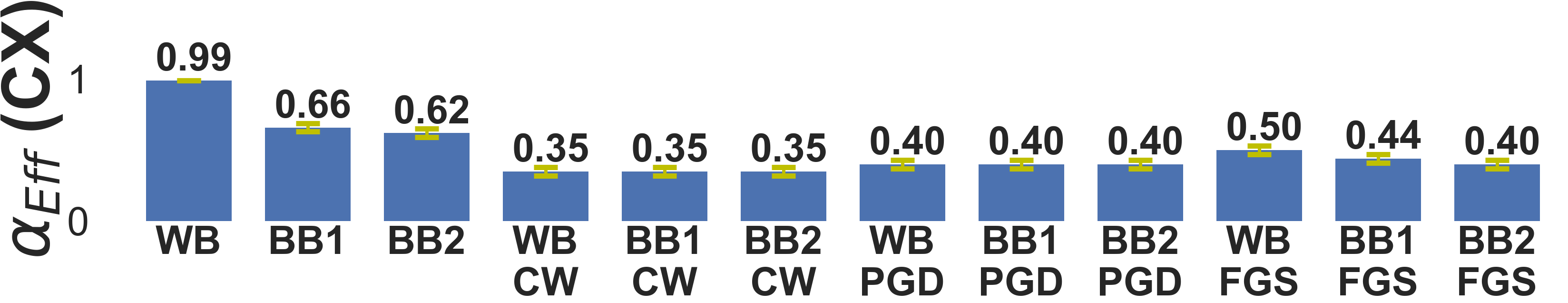}
        \end{minipage} 
        \begin{minipage}{.49\linewidth}
            \includegraphics[width=\linewidth]{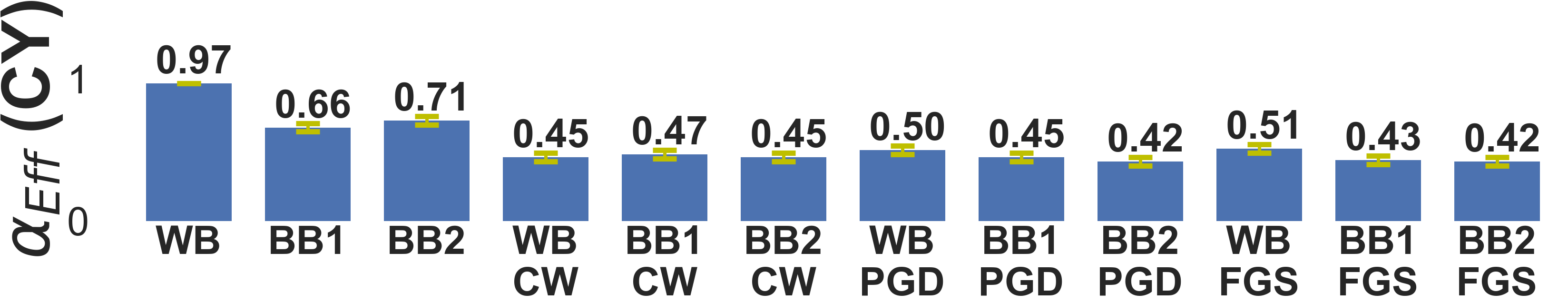}
        \end{minipage}%
        \begin{minipage}{.49\linewidth}
            \includegraphics[width=\linewidth]{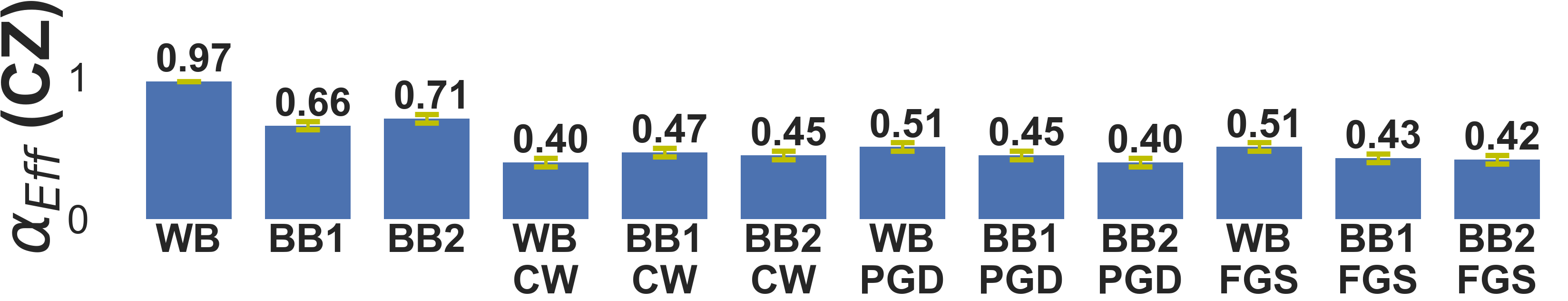}
        \end{minipage} 
        \begin{minipage}{.49\linewidth}
            \includegraphics[width=\linewidth]{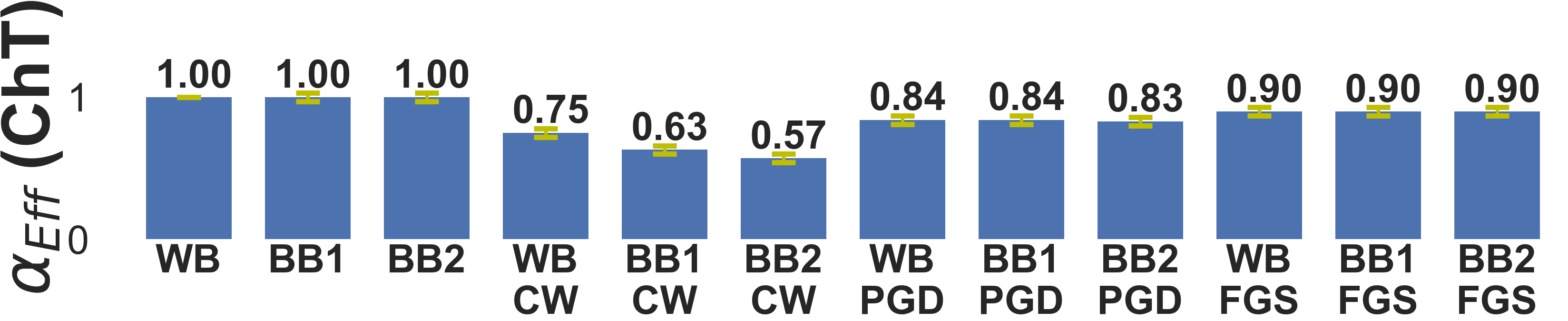}
        \end{minipage}
\end{minipage}
        \caption{Results for TSA-STAT universal adversarial examples on different deep models trained with clean data and adversarial training baselines.}
        \label{fig:univPerf}
        \vspace{-3em}
\end{figure}

We show the effectiveness of created adversarial examples for different settings (white-box, black-box etc.) to fool deep models for time-series domain. We evaluate TSA-STAT using the attack efficiency metric  $\alpha_{Eff} \in [0,1]$ over the created adversarial examples. $\alpha_{Eff}$ (higher means better attacks) measures the capability of targeted adversarial examples to fool a given DNN classifier $F_{\theta}$ to output the class-label $y_{target}$ (i.e., targeted attacks). Figure \ref{fig:singPerf} shows the results for instance-specific targeted attacks under white-box and black-box settings on different deep models. Figure \ref{fig:univPerf} shows the results for universal attacks using TSA-STAT. Unlike instance-specific attacks, universal attacks are created by directly using the resulting polynomial transformation $\mathcal{PT}(\cdot)$. Recall that for black-box attacks, we do not query the target deep model at any phase. While comparing TSA-STAT based attack with the existing attacks using success rate provides an assessment about the performance of different attacks, it does not show which attack is stronger. To investigate the real performance of TSA-STAT, we show the effectiveness of the TSA-STAT based attack to fool deep models for time-series domain using both standard and adversarial training. If any baseline algorithm were better attacks than TSA-STAT, the adversarial training using that baseline will be robust towards TSA-STAT's attacks.

\begin{figure}[!h]
\centering
    \begin{minipage}[t]{\linewidth}
    \centering
        \begin{minipage}{.49\linewidth}
            \includegraphics[width=\linewidth]{Figures/SingPeroformancewisdm.png}
        \end{minipage}%
        \begin{minipage}{.49\linewidth}
            \includegraphics[width=\linewidth]{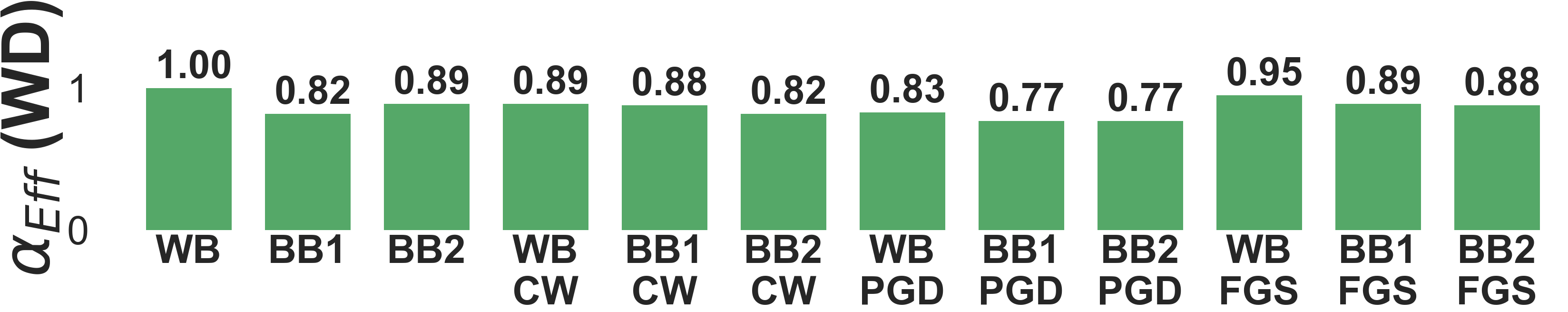}
        \end{minipage}
        \begin{minipage}{.49\linewidth}
            \includegraphics[width=\linewidth]{Figures/SingPeroformancehapt.png}
        \end{minipage}%
        \begin{minipage}{.49\linewidth}
            \includegraphics[width=\linewidth]{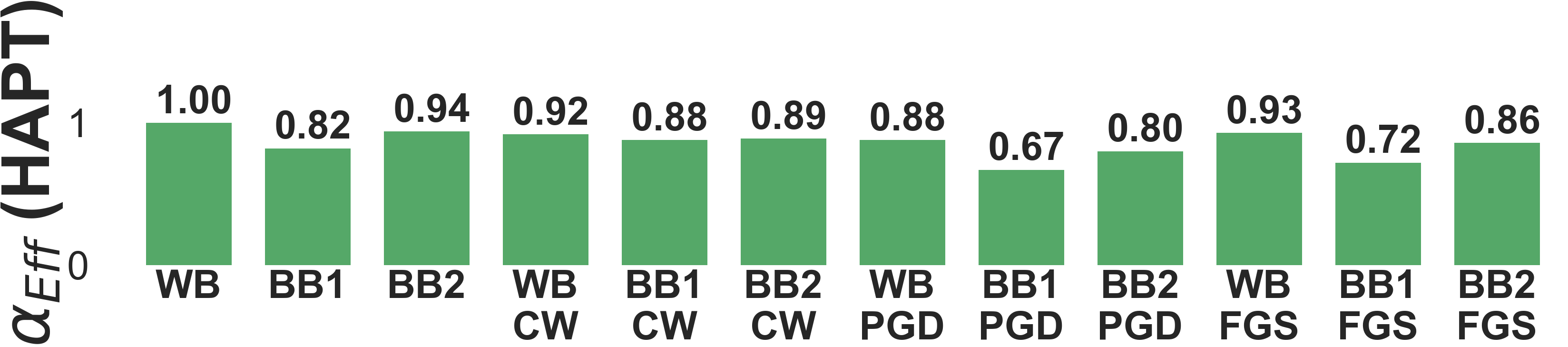}
        \end{minipage}
        \begin{minipage}{.49\linewidth}
            \includegraphics[width=\linewidth]{Figures/SingPeroformanceCC.png}
        \end{minipage}%
        \begin{minipage}{.49\linewidth}
            \includegraphics[width=\linewidth]{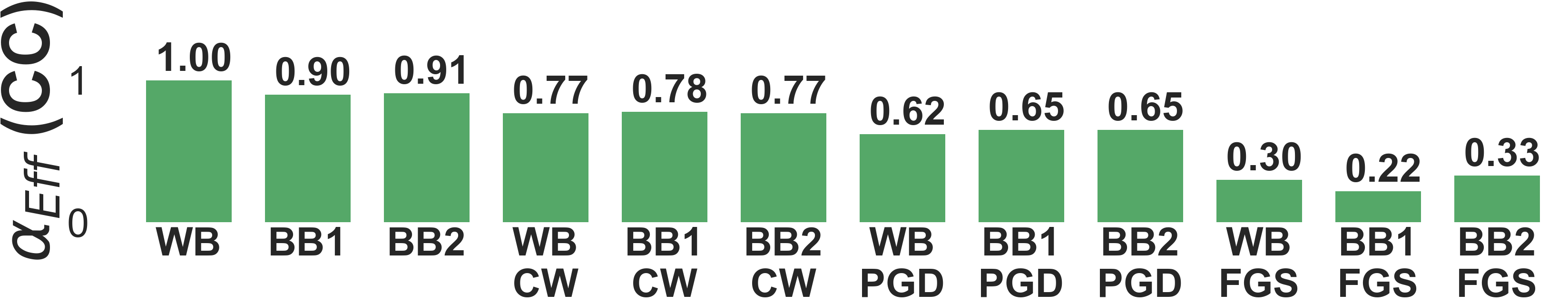}
        \end{minipage}
        \begin{minipage}{.49\linewidth}
            \includegraphics[width=\linewidth]{Figures/SingPeroformanceSC.png}
        \end{minipage} %
        \begin{minipage}{.49\linewidth}
            \includegraphics[width=\linewidth]{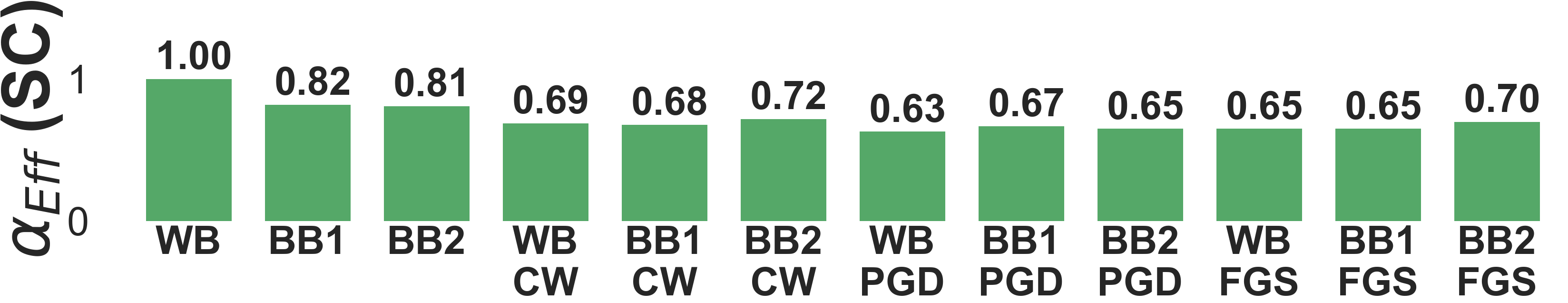}
        \end{minipage} 
\end{minipage}
        \caption{Results for TSA-STAT instance-specific adversarial examples on different deep models trained with clean data and adversarial training baselines using $l_\infty$ (shown in Blue) and $l_2$ (shown in Green) norm on the statistical features.}
        \label{fig:statnorm}
\end{figure}

\begin{figure}[!h]
\centering
\begin{minipage}[t]{\linewidth}
        \centering
        \begin{minipage}{.49\linewidth}
            \includegraphics[width=\linewidth]{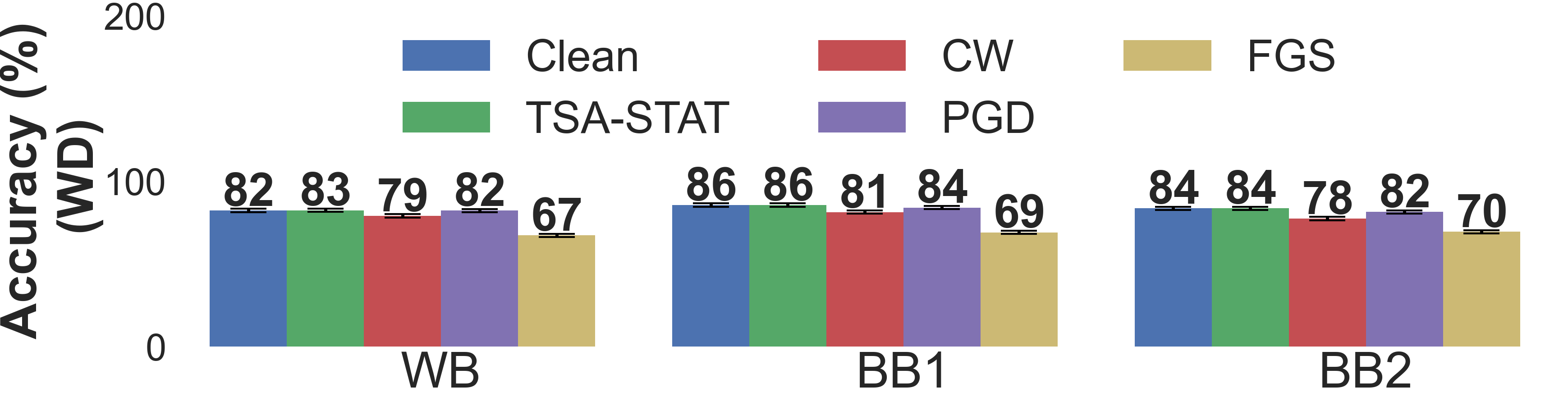}
        \end{minipage}%
        \begin{minipage}{.49\linewidth}
            \includegraphics[width=\linewidth]{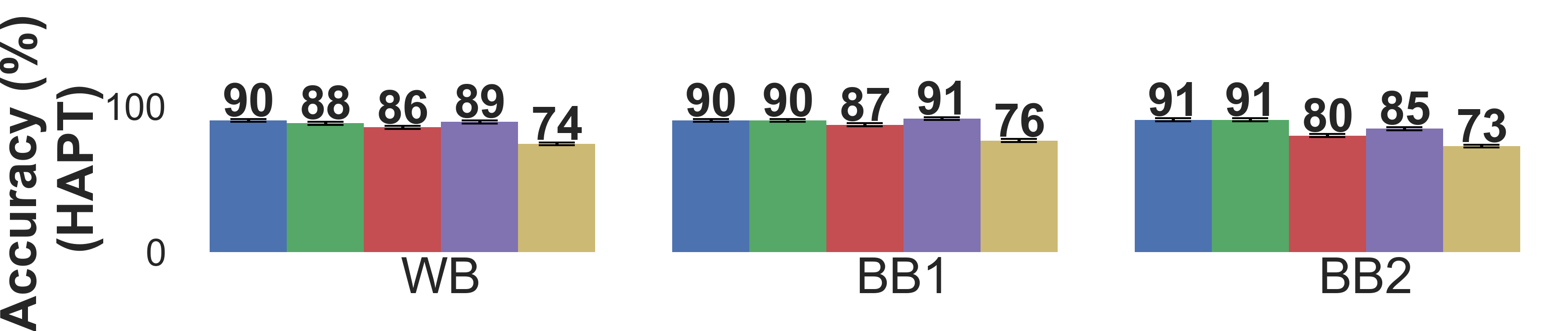}
        \end{minipage}
        \begin{minipage}{.49\linewidth}
            \includegraphics[width=\linewidth]{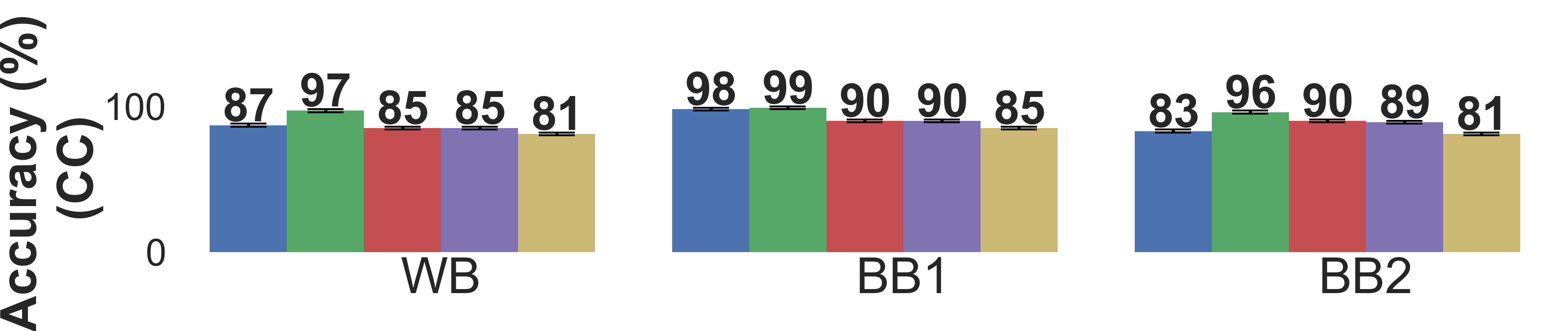}
        \end{minipage}%
        \begin{minipage}{.49\linewidth}
            \includegraphics[width=\linewidth]{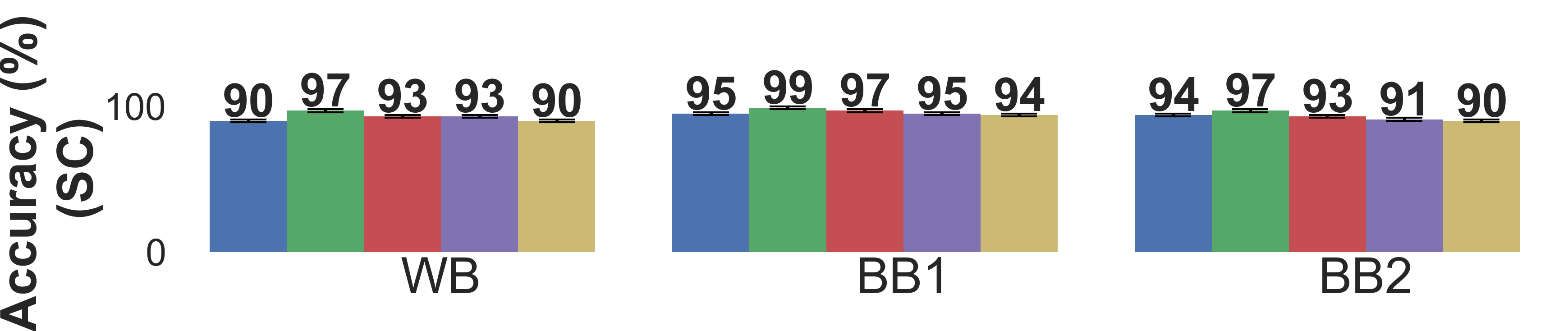}
        \end{minipage}
        \begin{minipage}{.49\linewidth}
            \includegraphics[width=\linewidth]{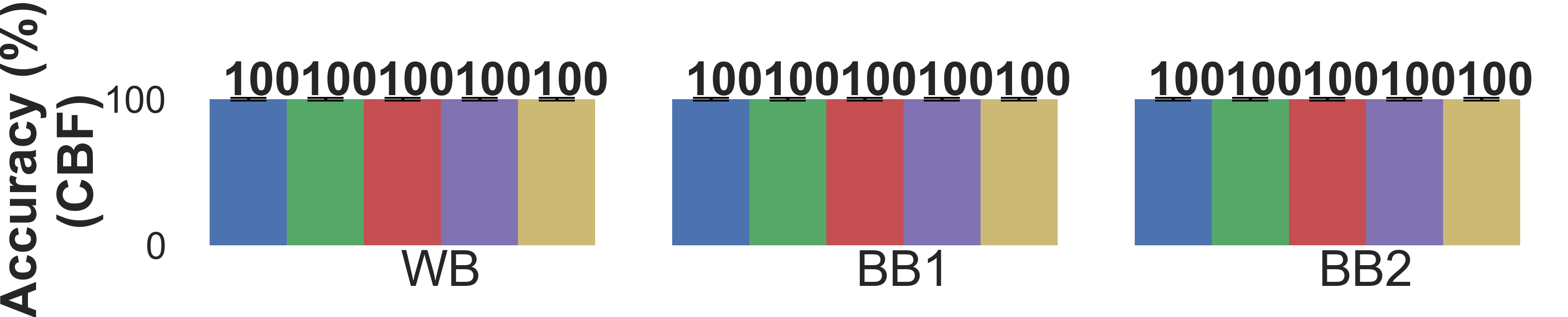}
        \end{minipage}%
        \begin{minipage}{.49\linewidth}
            \includegraphics[width=\linewidth]{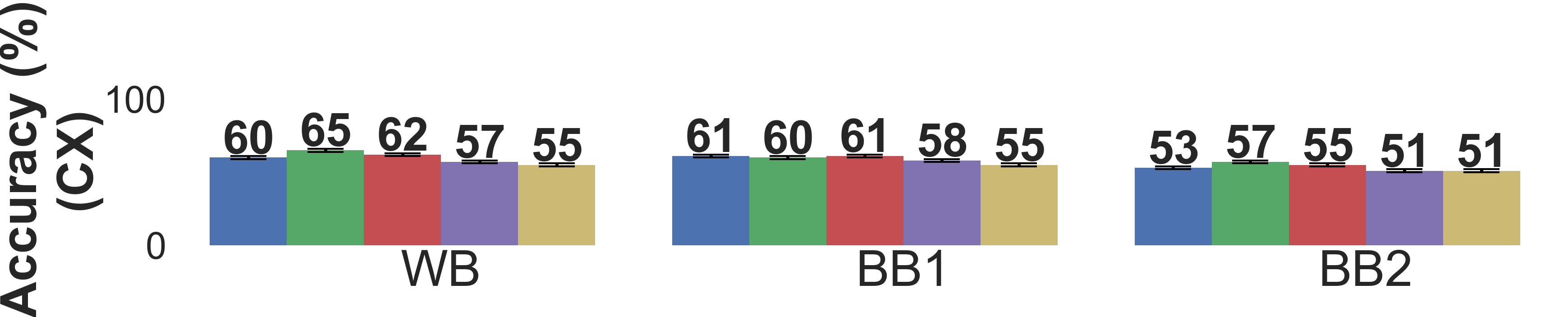}
        \end{minipage}
        \begin{minipage}{.49\linewidth}
            \includegraphics[width=\linewidth]{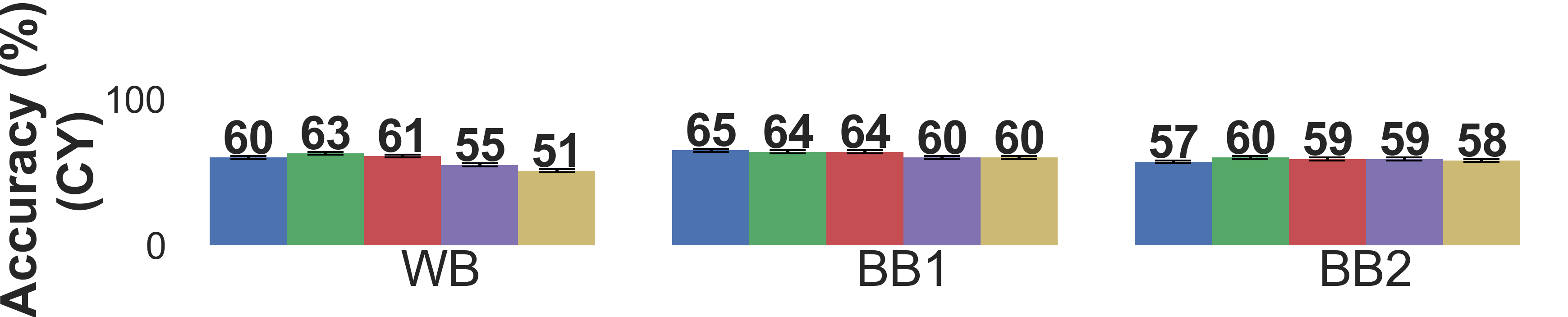}
        \end{minipage}%
        \begin{minipage}{.49\linewidth}
            \includegraphics[width=\linewidth]{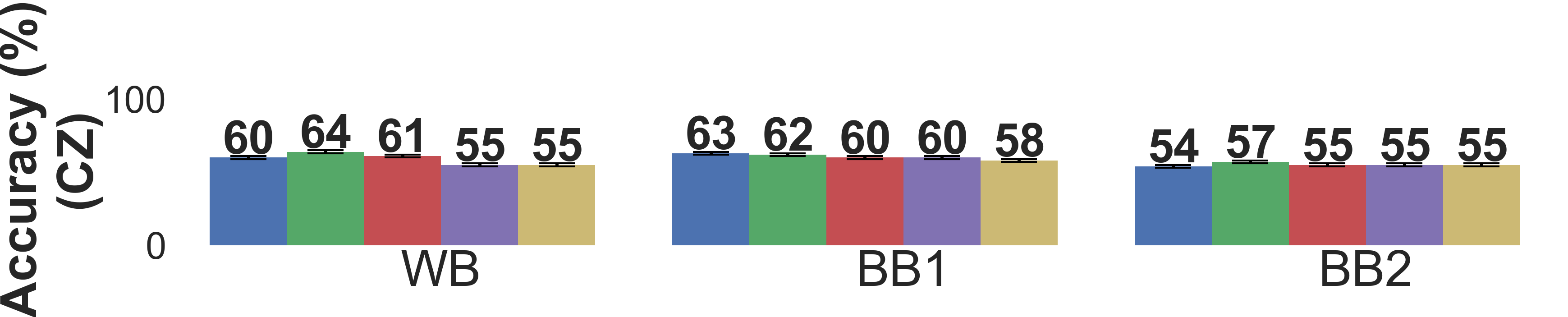}
        \end{minipage}
        \begin{minipage}{.49\linewidth}
            \includegraphics[width=\linewidth]{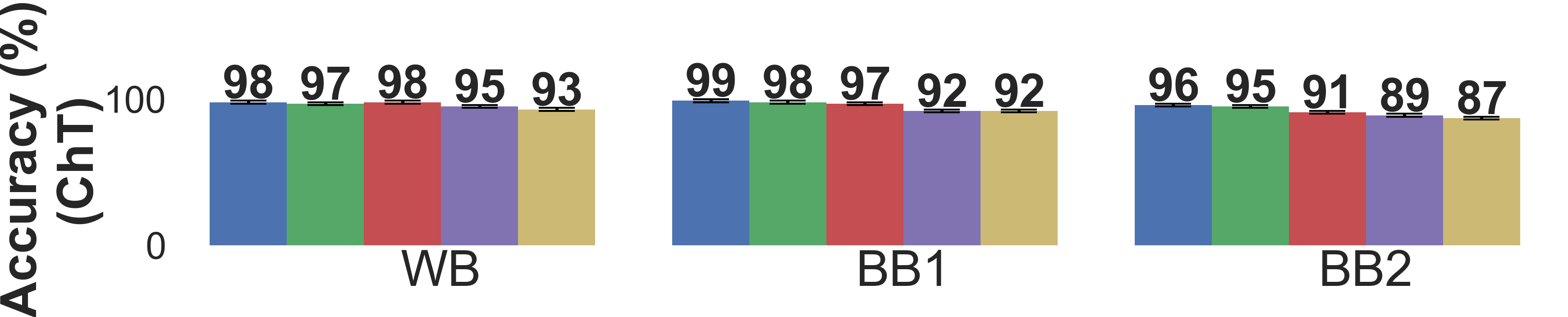}
        \end{minipage}
\end{minipage}
        \caption{Results for adversarial training using adversarial examples from different methods including TSA-STAT, FGS, CW, PGD, and standard training (Clean) on clean testing data for different deep models.}
        \label{fig:advTrnPerf}
\end{figure}

We can observe from both Figures \ref{fig:singPerf} and \ref{fig:univPerf} that on the multivariate WD and HAPT dataset, the fooling rate is good across all settings. Adversarial examples created by optimized $\mathcal{PT}(\cdot)$ are highly effective as $\alpha_{Eff} \ge 0.7$ for most cases. For CC and SC datasets, we see a lower performance for TSA-STAT, essentially at the level of FGS on CC. We believe that this is due to the effect of $l_p$-bounded adversarial examples that mislead the deep models during adversarial training. Additionally, we show in Figure \ref{fig:statnorm} that using $l_2$ norm or $l_\infty$ norm on the statistical features has no difference in the general performance of TSA-STAT based attacks.
Finally, Figure \ref{fig:advTrnPerf} shows the results of different deep models after adversarial training using adversarial examples from different methods including TSA-STAT. This performance is relative to the clean testing set of the data. We can easily observe from the results of using FGS, CW, and PGD for adversarial training (degrades overall performance), the validity of our claim: $l_p$ distance-based perturbation lacks true-label guarantees and can degrade the overall performance of deep models on real-world data. On the other hand, by using the adversarial examples from TSA-STAT, the overall performance did not decrease and has improved for some datasets: for SC, the accuracy increased from 90\% to 97\% for $WB$, and accuracy on CC improved from 83\% to 96\% for $BB_2$. We observe that FGS was the worst method in terms of preserving the performance of deep models.

We conclude from the experiments to test the effectiveness of adversarial examples from TSA-STAT that indeed using statistical features is well-justified for adversarial time-series data. If the standard $L_p$-norm-based methods from the image domain were to be very effective for the time-series domain:
    \begin{itemize}
        \item TSA-STAT based attacks will not be able to fool the models using baselines as a defense method as shown in Figures \ref{fig:singPerf} and \ref{fig:univPerf}.
        \item Adversarial training on clean data using baseline methods would have outperformed TSA-STAT unlike the observations from Figure \ref{fig:advTrnPerf}.
    \end{itemize}
    
\vspace{1.0ex}

\noindent \textbf{Certified bounds.} Using Algorithm \ref{alg:certif}, we can infer the robustness of an input $X$ by calculating the upper bound $\delta$ that limits the tolerable adversarial perturbation over the $\|\cdot\|_{\infty}$. Hence, for any generated perturbation on $X$ which employs $\hat{\delta} \le \delta$, the classification result is guaranteed to remain the same.  In other words, for a given time-series $X$ and its robustness bound $\delta$, the perturbation $\hat{\delta}$ can take any value  $\le \delta$. As a consequence, the classification of a time-series input $X$ with the perturbation $\hat{\delta}$ is stable/certified. For the following experiments, we employ $MAX=5\times 10^3$. For the generation of $\Sigma$, we use a random algorithm to generate a semi-definite positive matrix that has parameter $\sigma$ as diagonal elements.  

\begin{figure}[!h]
    \centering
    \begin{minipage}{.39\linewidth}
            \centering
            \includegraphics[width=\linewidth]{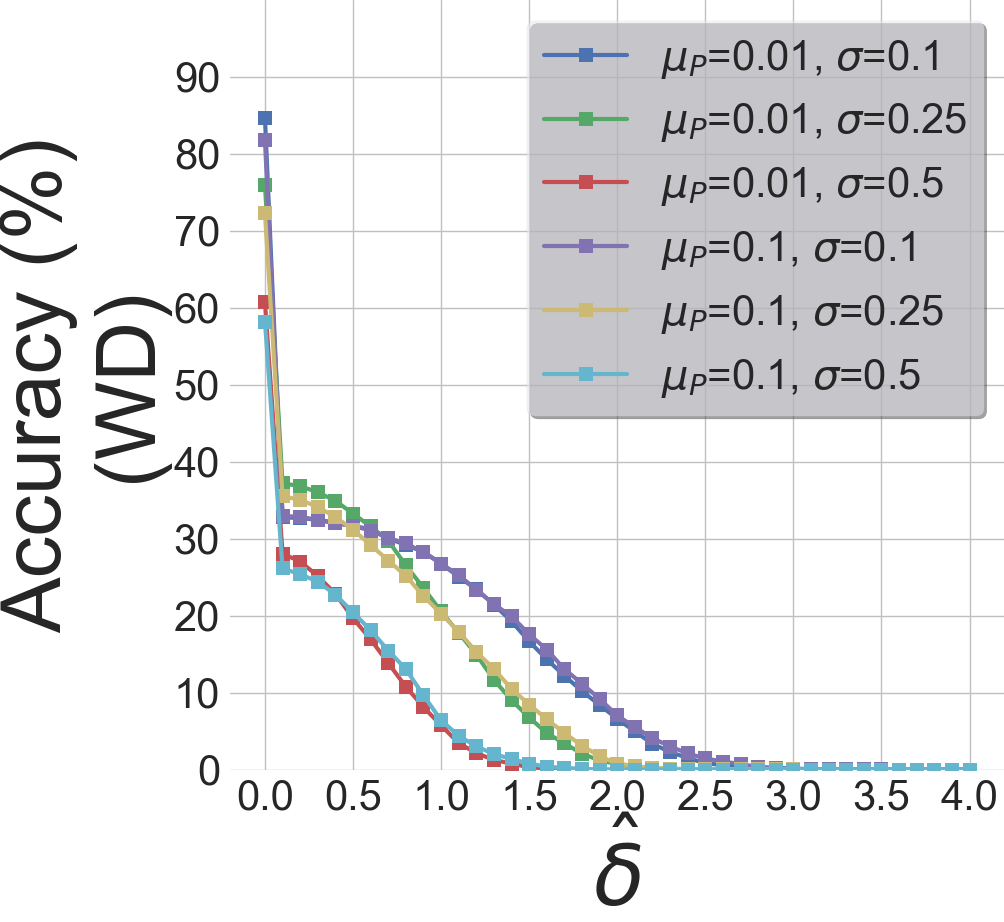}
        \end{minipage}%
        \begin{minipage}{.39\linewidth}
            \centering
            \includegraphics[width=\linewidth]{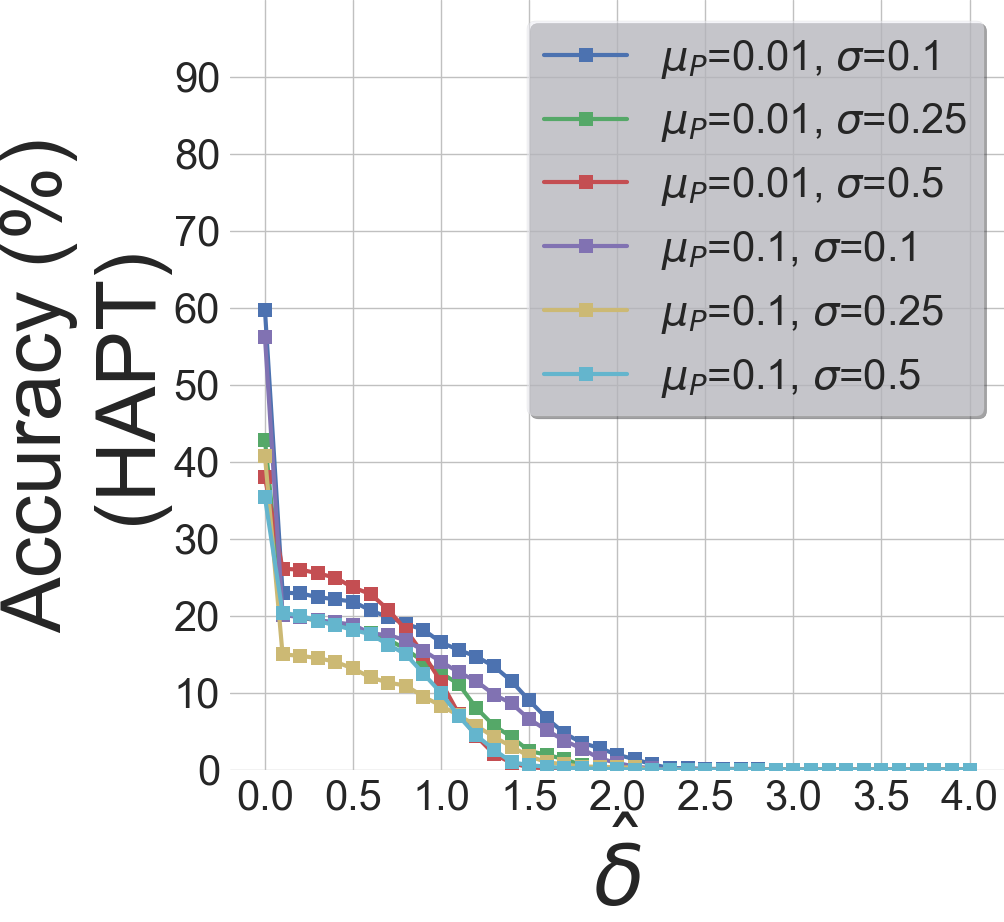}
        \end{minipage}
        \begin{minipage}{.39\linewidth}
            \centering
            \includegraphics[width=\linewidth]{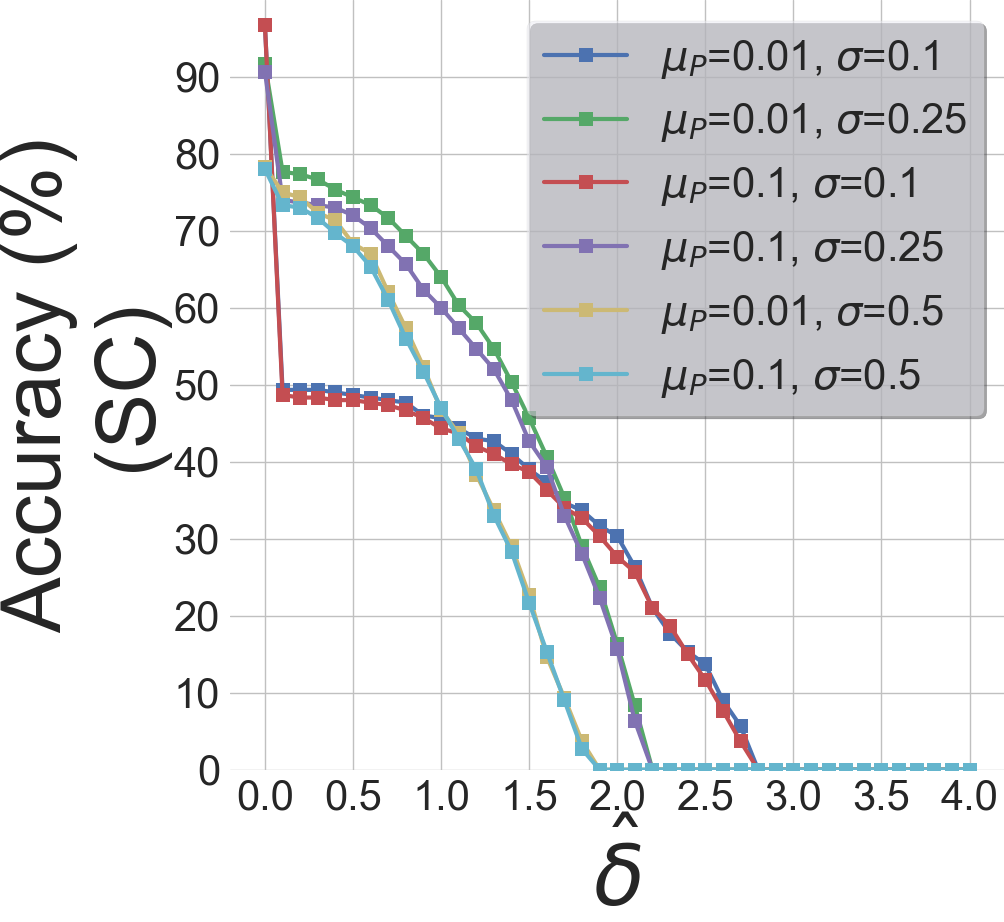}
        \end{minipage}%
        \begin{minipage}{.39\linewidth}
            \centering
            \includegraphics[width=\linewidth]{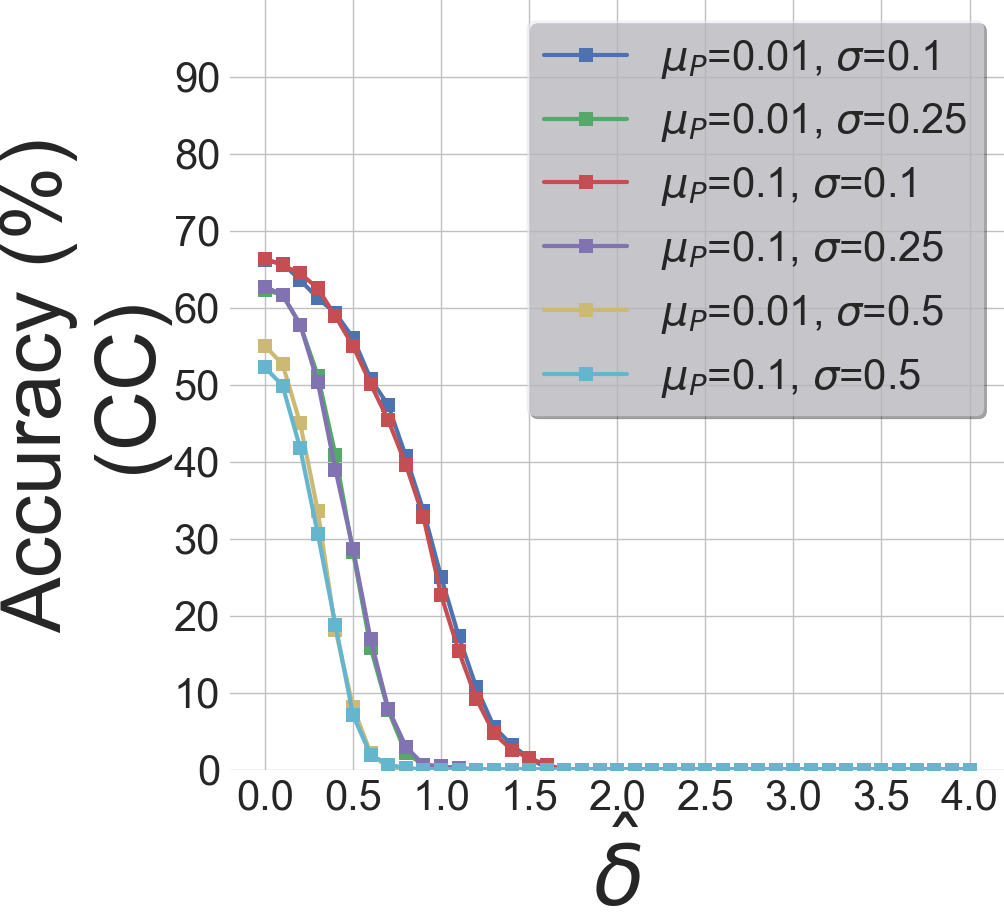}
        \end{minipage}%
    \caption{Certification lower bound accuracy on the testing data with varying $({\mu_P}, \sigma)$ for Algorithm \ref{alg:certif}.}
    \label{fig:certif}
\end{figure}

Figure \ref{fig:certif} shows the classification accuracy on testing set under the attack of different possible $\hat{\delta}$ with various choices of $\sigma=\|\sum_{i,i}\|_{\infty}$ for the multivariate Gaussian $n_p$ and $n_q$ of the Algorithm \ref{alg:certif} (noting that we observed similar findings on other datasets). $\sigma$ refers to the diagonal element of the covariance matrix $\sum$. As an example, for WD dataset where $(\mu_P=0.1, \sigma=0.1)$: At $\hat{\delta}=0$, we have a testing accuracy of 0.83, which translates to the fact that 83\% of the testing set is robust to the given perturbation and 19\% of the testing test is vulnerable to adversarial attacks. We also observe that the larger the value of $\sigma$ is, the faster the curve declines. This shows that inputs are unstable with respect to robustness to noises with higher $\sigma$.

\begin{figure}[!h]
    \centering
    \begin{minipage}{\linewidth}
    \centering
        \begin{minipage}{.33\linewidth}
            \centering
            \includegraphics[width=\linewidth]{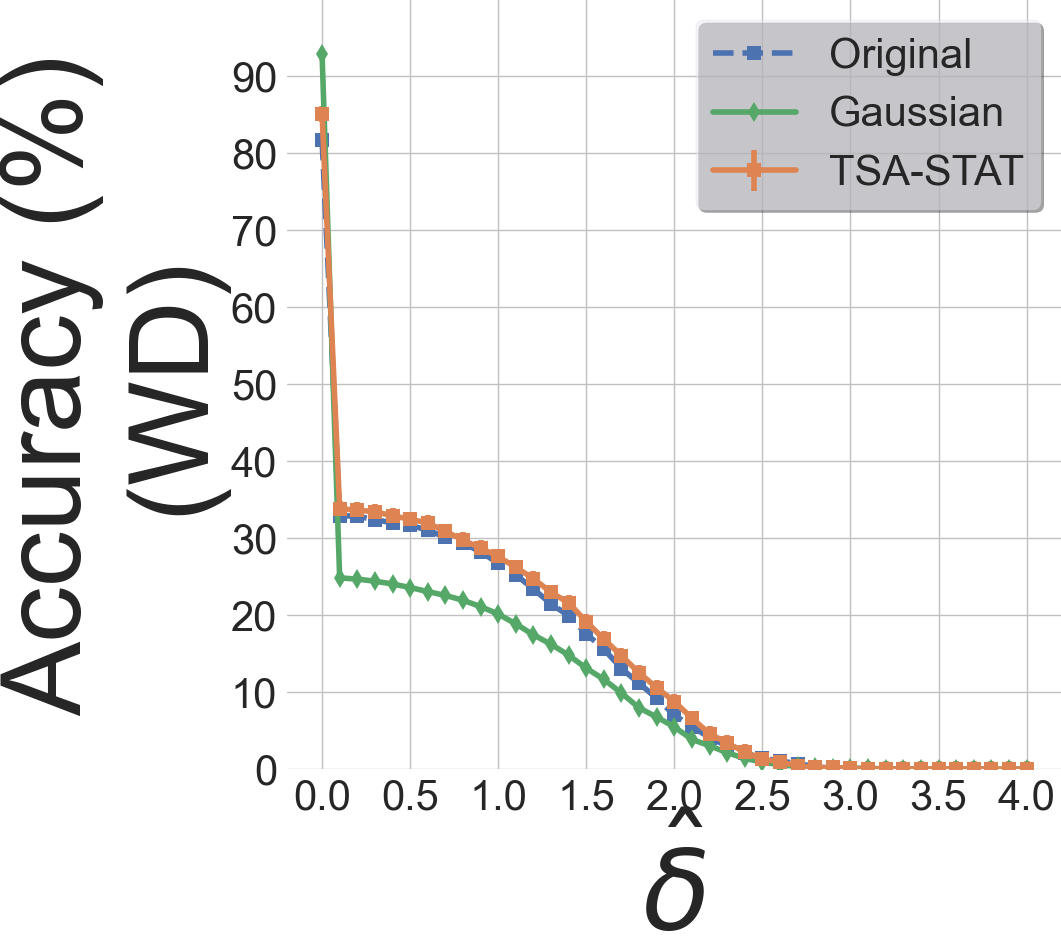}
        \end{minipage}%
        \begin{minipage}{.33\linewidth}
            \centering
            \includegraphics[width=\linewidth]{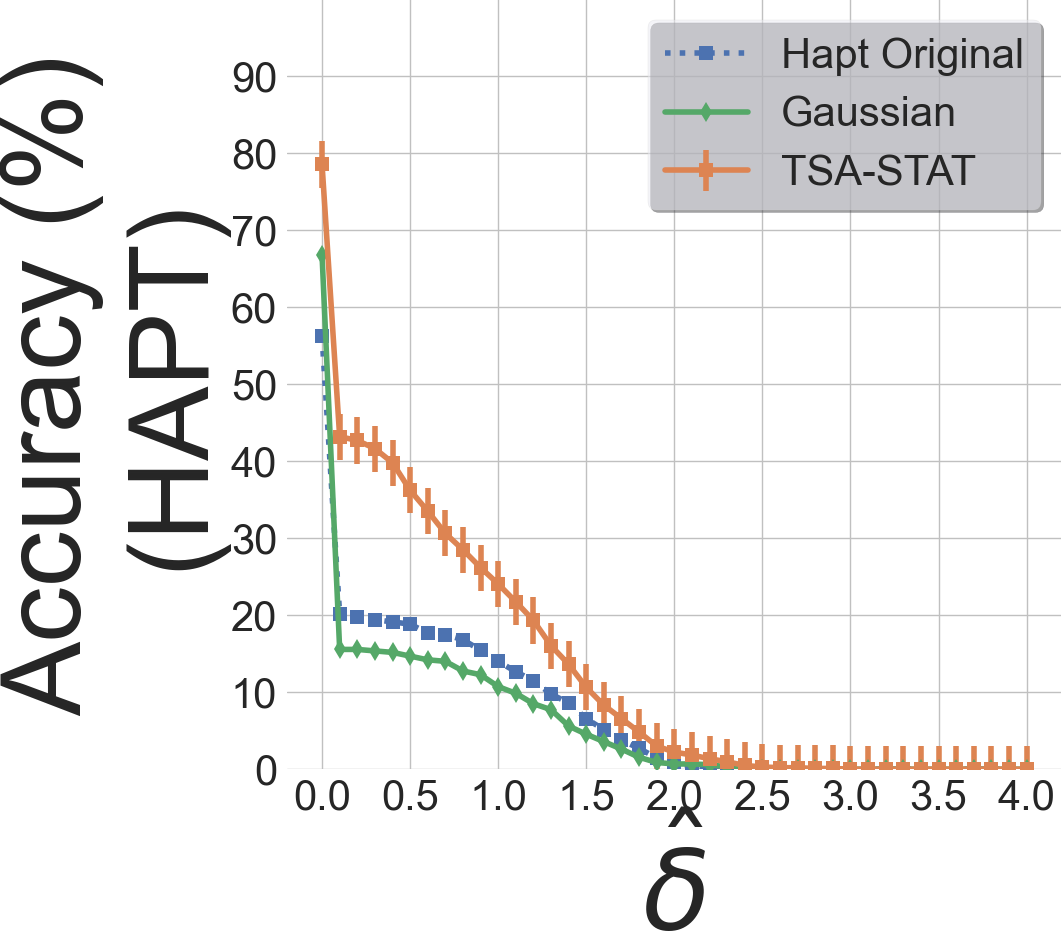}
        \end{minipage}%
        \begin{minipage}{.33\linewidth}
            \centering
            \includegraphics[width=\linewidth]{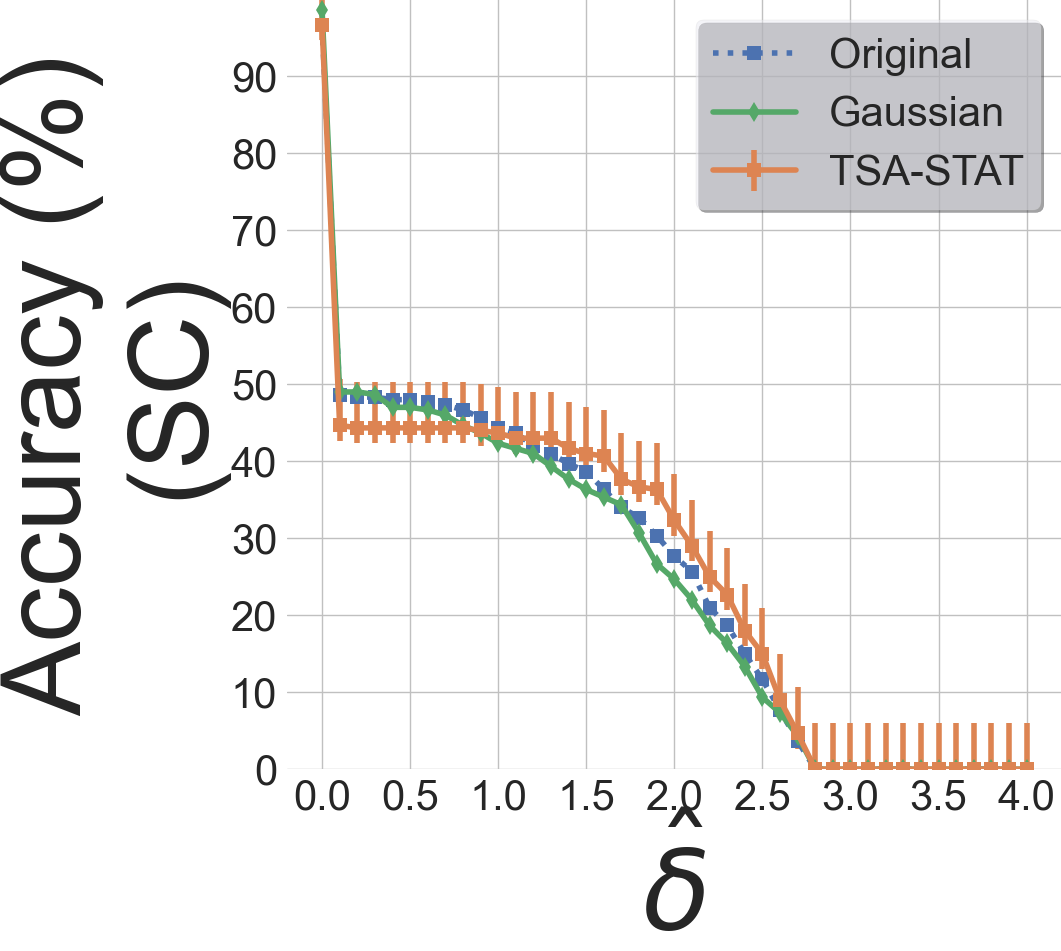}
        \end{minipage}
        \begin{minipage}{.33\linewidth}
            \centering
            \includegraphics[width=\linewidth]{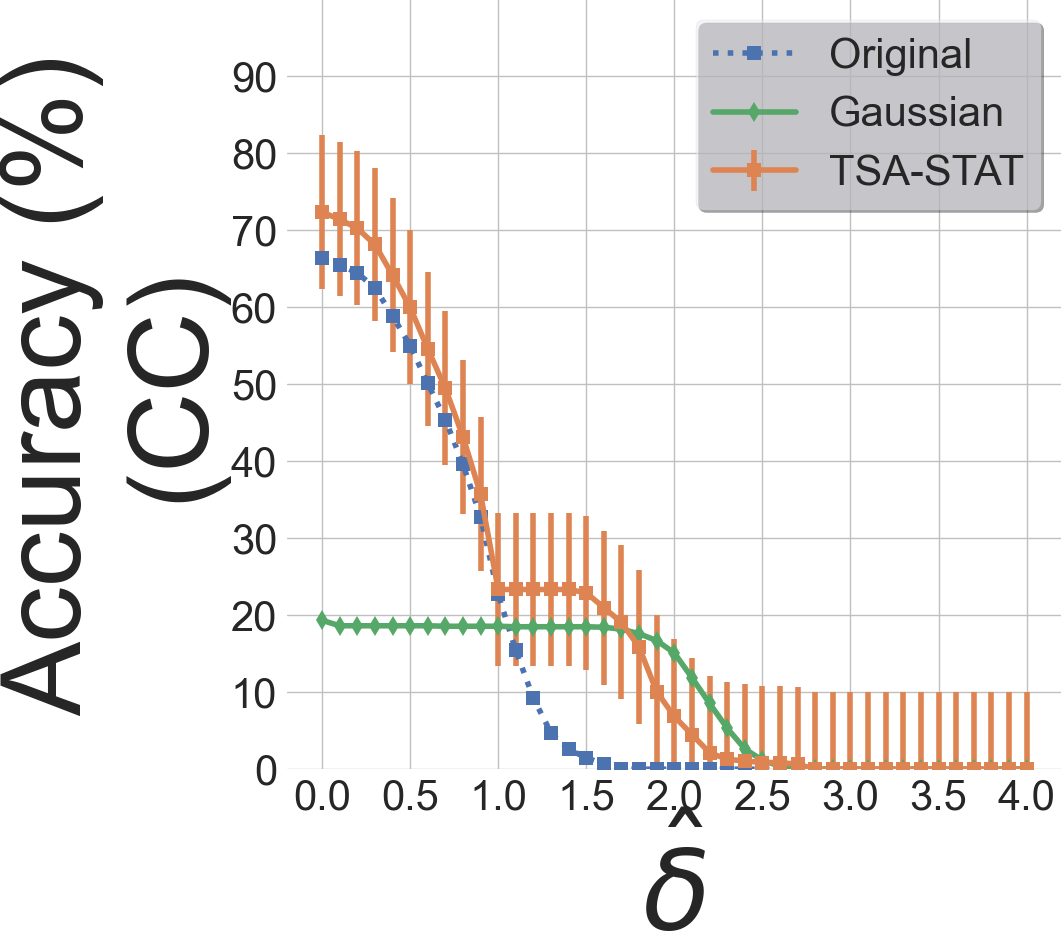}
        \end{minipage}%
        \begin{minipage}{.33\linewidth}
            \centering
            \includegraphics[width=\linewidth]{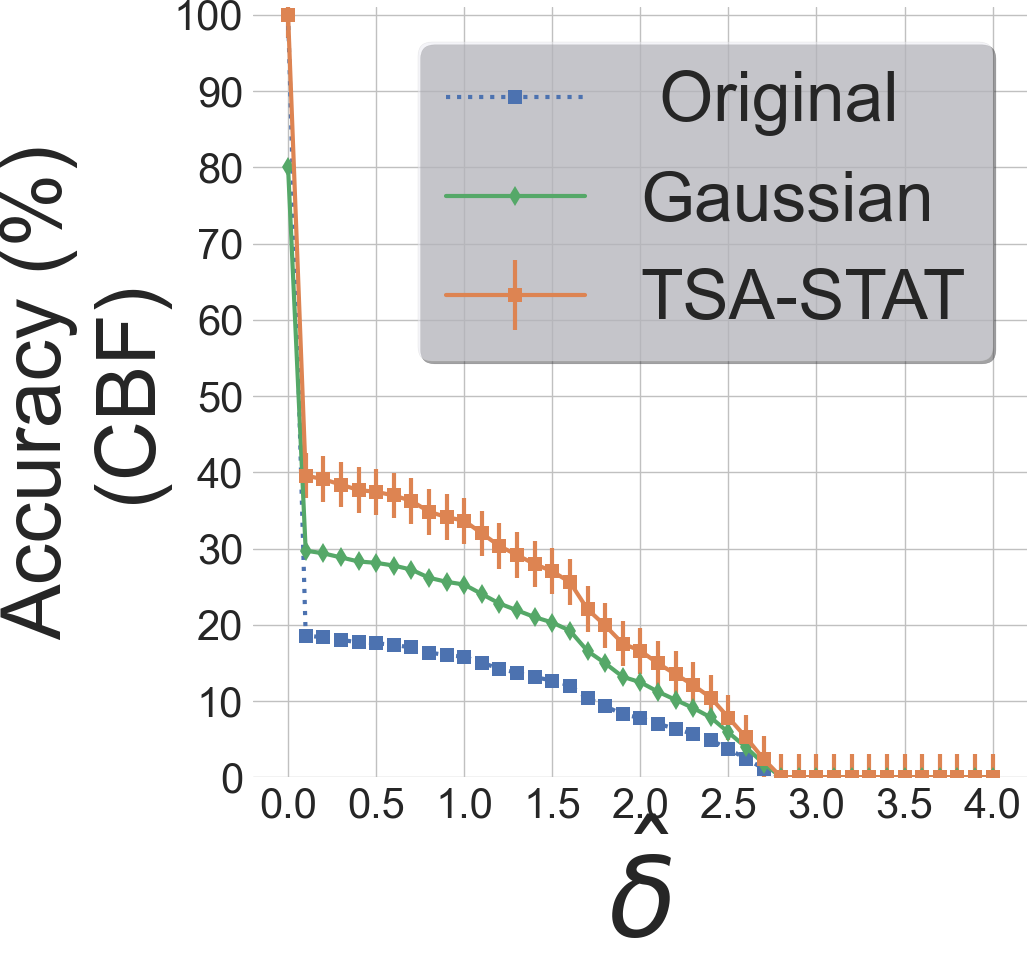}
        \end{minipage}%
        \begin{minipage}{.33\linewidth}
            \centering
            \includegraphics[width=\linewidth]{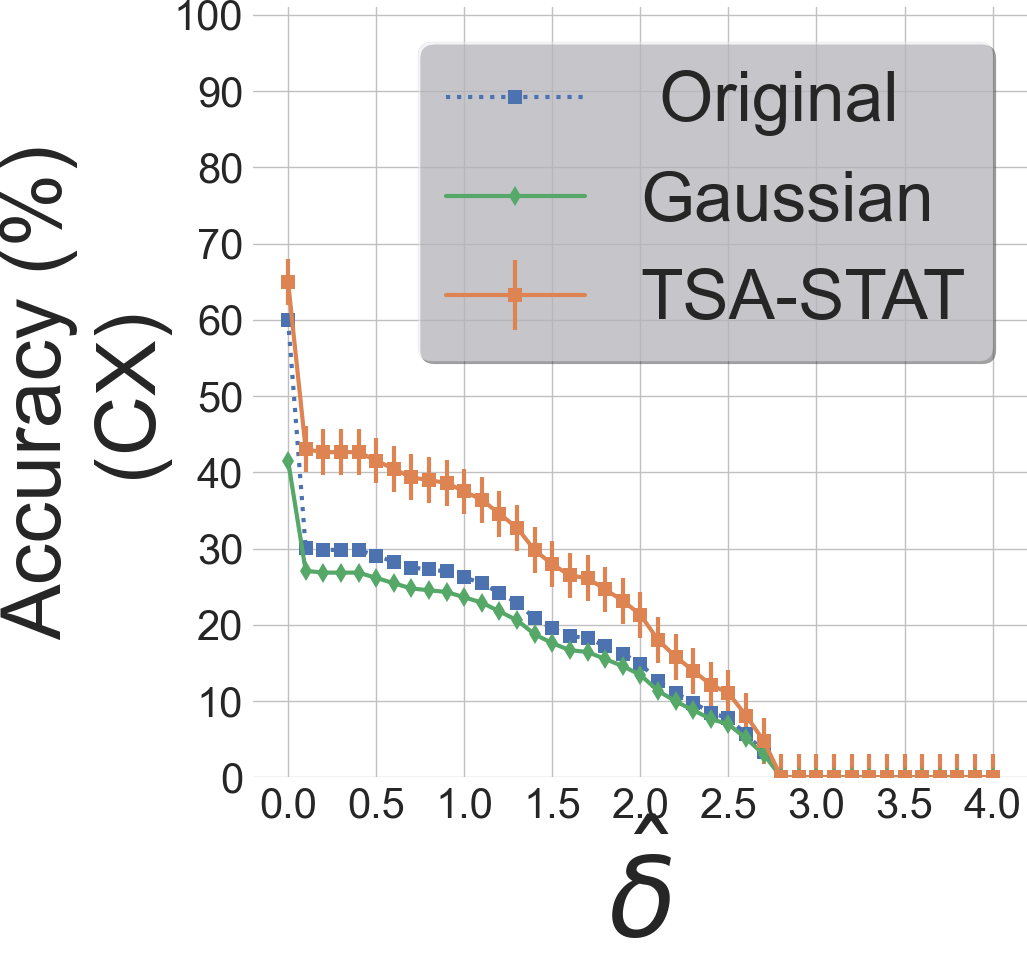}
        \end{minipage}
        \begin{minipage}{.33\linewidth}
            \centering
            \includegraphics[width=\linewidth]{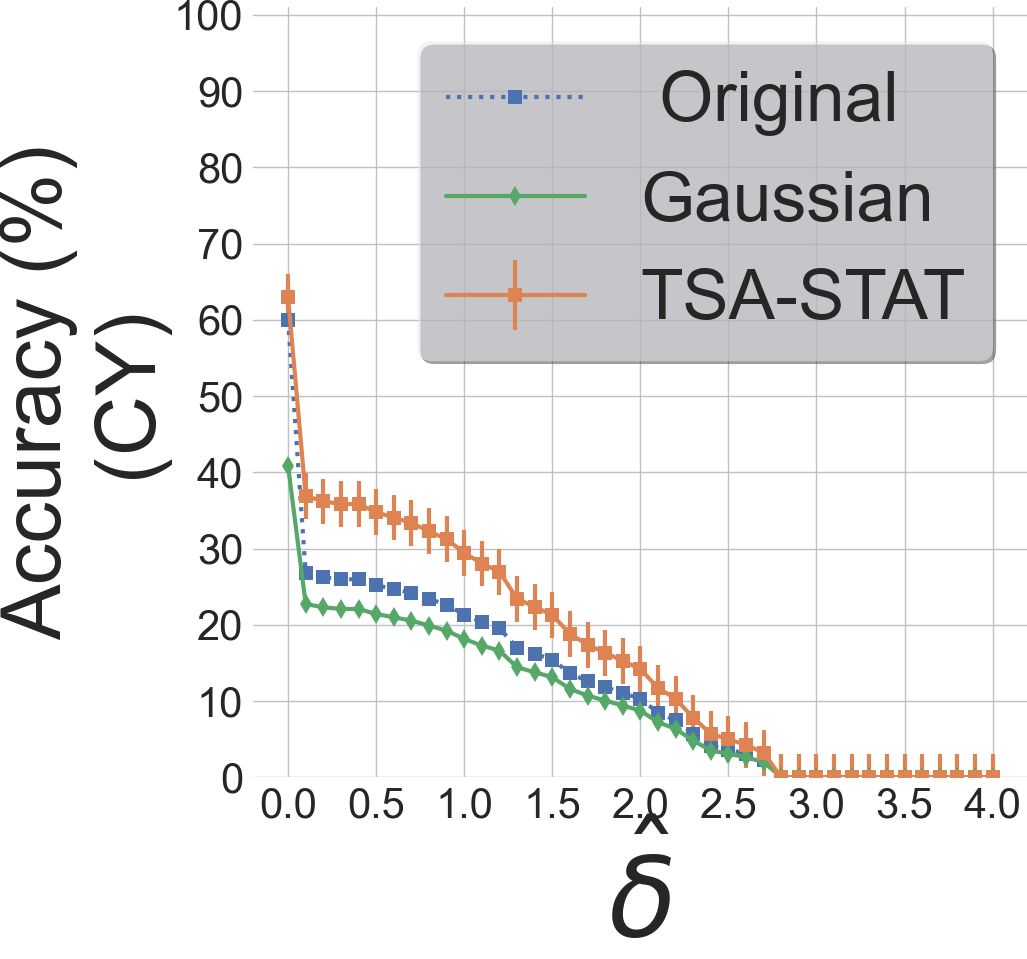}
        \end{minipage}%
        \begin{minipage}{.33\linewidth}
            \centering
            \includegraphics[width=\linewidth]{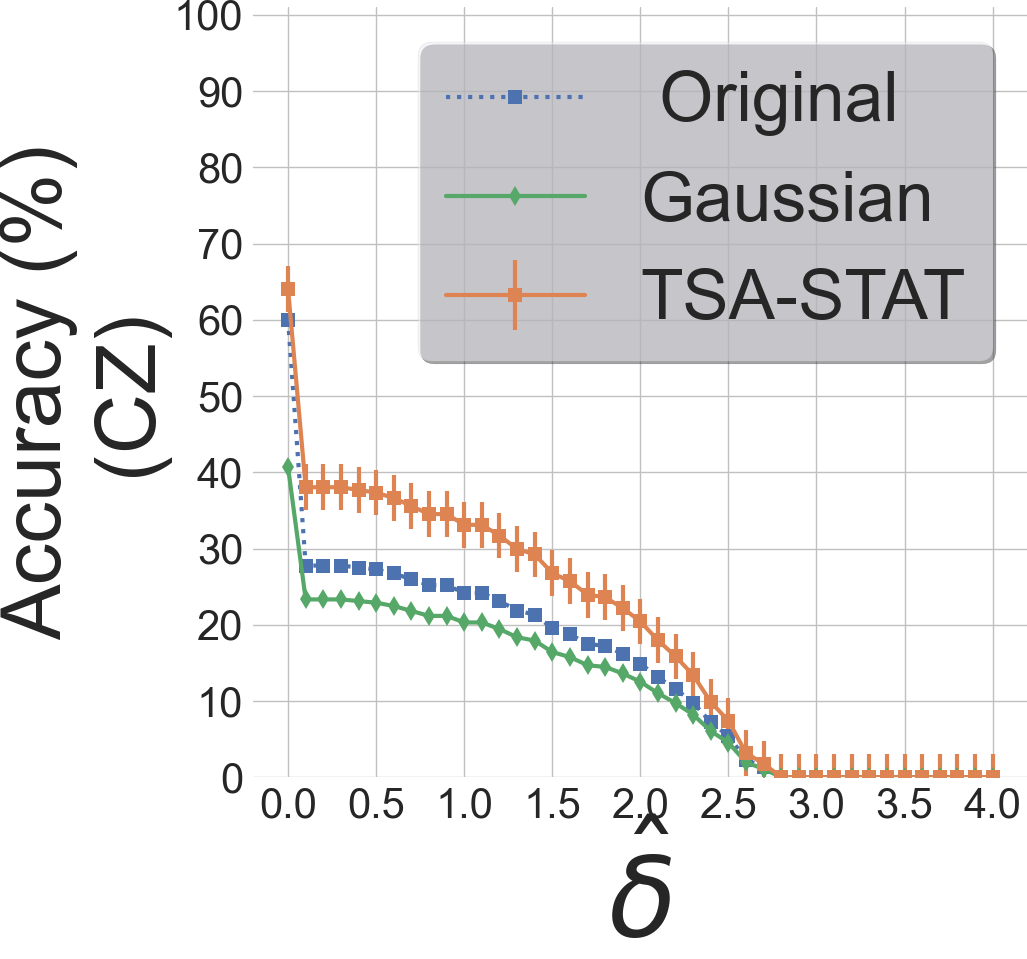}
        \end{minipage}%
        \begin{minipage}{.33\linewidth}
            \centering
            \includegraphics[width=\linewidth]{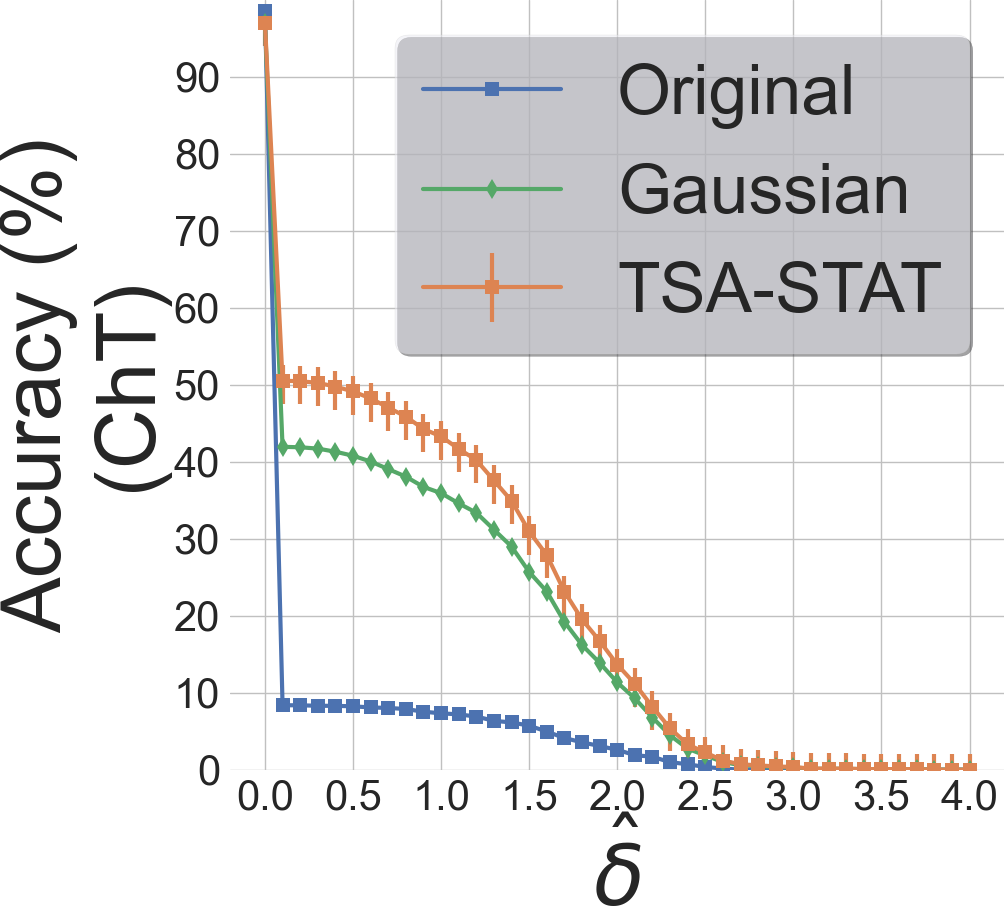}
        \end{minipage}
    \end{minipage}
    \caption{Robustness results with adversarial training. Comparison of the accuracy of Original model (standard training without adversarial examples) and adversarial training based on TSA-STAT and Gaussian augmentation. This figure illustrates that TSA-STAT is a better method to improve the robustness of deep model as it has the highest accuracy for a given $\hat{\delta}$ for most datasets.}
    \label{fig:robust}
        \vspace{-2em}
\end{figure}

Figure \ref{fig:robust} shows the robustness in accuracy of the deep model against perturbation $\hat{\delta}$ (In blue). It illustrates the classification accuracy on the testing set under attacks with different possible $\hat{\delta}$ values using $(\mu_P$=$0.01, \sigma$=$0.1, MAX$=$10^3)$ as parameters of the multivariate Gaussian for Algorithm \ref{alg:certif}. Consider the analysis for WD dataset as an example. At $\hat{\delta}=0$, we have a test accuracy of 0.83, which translates to the fact that 83\% of the test set inputs are robust to the given perturbation and 17\% of the test is vulnerable to adversarial attacks. At $\hat{\delta}$=$1$, the plot shows that around 28\% of the dataset has a certified bound $\delta \ge 1$.

The same figure shows the influence of adversarial training on the certification bound of the deep model via Algorithm \ref{alg:certif}. We also provide a comparison using Gaussian augmentation \cite{dodge2017study} to show the substantial role of TSA-STAT in using statistical features vs. using a standard Gaussian noise. We can observe the effect of adversarial training with TSA-STAT on increasing the robustness of most inputs on the different datasets. For example, on HAPT dataset, the initial region of $\hat{\delta} \le 1.5$, the robustness of several inputs have increased (20\% increase at $\hat{\delta}$=$0$ and 10\% increase at $\hat{\delta}$=$1.0$).

\vspace{1.0ex}

\noindent \textbf{Transferability of attacks.} Prior work has shown that RNN models are competitive with 1D-CNNs for time-series domain.  Therefore, we evaluate the transferability of adversarial examples from $WB$ (1D-CNN) to an RNN model. Table \ref{tab:lstm} shows the percentage of dataset that has targeted adversarial capabilities to fool a Long short-term memory (LSTM) model and a Gated recurrent unit (GRU) model using  TSA-STAT. We can observe that TSA-STAT attacks have the transfer potential to fool other deep models such as RNNs.  From Table \ref{tab:lstm}, we make the following observations. First, TSA-STAT is able to generate targeted adversarial examples that are able to fool RNN models. Second, the fooling efficiency increases $\alpha_{Eff}\rightarrow 0.8$ if we employ TSA-STAT in an untargeted setting for most of the datasets. Since this paper only studies the setting of no queries to the target deep model, the attacks would have an increased efficiency if the target deep model is available for label queries \cite{papernot2017practical}. Finally, we observe poor attack performance specifically on the ChT dataset under any black-box setting. This low performance is not restricted to the attacks from TSA-STAT. CW attacks on ChT also have poor transferability performance in black-box settings ($\alpha_{Eff}\le 0.1$). The analysis of robustness of this dataset is shown in Figure \ref{fig:robust}. We can clearly observe the low robustness performance and resilience to noise of the original model. Hence, specific analysis is needed to adapt to  datasets such as ChT, where adaptive attacks should be pursued \cite{tramer2020adaptive}.
\begin{table}[!t]
\centering
\caption{\centering Results for transferability of TSA-STAT attacks across RNN models.}
\begin{tabular}{|l|l|l|l|l|l|l|l|l|l|} 
\hline
  & \textbf{CC}  & \textbf{ SC} & \textbf{HAPT}  & \textbf{WD} & \textbf{CBF} & \textbf{CX} & \textbf{CY} & \textbf{CZ} & \textbf{ChT}    \\
 \hline
LSTM & 39\% &80\%& 37\%  & 50\% & 89\% & 58\% & 58\% & 57\% & 5.7\%\\ \hline
GRU & 49\% &71\%& 43\%  & 41\% & 87\% & 56\% & 53\% & 55\% & 5.5\%\\ \hline
\end{tabular}
\label{tab:lstm}
\end{table}

\vspace{1ex}
\noindent \textbf{Comparison with the work of \cite{karim2020adversarial}.} We mentioned in the related Work section that there is a recent work that proposed an approach for studying adversarial attacks for the time-series domain \cite{karim2020adversarial}. This method employs network distillation to train a student model for creating adversarial attacks. We provide a comparison between TSA-STAT and the network distillation approach to show the effectiveness of our proposed framework.  First, the method in \cite{karim2020adversarial} is severely limited: only a small number of target classes yield to a generation of adversarial examples and the method does not guarantee a generation of adversarial example for every input. \cite{karim2020adversarial} showed that for many datasets, this method creates a limited number of adversarial examples in the white-box setting. To test the effectiveness of this attack against TSA-STAT, we employ adversarial training using adversarial examples generated by the model proposed in \cite{karim2020adversarial} under the black-box setting. We use the code \footnote{https://github.com/titu1994/Adversarial-Attacks-Time-Series.git} provided by the authors to generate the adversarial examples using this baseline method.

\begin{figure}[!h]
    \centering    
    \begin{minipage}{\linewidth}
    \centering
        \includegraphics[width=.8\linewidth]{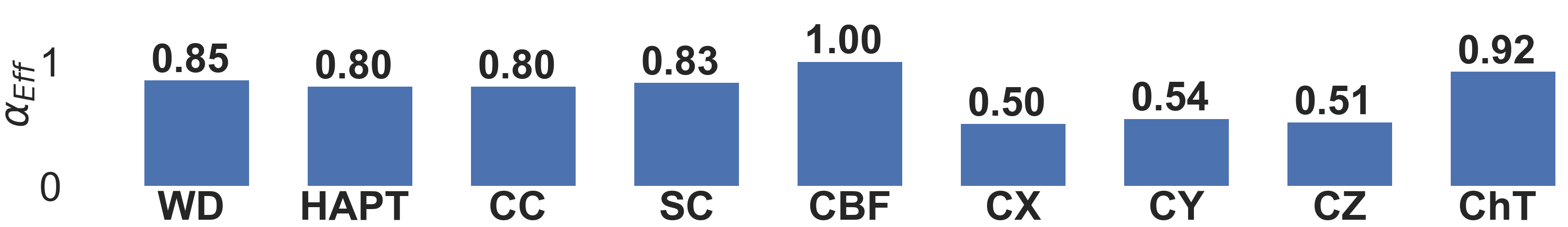}
        \caption{Results for effectiveness of TSA-STAT on  deep models via adversarial training using the augmented data generated from \cite{karim2020adversarial}.}
        \label{fig:karim}
    \end{minipage}
    \hspace{1em}
    \begin{minipage}{\linewidth}
    \centering
        \begin{minipage}{.8\linewidth}
            \includegraphics[width=\linewidth]{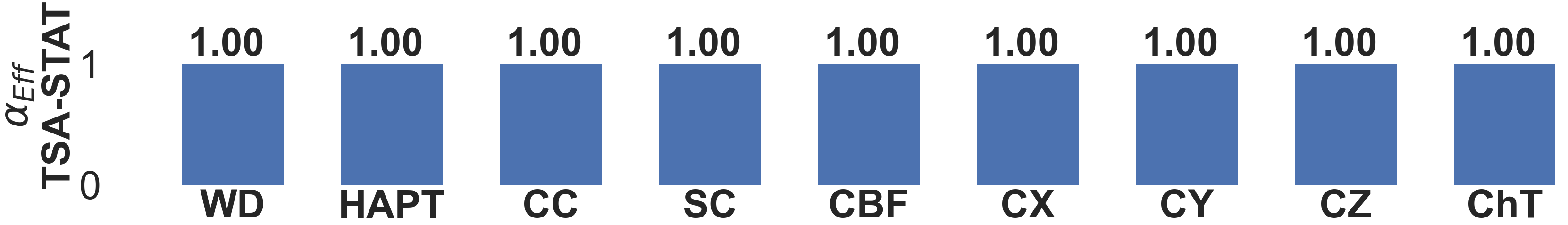}
        \end{minipage}
        \begin{minipage}{.8\linewidth}
            \includegraphics[width=\linewidth]{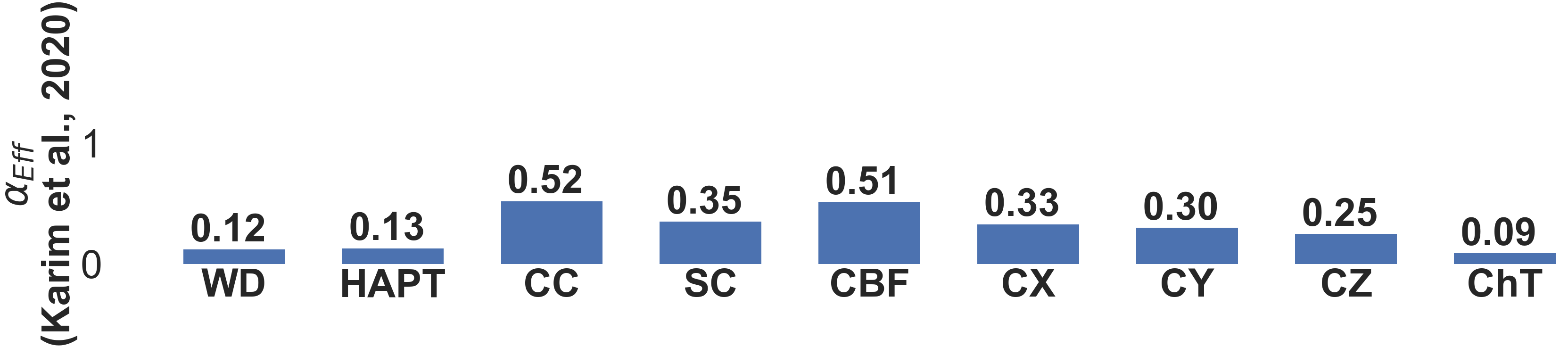}
        \end{minipage} 
        \caption{Results for the effectiveness of TSA-STAT and \cite{karim2020adversarial} method under the white-box setting $WB$.}
        \label{fig:karim3}
\end{minipage}    
    \begin{minipage}{\linewidth}
    \centering
         \includegraphics[width=.8\linewidth]{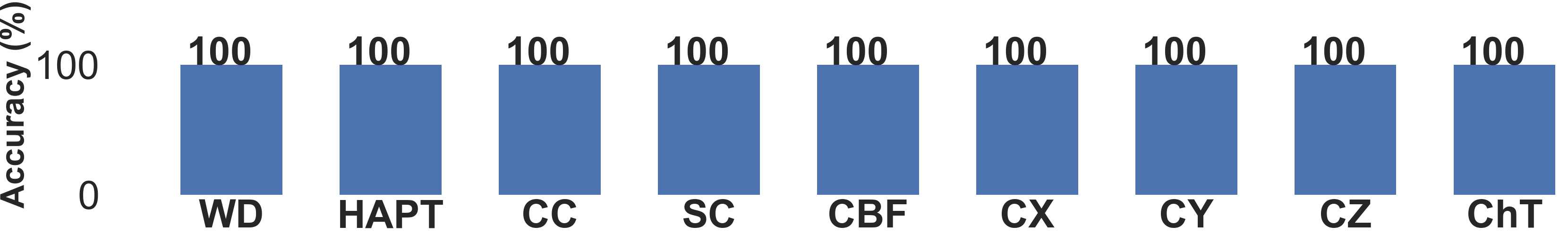}
        \caption{Results of TSA-STAT based adversarial training performance on predicting the true labels of adversarial attacks generated by \cite{karim2020adversarial}.}
        \label{fig:karim2}
    \end{minipage}%
    \vspace{-1em}
\end{figure}

Figure \ref{fig:karim} shows the fooling rate of TSA-STAT generated attacks on different datasets. We can conclude that adversarial training using \cite{karim2020adversarial} does not improve the robustness of the models against our proposed attack. Additionally, we show a direct comparison between TSA-STAT and \cite{karim2020adversarial} using the attack performance in Figure \ref{fig:karim3}. This figure shows the results comparing both attack performances under the white-box setting $WB$. We observe that the attack success rate ($\alpha_{ref}$) of TSA-STAT outperforms the adversarial attacks created by \cite{karim2020adversarial} method. Figure \ref{fig:karim2} shows the effectiveness of adversarial examples generated from \cite{karim2020adversarial} on the deep models created via adversarial training using augmented data from TSA-STAT. We can see that using TSA-STAT for adversarial training results in a robust model against any attack generated by the method in \cite{karim2020adversarial}.

\subsection{Summary of Key Experimental Findings}

Our comprehensive experimental evaluation demonstrated that TSA-STAT is an effective adversarial framework for time-series domain. We briefly summarize the main experimental findings below.
\begin{itemize}
\setlength\itemsep{0em}
    \item The similarity measure based on statistical features of time-series used by TSA-STAT is more effective in capturing the unique characteristics of time-series data when compared to the standard algorithms which rely on $l_p$-norm distance (Figure \ref{fig:tsne}).
    \item Figures \ref{fig:singPerf} and \ref{fig:univPerf} demonstrate that the instance-specific and universal adversarial attacks created by TSA-STAT are very effective in fooling DNNs for time-series classification tasks and evading adversarial training based on adversarial examples created by prior methods.
    \item Adversarial examples created by TSA-STAT provide better true-label guarantees (examples belonging to the semantic space of true label) than those based on prior methods relying on $l_p$-norm distance. As a result, adversarial training based on TSA-STAT improves the robustness of deep models more than adversarial training with prior methods (Figure \ref{fig:advTrnPerf}).
    \item Figure \ref{fig:robust} demonstrates that adversarial training based on TSA-STAT provides better robustness certification for time-series classifiers than prior methods.
    \item Table \ref{tab:lstm} results show that TSA-STAT supports transferability: optimized polynomial transformations can be reused to create effective adversarial examples for unseen deep models and time-series signals.
    
\end{itemize}

\section{Conclusions}

We introduced the TSA-STAT framework to study adversarial robustness of deep models for  time-series domain. TSA-STAT relies on two key ideas to create more effective adversarial examples for the time-series domain: 1) Constraints over statistical features of time-series signals to preserve similarities between original input and adversarial examples; and 2) Polynomial transformations to expand the space of valid adversarial examples compared to prior methods. TSA-STAT synergistically combines these two key ideas to overcome the drawbacks of prior methods from the image domain which rely on $l_p$-distance and are not suitable for the time-series domain. We provided theoretical and empirical analysis to explain the importance of these two key ideas in making TSA-STAT more suitable to create adversarial attacks for the time-series domain. We also provided certification guarantees for adversarial robustness of the TSA-STAT framework. We theoretically derived the computation of certification bound for TSA-STAT and provided a concrete algorithm that can be used with any deep model for the time-series domain. Finally, we empirically demonstrated the effectiveness of TSA-STAT on diverse real-world datasets and different deep models in terms of fooling rate and improved robustness with adversarial training. Our work concludes that time-series domain requires separate investigation for robustness analysis due to its unique characteristics and shows the effectiveness of the TSA-STAT framework towards this goal.

\vspace{1.0ex}

\noindent {\bf Acknowledgements.} This research is supported in part by the AgAID AI Institute for Agriculture Decision Support, supported by the National Science Foundation and United States Department of Agriculture - National Institute of Food and Agriculture award \#2021-67021-35344.
\newpage
\appendix

\section{Proofs}
\label{append:proof}
\subsection{Proof of Theorem \ref{th:poly}}
\textit{For a given input space $\mathbb{R}^{n\times T}$ and $d \ge 1$, polynomial transformations allow more candidate adversarial examples than additive perturbations in a constrained space. If $X\in \mathbb{R}^{n\times T}$ and $\mathcal{PT}:X\rightarrow \sum_{k=0}^{d} a_k~X^k$, then $\forall X_{adv}$ s.t. $\|S_i(X_{adv})-S_i(X)\|_{\infty} \le \epsilon_i$:
\begin{align*}
\bigg\{X_{adv}=\mathcal{PT}(X), ~\forall a_k  \bigg\} 
\supsetneq
\bigg\{X_{adv}=X+\delta, ~\forall \delta  \bigg\}
\end{align*}
, $S_i \in  \mathcal{S}^m(X)\bigcup I$dentity.}

Let $X \in \mathbb{R}^{n\times T}$ and $d \ge 1$. Let $\mathcal{PT}(\cdot)$ a polynomial adversarial transformation such that $\mathcal{PT}:X\rightarrow \sum_{k=0}^{d} a_k~X^k$. We want to prove that a polynomial transformation can create an adversarial example $X_{adv}$ that is out of the scope for additive perturbation with a constant $\delta$. The main condition on $X_{adv}$ is that $\|S_i(X_{adv})-S_i(X)\|_{\infty} \le \epsilon_i$ with $S_i \in  \mathcal{S}^m(X)\bigcup I$dentity.

In other words, if the given condition is satisfied, we will have:
\begin{align*}
\bigg\{X_{adv}=\mathcal{PT}(X), ~\forall a_k  \bigg\} 
\supsetneq
\bigg\{X_{adv}=X+\delta, ~\forall \delta  \bigg\}
\end{align*}
Suppose $\mathcal{A}$ be the space of all possible adversarial examples $\big\{X_{adv}=\mathcal{PT}(X), ~\forall a_k  \big\}$ and $\mathcal{B}$ be the space of all possible adversarial examples  $\big\{X_{adv}=X+\delta, ~\forall \delta  \big\}$

\hspace{1em}$\bullet$ $S_i = I$dentity: For $X_{adv}=\mathcal{PT}(X)$:
\begin{align*}
    \|X_{adv}-X\|_{\infty} \le \epsilon_i\\
    \|\sum_{k=0}^{d} a_k~X^k - X\|_{\infty} \le \epsilon_i\\
    \|a_0 + (a_1-1)X +\sum_{k=0}^{d} a_k~X^k\|_{\infty} \le \epsilon_i
\end{align*}
Without loss of generality, let us consider $\|\cdot\|_{\infty}$ on the component $l\le n$.
\begin{align*}
   |a_0 + (a_1-1)X_l +\sum_{k=0}^{d} a_k~X_l^k| \le \epsilon_i\\
    |a_0 + \beta(\{a_k, X_l\})| \le \epsilon_i
\end{align*}
Then $X_{adv} \in \mathcal{B}$ only if the function $\beta(\{a_k, X_l\})=0$ and $|a_0|\le \epsilon_i$. Hence, by construction on the set of $\{a_k\}$, if $|a_0|> \epsilon_i$, we can create $X_{adv}$ such that $|a_0 + \beta(\{a_k, X_l\})| \le \epsilon_i$. Hence, we have $X_{adv} \in \mathcal{A}$ and $X_{adv} \notin \mathcal{B}$ ($\beta$ depends on $X$, so it cannot be considered as a constant perturbation $\delta$ to be in $\mathcal{B}$).

\hspace{1em}$\bullet$ $S_i \in \mathcal{S}^m(X)$: Let us start with $S_i(\cdot) = \mu(\cdot)$. Similar to the previous case, and if we consider $\|\cdot\|_{\infty}$ on the component $l\le n$:
\begin{align*}
    \|\mu(X_{adv})-\mu(X)\|_{\infty} \le \epsilon_i\\
    \bigg|\mu\bigg(\sum_{k=0}^{d} a_k~X_l^k\bigg) - \mu(X_l)\bigg| \le \epsilon_i\\
    \bigg|\sum_{j=0}^T\frac{\sum_{k=0}^{d} a_k~X_{l,j}^k}{T} - \sum_{j=0}^T\frac{X_{l,j}}{T}\bigg| \le \epsilon_i\\ \bigg|\sum_{j=0}^T\frac{a_0 + (a_1-1)X_{l,j} +\sum_{k=0}^{d} a_k~X_{l,j}^k}{T}\bigg| \le \epsilon_i
\end{align*}
If $X_{adv} \in \mathcal{B}$, then $\|\mu(X_{adv})-\mu(X)\|_{\infty}=|\sum_{j=0}^T\frac{a_0}{T}|$. With the same construction logic as in the previous case, we can end with $X_{adv} \in \mathcal{A}$ and $X_{adv} \notin \mathcal{B}$. For the remaining cases of $S_i(\cdot)$ used in this work, as they are correletaed with $\mu$, similar construction can be used.

\subsection{Proof of Theorem \ref{th:bound}}
\textit{Let $X\in \mathbb{R}^{n\times T}$ be an input time-series signal. Let $n_P\sim\mathcal{N}(\mu_P, \sum)$ and $n_0 \sim\mathcal{N}(0, \sum)$. Given a classifier $F_{\theta}: \mathbb{R}^{n\times T}\rightarrow Y$ that produces a probability distribution $(p_1,\cdots,p_k)$ over $k$ labels for $F_{\theta}(X+n_P)$ and another probability distribution $(p^0_1,\cdots,p^0_k)$ for $F_{\theta}(X+n_0)$. To guarantee that $\displaystyle arg\max_{p_i} ~ p_i = \displaystyle arg\max_{p^0 } ~ p^0$, the following condition must be satisfied:
\begin{center}
$\|\mu_P\|^2_{\infty} \le  \displaystyle \max_{\alpha \neq 1} \frac{2}{\alpha \cdot \sum\nolimits^{(S)}}  \cdot \left( -ln \left(1-p_{(1)}-p_{(2)} +2\left( \frac{1}{2}\left( p_{(1)}^{1-\alpha} + p_{(2)}^{1-\alpha} \right)  \right)^{\frac{1}{1-\alpha}} \right)\right)$
\end{center}
where $\|\mu_P\|_{\infty}$ is the maximum perturbation over the mean of the input's channels and $\sum\nolimits^{(S)}$ is the sum of all elements of ~$\sum$. }

To prove this theorem, we call for a second Lemma  provided in \cite{li2019certified}):
\begin{lemma}
Let $EP$ and $E0$ be two probability distributions where $EP$=$(p_1, \cdots, p_k)$ and $E0$=$(p^0_1, \cdots, p^0_k)$. If $\displaystyle arg\max_{p_i \in EP} ~ p_i \neq \displaystyle arg\max_{p^0_i \in E0} ~ p^0_i$, then:
\begin{equation}
\begin{split}
     D_{\alpha}(EP\|E0) \ge  -ln \left(1-p_{(1)}-p_{(2)}+2\left( \frac{1}{2}\left( p_{(1)}^{1-\alpha} + p_{(2)}^{1-\alpha} \right)  \right)^{\frac{1}{1-\alpha}}  \right) 
     \end{split}
     \label{eq:lem26}
\end{equation}
where $p_{(1)}$ and $p_{(2)}$ are respectively the largest and second largest $p_i\in EP$.
\label{lem:upbound}
\end{lemma}
This Lemma provides a lower bound of the Rényi divergence for changing the index of the maximum of $EP$, which is useful for the derivation of our certification bound. If the estimated distributions $EP$ and $E0$ have different indices for the maximum class probabilities, then $ D_{\alpha}(EP\|E0) < $ RHS of Equation \ref{eq:lem26}. 

Let $X\in \mathbb{R}^{n\times T}$ an input time-series signal, $n_P\sim\mathcal{N}(\mu_P, \sum)$ and $n_0 \sim\mathcal{N}(0, \sum)$, and a DNN classifier $F_{\theta}: \mathbb{R}^{n\times T}\rightarrow Y$ that produces a probability distribution over $k$ candidate class labels: $EP$=$(p_1,\cdots,p_k)$ for $F_{\theta}(X+n_P)$ and another probability distribution $E0$=$(p^0_1,\cdots,p^0_k)$ for $F_{\theta}(X+n_0)$. \\

As a direct result from Lemma \ref{lemma:mvgdist}:
\begin{center}
    
$D_\alpha(EP\|E0) = \frac{\alpha}{2}(\mu_P-0)^T\sum_{\alpha}(\mu_P-0) - \frac{1}{2(\alpha-1)}~ ln\frac{|\sum_{\alpha}|}{|\sum|^{1-\alpha}|\sum|^{\alpha}}$
\end{center}
where $\sum_{\alpha}=\alpha\sum+(1-\alpha)\sum = \sum$. \\

This results to:

\begin{align*}
D_\alpha(EP\|E0) = \frac{\alpha}{2}\mu_P^T\sum\mu_P - \frac{1}{2(\alpha-1)}~ ln(1)\\
=\frac{\alpha}{2}\mu_P^T\sum\mu_P~~~~~~~~~~~~~~~~~~~~~~~~~~~~
\end{align*}

Since $\forall i: \mu_{P, i} \le \|\mu_P\|_{\infty}$, we get
\begin{align*}
D_\alpha(EP\|E0) =\frac{\alpha}{2}\mu_P^T\sum\mu_P
=\frac{\alpha}{2} \displaystyle \sum_i \displaystyle \sum_j \mu_{P, i}\times \mu_{P, j}\times \textstyle \sum_{i,j}\\
\le \frac{\alpha}{2} \|\mu_P\|_{\infty}^2 \sum\nolimits^{(S)}~~~~~~~~~~~~~~~~~~~~~~~~~~~~~~~~~~~~~~~~~~~~
\end{align*}
where $\sum\nolimits^{(S)}$ is the sum of all elements of ~$\sum$.

To guarantee that $\displaystyle arg\max_{p_i} \; p_i = \displaystyle arg\max_{p^0_i } \; p^0_i$, the following condition must be satisfied from Lemma \ref{lem:upbound}:
\begin{align*}
    D_\alpha(EP\|E0) < -ln(1-p_{(1)}-p_{(2)}+2\left( \frac{1}{2}\left( p_{(1)}^{1-\alpha} + p_{(2)}^{1-\alpha} \right)  \right)^{\frac{1}{1-\alpha}} 
\end{align*}
This implies:
\begin{align*}
    \frac{\alpha}{2} \|\mu_P\|_{\infty}^2 \sum\nolimits^{(S)} < -ln(1-p_{(1)}-p_{(2)}+2\left( \frac{1}{2}\left( p_{(1)}^{1-\alpha} + p_{(2)}^{1-\alpha} \right)  \right)^{\frac{1}{1-\alpha}}
\end{align*}
which leads us to:
\begin{align*}
     \|\mu_P\|_{\infty}^2 < \frac{2}{\alpha\times \sum\nolimits^{(S)}} \times \left(  -ln(1-p_{(1)}-p_{(2)}+2\left( \frac{1}{2}\left( p_{(1)}^{1-\alpha} + p_{(2)}^{1-\alpha} \right)  \right)^{\frac{1}{1-\alpha}}\right)
\end{align*}
Hence, our result.

\vspace{2.0ex}

\noindent While the proposed certification in Theorem 2 uses the Lemma3 provided in \cite{li2019certified}, the derivation of the bound is different in terms of the following aspects that are not covered in \cite{li2019certified} :
\begin{itemize}
    \item The use of a multivariate Gaussian distribution of the noise. The main difference in our work is that we use a multivariate Gaussian distribution (Lemma 1) that is characterized by mean vector $\mu$ and a covariance matrix $\sum$. This general formulation of multivariate Gaussian distributions results in the conclusion of Theorem 1 because it is applied on multivariate inputs $X\in \mathcal{R}^{n\times T}$. However, the standard Gaussian noise with $0$-value mean and a single-value standard deviation $\sigma$ employed in (Li et. al., 2019) is limited for our case. The theoretical analysis in \cite{li2019certified} is not general to the multivariate Gaussian distributions used in our work. This is due to the fact that the Renyi divergence of the noise distribution in (Li et. al., 2019) can be upper-bounded by a factor of the $L_2$ norm. This upper bound is not applicable for the multivariate Gaussian noise. Hence, we provide the proof of Theorem 2 to derive a certification robustness. The approach in \cite{li2019certified} can only be used for {\em univariate} time-series data. We provided a more general derivation for multi-variate time-series.
    \item The certification in \cite{li2019certified} is only provided for the Euclidean distance. For the theoretical analysis of TSA-STAT, we had to introduce the statistical features of time-series instead of euclidean distance. Hence, we proposed in our proof the use of a mean vector $\mu_P$. This proof is not similar to the one provided in \cite{li2019certified}. Additionally, we provided Lemma 2 with proof to extend the certification to other statistical features.
\end{itemize}

\subsection{Proof of Lemma \ref{lem:otherbounds}}
\textit{If a certified bound $\delta$ has been generated for the mean of input time-series signal $X\in \mathbb{R}^{n\times T}$ and classifier $F_{\theta}$, then certified bounds for other statistical/temporal features can be derived consequently.}

In this section, we will work on other statistical constraints used in our experimental evaluation. Let $X\in \mathbb{R}^{n\times T}$ an time-series input signal. Let $\Sigma$ be the positive semi-definite covariance matrix used for the additive multivariate Gaussian noise. The bound on the mean $\delta$ value is given by TSA-STAT certification algorithm. As $\|S_i(X)\|_{\infty}$ is equal to the value of $S_i$ on one of the  channels $n$, let us consider for simplicity of this proof only that channel. Hence, the derivation of the bounds for other statistical features is as follows:
\begin{itemize}
    \item RMS =  $\sum \frac{x_i^2}{n}$
    \begin{equation*}
        \sigma^2 = \sum \frac{(x_i-\mu)^2}{n} = \sum \frac{x_i^2-2x_i\mu+\mu^2}{n} = RMS^2 - 2\mu\sum\frac{x_i}{n}+\sum\frac{\mu^2}{n} = RMS^2-\mu^2
    \end{equation*}
    $\Rightarrow \text{max}\|RMS\|_{\infty}=\delta^2+\sigma^2$
    \item Skewness $g = \sum\frac{(x_i-\mu)^3}{n\times \sigma^3}$
    
Let $G(\mu)= \sum\frac{(x_i-\mu)^3}{n}$
    \begin{equation*}
    \frac{\partial G}{\partial \mu} = \sum \frac{\partial}{\partial \mu} \frac{(x_i-\mu)^3}{n} = -3 \times \sum \frac{(x_i-\mu)^2}{n} \neq 0 ~\forall \mu ~\text{as ($\sigma \neq 0$)}
    \end{equation*}
    Therefore, $G(\mu)$ is monotonic $\Rightarrow  \text{max}\|g\|_{\infty}=\frac{G(\delta)|}{\sigma^3}$
    \item Kurtosis $k = \sum\frac{(x_i-\mu)^4}{n\times \sigma^4}-3$
    
Let $K(\mu)= \sum\frac{(x_i-\mu)^4}{n}$, following the previous result on the skewness:
    \begin{equation*}
    \frac{\partial K}{\partial \mu} = \sum \frac{\partial}{\partial \mu} \frac{(x_i-\mu)^4}{n}  \neq 0 ~\forall \mu
    \end{equation*}
    Therefore, $K(\mu)$ is monotonic $\Rightarrow  \text{max}\|k\|_{\infty}=\frac{|K(\delta)|}{\sigma^4}-3$ 
\end{itemize}
\newpage
\bibliography{zaistats}
\bibliographystyle{theapa}

\end{document}